%% file: tcc.tex
\pdfoutput=1
\documentclass[english, brazil, bsc]{dcomp-abntex2}

\usepackage{caption}			% Para legendas em floats
\usepackage{subcaption}			% Para apresentação de subfloats
\usepackage{media9}				% Para vídeos

\makeindex

\begin{document}
	
	% Seleciona o idioma do documento (conforme pacotes do babel)
	\selectlanguage{brazil}
	
	% Retira espaço extra obsoleto entre as frases.
	\frenchspacing 
	
	% ----------------------------------------------------------
	% ELEMENTOS PRÉ-TEXTUAIS
	% ----------------------------------------------------------
	\pretextual
	\titulo{Sistema de Navegação Autônomo Baseado em Visão Computacional}
	\autor{Michel Conrado Cardoso Meneses}
	\orientador{Prof. Dr. Leonardo Nogueira Matos}
	\coorientador{Prof. Dr. Bruno Otavio Piedade Prado}
	\curso{Engenharia de Computação}
	
	\imprimircapa
	\imprimirfolhaderosto*

	\include{Pre_Textual/Dedicatoria}
	\include{Pre_Textual/Agradecimentos}
	\include{Pre_Textual/Epigrafe}
	\include{Pre_Textual/Resumo}
	\include{Pre_Textual/Abstract}
	
	% Lista de Figuras
	\pdfbookmark[0]{\listfigurename}{lof}
	\listoffigures*
	\cleardoublepage
	
	% Lista de Tabelas
	\pdfbookmark[0]{\listtablename}{lot}
	\listoftables*
	\cleardoublepage
	
	\include{Pre_Textual/Abreviaturas}
	\include{Pre_Textual/Simbolos}
	
	\pdfbookmark[0]{\contentsname}{toc}
	\tableofcontents*
	\cleardoublepage
	
	% ----------------------------------------------------------
	% ELEMENTOS TEXTUAIS
	% ----------------------------------------------------------
	\textual
	\include{Conteudo/Introducao}
	\include{Conteudo/TrabalhosRelacionados}
	\include{Conteudo/FluxoOptico}
	\include{Conteudo/PlataformaProposta}
	\include{Conteudo/ExperimentosResultados}
	\include{Conteudo/Conclusao} 
	
	\bibliography{bibliografia}
	
	% ----------------------------------------------------------
	% ELEMENTOS PÓS-TEXTUAIS
	% ----------------------------------------------------------
	\postextual
	
	\renewcommand{\chapnumfont}{\chaptitlefont}
	\renewcommand{\afterchapternum}{}
	\include{Pos_Textual/Apendices}
	
\end{document}

%% file: Pre_Textual/Dedicatoria.tex
\begin{dedicatoria}
   \vspace*{\fill}
   \centering
   \noindent
   \textit{À minha família e a todos aqueles que de alguma\\ 
   	forma contribuíram para esta conquista.} \vspace*{\fill}
\end{dedicatoria}
% ---

%% file: Pre_Textual/Agradecimentos.tex
\begin{agradecimentos}

A realização deste trabalho, juntamente com todo o contexto que o cerca, deve-se aos seguintes envolvidos:

\begin{itemize}
	\item Minha família, maior apoiadora e principal inspiração para os meus sonhos;
	\item Os amigos que estiveram ao meu lado durante esta jornada, com os quais pude compartilhar experiências e aprendizados valiosíssimos;
	\item Os professores que se empenharam em transmitir seu conhecimento e cujos ensinamentos jamais serão esquecidos;
	\item Todos os colegas e instituições que direta ou indiretamente contribuíram para esta realização.
\end{itemize}

A todos, minha imensurável gratidão.

\end{agradecimentos}
% ---

%% file: Pre_Textual/Epigrafe.tex
\begin{epigrafe}[]
    \vspace*{\fill}
	\begin{flushright}
	
		\textit{Your time is limited, so don't waste it\\ 
			living someone else's life. \\
			Don't be trapped by dogma - which is\\
			living with the results of other people's thinking.\\ 
			Don't let the noise of others' opinions\\ 
			drown out your own inner voice.\\ 
			And most important, have the courage\\ 
			to follow your heart and intuition.\\
			(Steve Jobs)}
		
	\end{flushright}
\end{epigrafe}
% ---

%% file: Pre_Textual/Resumo.tex
% resumo em português
\setlength{\absparsep}{18pt} % ajusta o espaçamento dos parágrafos do resumo
\begin{resumo}

Robôs autônomos são utilizados como ferramenta para a solução de diversos tipos de problemas, como o mapeamento e o monitoramento de ambientes. Seja pelas condições adversas à presença humana ou mesmo pela necessidade de reduzirem-se custos, é certo que vários esforços vêm sendo somados para desenvolverem-se robôs com nível de autonomia cada vez maior. Estes devem ser capazes de se locomover em ambientes dinâmicos, sem contar com o auxílio de operadores humanos ou sistemas externos. Nota-se, portanto, que a forma de percepção e modelagem do ambiente torna-se significativamente relevante para a navegação. Dentre os principais métodos de sensoriamento, destacam-se aqueles baseados na visão. Através desta é possível gerar modelos com alto nível de detalhamento acerca do ambiente, uma vez que diversas características podem ser aferidas, tais como textura, cor e luminosidade. No entanto, as técnicas mais precisas de navegação autônoma baseadas em visão apresentam custo computacional superior ao suportado por plataformas móveis de baixo custo, tais como a Raspberry Pi. Esta, por sua vez, vem ganhando cada vez mais espaço em aplicações comerciais e científicas, devido ao seu eficiente gerenciamento de energia e às suas dimensões reduzidas. Sendo assim, o trabalho realizado teve como objetivo desenvolver um robô de baixo custo, controlado por uma Raspberry Pi e cujo sistema de navegação autônomo é baseado em visão. Para tanto, a estratégia utilizada consistiu na identificação de obstáculos a partir do reconhecimento de padrões de fluxo óptico. Através deste sinal é possível inferir a movimentação relativa entre o robô e os demais elementos inseridos no ambiente. Para estimá-lo, utilizou-se o algoritmo de Lucas-Kanade, o qual é capaz de ser executado pela Raspberry Pi sem que seu desempenho seja prejudicado. Finalmente, utilizou-se um classificador SVM para identificar padrões deste sinal associados à movimentação de obstáculos. O sistema desenvolvido foi avaliado considerando-se sua execução sobre uma base de dados formada por padrões de fluxo óptico extraídos de um ambiente real de navegação. Ao fim, seu desempenho foi comparado ao obtido por outros trabalhos relacionados. Constatou-se que a frequência de processamento deste sistema foi superior à apresentada pelos demais. Além disso, sua acurácia e seu custo de aquisição foram, respectivamente, superior e inferior ao da maioria dos trabalhos considerados.
 
 \textbf{Palavras-chave}: Navegação autônoma, Visão computacional, Fluxo óptico, Reconhecimento de padrões, Raspberry Pi.
\end{resumo}

%% file: Pre_Textual/Abstract.tex
% resumo em inglês
\setlength{\absparsep}{18pt} % ajusta o espaçamento dos parágrafos do resumo
\begin{resumo}[Abstract]
 \begin{otherlanguage*}{english}
   
   Autonomous robots are used as the tool to solve many kinds of problems, such as environmental mapping and monitoring. Either for adverse conditions related to human presence or even for the need to reduce costs, it is certain that many efforts have been made to develop robots with increasingly high level of autonomy. They must be capable of locomotion through dynamic environments, without human operators or assistant systems' help. It is noted, thus, that the form of perception and modeling of the environment becomes significantly relevant to navigation. Among the main sensing methods are those based on vision. Through this it is possible to create highly-detailed models about the environment, since many characteristics can be measure, such as texture, color and illumination. However, the most accurate vision based navigation techniques are computationally expensive to run on low-cost mobile platforms, as the Raspberry Pi. This computer, in turn, has been increasingly used in scientific and commercial applications, due to its efficient power supply management and reduced dimensions. Therefore, the goal of this work was to develop a low-cost robot, controlled by a Raspberry Pi, whose navigation system is based on vision. For this purpose, the strategy used consisted in identifying obstacles via optical flow pattern recognition. Through this signal it is possible to infer the relative displacement between the robot and other elements in the environment. Its estimation was done using the Lucas-Kanade algorithm, which can be executed by the Raspberry Pi without harming its performance. Finally, a SVM based classifier was used to identify patterns of this signal associated to obstacles movement. The developed system was evaluated considering its execution over an optical flow pattern dataset extracted from a real navigation environmet. In the end, it was verified that the processing frequency of the system was superior to the others. Furthermore, its accuracy and acquisition cost were, respectively, higher and lower than most of the cited works.
   
 \end{otherlanguage*}
\end{resumo}

%% file: Pre_Textual/Abreviaturas.tex
% ---
% inserir lista de abreviaturas e siglas
% ---

\begin{siglas}
	\item[A]{Ampere}
	\item[ARM]{Advanced RISC Machine}
	\item[AVI]{Audio Video Interleave}
	\item[CI]{Circuito Integrado}
	\item[cm]{Centímetro}
	\item[cos]{Função cosseno}
	\item[DC]{Direct Current}
	\item[DFT]{Discrete Fourier Transform}
	\item[EM]{Expectation Maximization}
	\item[FN]{False Negative}
	\item[FP]{False Positive}
	\item[FP]{Frames Per Second}
	\item[GB]{Gigabytes}
	\item[GPIO]{General Purpose Input/Output}
	\item[GPU]{Graphics Processing Unit}
	\item[GHz]{Giga-hertz}
	\item[GPS]{Global Positioning System}
	\item[HDMI]{High-Definition Multimedia Interface}
	\item[Hz]{Hertz}
	\item[JPEG]{Joint Photographic Experts Group}
	\item[Kg]{Quilograma}
	\item[LAN]{Local Area Network}
	\item[m]{Metro}
	\item[mm]{Milímetro}
	\item[mA]{Miliampere}
	\item[mAH]{Miliampere-hora}
	\item[ms]{Milissegundo}
	\item[MB]{Megabytes}
	\item[MHz]{Mega-hertz}
	\item[PCA]{Principal Component Analysis}
	\item[PNG]{Portable Network Graphics}
	\item[PWM]{Pulse Width Modulation}
	\item[rad]{Radianos}
	\item[RAM]{Random Access Memory}
	\item[RBF]{Radial Basis Function}
	\item[RPM]{Rotations Per Minute}
	\item[sen]{Função seno}
	\item[SD]{Secure Digital}
	\item[SFTP]{SSH File Transfer Protocol}
	\item[SIFT]{Scale-Invariant Feature Transform}
	\item[SSH]{Secure Shell}
	\item[SURF]{Speeded-Up Robust Features}
	\item[SVM]{Support Vector Machine}
	\item[SVR]{Support Vector Regressor}
	\item[TB]{Terabytes}
	\item[TBB]{Threading Building Blocks}
	\item[TN]{True Negative}
	\item[TP]{True Positive}
	\item[UML]{Unified Modeling Language}
	\item[USB]{Universal Serial Bus}
	\item[W]{Watt}
	\item[V]{Volt}
\end{siglas}
% ---

%% file: Pre_Textual/Simbolos.tex
% ---
% inserir lista de símbolos
% ---

\begin{simbolos}
	\item[$ GB/s $] Gigabyte por segundo
	\item[$ K\Omega $] Quilo-ohm
	\item[$ MB/s $] Megabyte por segundo
	\item[$ m/s $] Metro por segundo
	\item[$ \mu $s] Microssegundo
\end{simbolos}
% ---

%% file: Conteudo/Introducao.tex
\chapter{Introdução}
\begin{figure}[htb]
	\centering
	\includegraphics[width=0.5\linewidth]{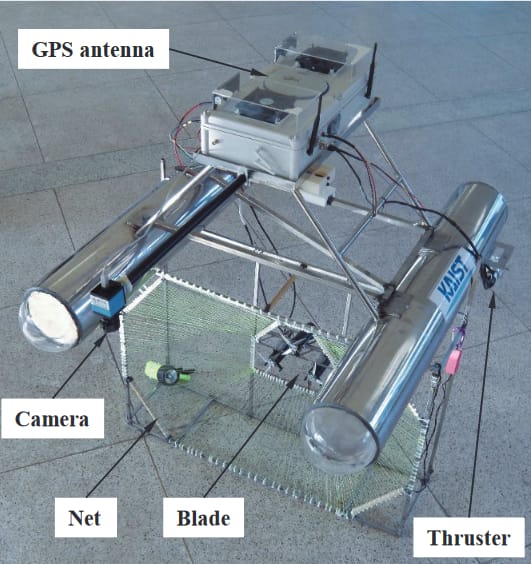}
	\caption{Robô aquático capaz de capturar águas-vivas \cite{Kim2012}.\label{fig:jellyfish_removal_robot}}
\end{figure}

O desenvolvimento de robôs capazes de se locomover de maneira autônoma pode ser considerado um dos temas de estudo mais promissores da robótica \cite{Bekey2005}. Tal importância deve-se à quantidade quase que irrestrita de aplicações relevantes e inovadoras proporcionadas pela utilização destes instrumentos. De uma maneira generalizada, estas aplicações podem ser catalogadas de acordo com uma das seguintes categorias: transporte, busca, mapeamento e monitoramento \cite{Siegwart2004}. No entanto, tal categorização não significa que o uso dessas ferramentas está restrito a um contexto ou a uma área de atuação específica. De fato, atualmente é possível observar a utilização de robôs autônomos como a principal ferramenta para a solução dos mais variados tipos de problemas. Como exemplos, podem-se citar a utilização de robôs aquáticos para a captura de enxames de águas-vivas com o intuito de realizar o controle populacional de tal espécie (\autoref{fig:jellyfish_removal_robot}), o uso de veículos aéreos autônomos para monitorar o tráfego de veículos em grandes metrópoles e assim extrair dados que possam ajudar a diagnosticar e a solucionar problemas no trânsito da região \cite{Wang2015}, a aplicação de robôs em ambientes industriais com o objetivo de transportar mercadorias pesadas entre pontos de carregamento e descarregamento \cite{Kadir2015} e a utilização de robôs para coleta de índices de radiação em áreas contaminadas que apresentam risco à saúde humana \cite{Chaiyasoonthorn2015}.

Além de promissor, o desenvolvimento de robôs que naveguem de forma autônoma também pode ser considerado um tema de estudo extremamente desafiador. De fato, tal desafio pode ser justificado pela complexidade das diferentes etapas que compõem este tipo de navegação. Embora o objetivo final seja fazer com que o robô se locomova de maneira segura entre um ponto de origem e um ponto de destino, tal tarefa é composta por pelo menos quatro subprocessos de natureza multidisciplinar, incluindo as áreas de Computação, Física, Engenharia Mecânica, Engenharia Eletrônica e Biologia \cite{Siegwart2004}. Estes quatro subprocessos podem ser enumerados como:

\begin{enumerate}
	\item Modelagem do ambiente no qual o robô está inserido;
	\item Localização do robô em relação ao modelo gerado do ambiente;
	\item Elaboração da rota a ser percorrida pelo robô;
	\item Controle dos atuadores do robô de modo a garantir o cumprimento da rota elaborada.
\end{enumerate}

O primeiro subprocesso refere-se à elaboração de um mapa do ambiente que envolve o robô. Pelo fato de ser apenas um modelo, este mapa apresenta uma quantidade restrita de informações acerca do ambiente, sendo que seu nível de detalhamento é estipulado a partir das necessidades do robô e das especificações do caso de uso que o mesmo deve realizar. A elaboração deste mapa é feita com base na medição de algumas das características do ambiente, como variação da taxa de ocupação, distribuição de profundidade, geometria, entre outras. A partir do mapa gerado, parte-se para o segundo subprocesso, o qual tem como objetivo identificar a posição do robô neste mapa. Este processo de localização pode ser realizado tanto com base em novas medições do ambiente como também na extração de dados do próprio robô, processo este conhecido na literatura como odometria \cite{Cho2013}. Conhecendo-se a localização do robô no ambiente, é possível então traçar o caminho a ser percorrido. Em geral, escolhe-se o trajeto que apresente maior nível de segurança e menor custo para o robô. Por exemplo, pode-se buscar o caminho que apresente a menor quantidade de obstáculos, que leve o robô ao ponto de destino em menor tempo ou mesmo que apresente superfície com maior regularidade \cite{Ganganath2015}. Finalmente, definido o caminho a ser percorrido parte-se para a última etapa do ciclo de navegação do robô, a qual consiste em definir a sequência de acionamento dos seus atuadores que efetivamente garanta o cumprimento do trajeto estabelecido. Esta sequência de acionamento pode consistir em alterar a direção das rodas do robô, em variar a velocidade de rotação dos seus motores, etc.

\begin{figure}[htb]
	\centering
	\includegraphics[width=0.5\linewidth]{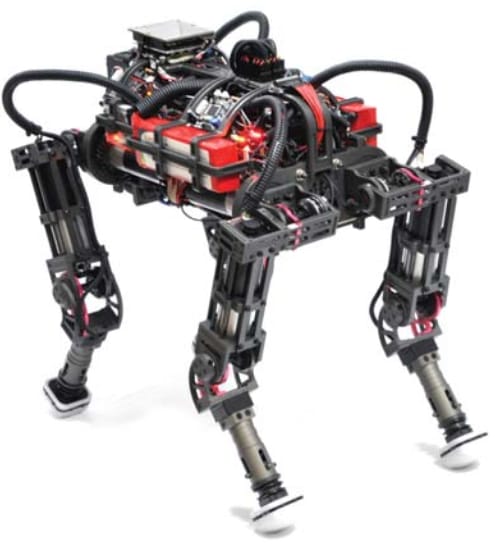}
	\caption{Robô AiDIN III, cujos atuadores são pernas articuladas \cite{Koo_2013}.\label{fig:leg_robot}}
\end{figure}

Obviamente, cada um destes subprocessos pode apresentar maior ou menor relevância a depender das características do contexto no qual o robô está inserido. Por exemplo, há situações nas quais antes mesmo do início da navegação o robô já possui alguma representação do ambiente. Nesses casos, geralmente não há a necessidade de se criar um novo mapa, pois se assume que o ambiente em questão seja estático, isto é, suas características relevantes não são alteradas ao longo do tempo. Há também casos nos quais o robô é capaz de receber, através de outro dispositivo, informações sobre sua própria localização. Nesse tipo de contexto, não há a necessidade de ser estimado tal posicionamento através de medições do ambiente. É o caso da navegação baseada em GPS \cite{Lee2015}. A elaboração do caminho a ser percorrido também apresenta relevância variável a depender das condições impostas ao robô. Há contextos que permitem a construção de tal caminho antes mesmo de iniciar a navegação. Nesses casos, o trajeto a ser seguido é calculado uma única vez, não havendo mais a necessidade do robô realizar este cálculo novamente ao longo de sua locomoção. Finalmente, pode-se dizer que a relevância do controle dos atuadores utilizados pelo robô está relacionada à complexidade mecânica apresentada pelos mesmos. Por exemplo, no caso de robôs cujos atuadores correspondem a rodas, seu controle pode resumir-se simplesmente a acionar ou não tais atuadores. Já no caso de robôs cuja locomoção baseia-se no movimento de pernas mecânicas, o controle dos atuadores pode envolver questões mais complexas, como a manutenção do equilíbrio ou mesmo a definição da amplitude de cada passada (\autoref{fig:leg_robot}).

Dessa forma, em contextos nos quais o robô deve apresentar autonomia completa percebe-se que o problema de percepção e modelagem do ambiente torna-se significativamente relevante para os demais subprocessos envolvidos na navegação. Esta autonomia não se refere somente à total independência de intervenção humana durante a navegação, mas também à não utilização de qualquer espécie de sistema auxiliar externo (como radares ou sistemas de geolocalização) e até mesmo à necessidade de conhecimentos prévios mínimos a respeito do ambiente a ser explorado. De fato, nestas situações a única fonte de informação sobre o ambiente utilizada pelos demais subprocessos corresponde justamente ao modelo gerado pelo próprio robô. Consequentemente, o nível de detalhamento deste modelo é um dos principais responsáveis por definir a complexidade e as limitações envolvidas na realização das demais etapas da navegação. É devido a isso que as técnicas existentes para resolução do problema de navegação autônoma podem ser superficialmente categorizadas de acordo com o método utilizado para o sensoriamento do ambiente e para a representação das informações extraídas. Dentre os principais métodos destacam-se aqueles baseados na utilização de sensores de alcance e aqueles baseados em visão \cite{Siegwart2004}.

\begin{figure}[htb]
	\centering
	\includegraphics[width=0.7\linewidth]{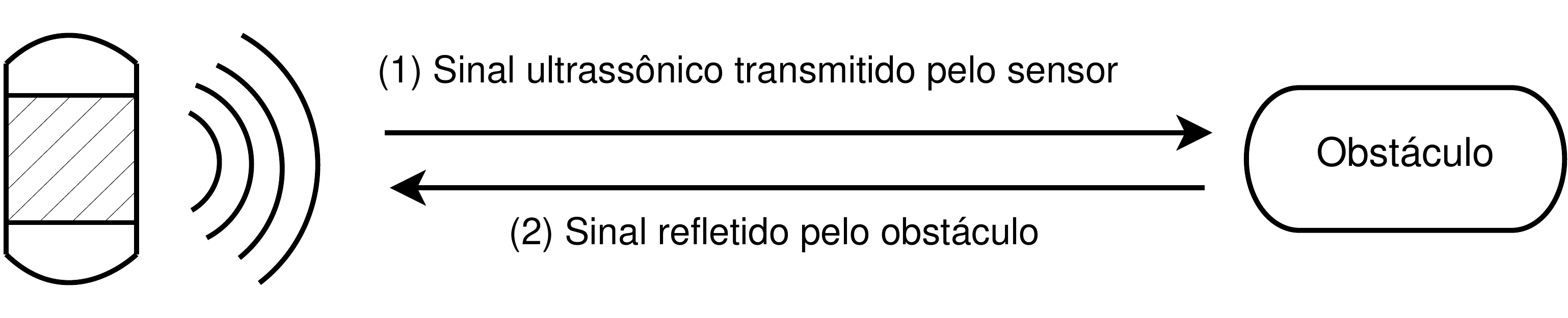}
	\caption{Funcionamento de sensor ultrassônico de alcance.\label{fig:diagrama_sensor_ultrassonico}}
\end{figure}

Assim como qualquer sensor ativo, sensores de alcance funcionam com base na emissão de energia e na medição da resposta gerada pelo ambiente (\autoref{fig:diagrama_sensor_ultrassonico}). Mais especificamente, estes sensores emitem algum tipo de sinal (como uma onda sonora ou um sinal eletromagnético) e medem o tempo que o mesmo leva para ser refletido pelo ambiente. Este tipo de sensor é amplamente utilizado em sistemas robóticos devido à sua simples manipulação e pelo fato das medidas obtidas serem de fácil interpretação. No entanto, por serem ativos há sempre o risco destes sensores alterarem características de interesse do ambiente. Além disso, a utilização de sensores mais simples (como o ultrassônico) provê apenas medidas unidimensionais do ambiente, de modo que o modelo gerado é consideravelmente simplificado. Ainda, sensores que permitem a obtenção de medidas multidimensionais, como \textit{lasers} \cite{JINXUE2011}, apresentam alto custo de aquisição, principalmente quando utilizados em sistemas robóticos de pequeno porte. Por sua vez, o sensoriamento através de visão baseia-se na utilização de câmeras para a obtenção de dados tanto dimensionais quanto relacionados a características puramente visuais do ambiente, como textura, cor, luminosidade, etc. Em muitas situações, estas características apresentam importância fundamental para a correta modelagem do ambiente, sendo que sensores de alcance são incapazes de medi-las. Além disso, o custo de aquisição de câmeras com desempenho razoável é consideravelmente menor que o de \textit{lasers}. Sendo assim, técnicas de navegação autônoma baseadas em visão têm sido amplamente desenvolvidas, sendo atualmente aplicadas em diversos problemas relevantes.

As técnicas de sensoriamento baseadas em visão podem ser divididas em duas classes: as que utilizam visão estereoscópica e as que usam visão monocular \cite{Szeliski2010}. Técnicas que utilizam visão estereoscópica são inspiradas no funcionamento do sistema de visão humano e, de modo simplificado, consistem na captura de múltiplas imagens do ambiente sob diferentes perspectivas para a construção de um modelo tridimensional. Para isso, identificam-se pontos correspondentes contidos nas múltiplas imagens relacionadas a uma mesma cena. Esta etapa geralmente é realizada utilizando-se algoritmos de correlação (por exemplo, o SIFT ou o SURF \cite{Kostavelis2016}). Ao identificá-los, realiza-se o cálculo da profundidade através de algum método de triangulação. Dessa forma, é possível obter um modelo tridimensional do ambiente. Em geral, os modelos gerados através de visão estereoscópica apresentam alta exatidão e baixo índice de incerteza \cite{Lins2016}. No entanto, o custo computacional exigido para a execução dos algoritmos de correlação e de triangulação é elevado. Sendo assim, torna-se inviável a utilização desta abordagem em plataformas robóticas com especificações de \textit{hardware} mais econômicas ou que estejam inseridas em contextos que exijam baixo tempo de processamento.  

\begin{figure}[htb]
	\centering
	\begin{subfigure}[b]{0.4\linewidth}
		\includegraphics[width=\linewidth]{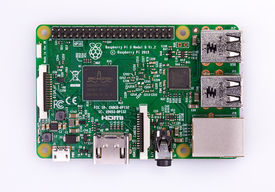}
		\caption{Raspberry Pi 3 Model B \cite{Raspberry}.\label{fig:raspberry_pi_3}}
	\end{subfigure}
	\qquad
	\begin{subfigure}[b]{0.4\linewidth}
		\includegraphics[width=\linewidth]{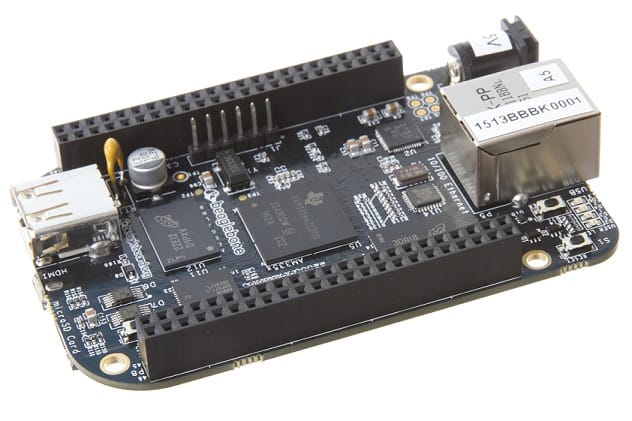}
		\caption{BeagleBone Black \cite{Beagle}.\label{fig:beagle_bone_black}}
	\end{subfigure}
	\caption{Exemplos de plataformas de computação móvel adequadas para o contexto de navegação autônoma.\label{fig:plataformas_recentes}}
\end{figure}

Por sua vez, as técnicas baseadas em visão monocular utilizam algoritmos que, em geral, têm como objetivo identificar padrões de cores, texturas e símbolos \cite{Li2010}, rastrear bordas \cite{Conrad2010} e calcular a movimentação aparente das imagens capturadas \cite{Boroujeni2012}. Estas operações apresentam baixo custo computacional, principalmente quando comparadas às realizadas por técnicas de visão estereoscópica. Devido a isso, abordagens baseadas em visão monocular vêm sendo amplamente aplicadas em sistemas robóticos de navegação autônoma, com ênfase naqueles de pequeno porte, de baixo custo e que devem operar em alta frequência. Ainda assim, observa-se a existência de plataformas móveis nas quais técnicas de visão monocular ainda não foram devidamente investigadas no contexto da navegação autônoma de robôs. Plataformas mais recentes, como as exibidas na \autoref{fig:plataformas_recentes}, apresentam especificações de \textit{hardware} e \textit{software} que as tornam adequadas para serem aplicadas neste tipo de contexto. Sendo assim, percebe-se a oportunidade de explorar o potencial destas plataformas para a implementação de sistemas de navegação autônomos baseados em visão monocular.

\section{Motivação}
A motivação encontrada para o desenvolvimento deste trabalho está relacionada às características promissoras de robôs capazes de se locomover de maneira completamente autônoma, ou seja, sem o auxílio de agentes ou sistemas externos. Devido a tais características, estes instrumentos podem ser aplicados numa grande variedade de situações sem que haja a necessidade de adaptá-los ou mesmo de alterar o ambiente em questão. Além disso, percebe-se que a maioria dos trabalhos voltados para o desenvolvimento deste tipo de ferramenta utiliza plataformas computacionais de alto custo de aquisição, dificultando sua utilização em aplicações cotidianas. Finalmente, o surgimento de novas plataformas de programação móvel capazes de apresentar poder de processamento relativamente alto, eficiente gerenciamento de energia e custo de aquisição não tão elevado representa uma oportunidade para o desenvolvimento de sistemas robóticos menos custosos e com dimensões reduzidas. Dentre tais plataformas, destaca-se a Raspberry Pi \cite{Raspberry}, a qual é amplamente comercializada e possui vasta documentação.

\section{Objetivos}
O objetivo principal deste trabalho é desenvolver um sistema robótico capaz de se locomover de maneira autônoma através de visão monocular e que seja baseado na plataforma de baixo custo Raspberry Pi.

Para tanto, os objetivos específicos deste trabalho são:

\begin{itemize}
	\item Realizar uma revisão bibliográfica sobre as principais técnicas de visão monocular utilizadas para o desenvolvimento de robôs autônomos;
	\item Investigar o desempenho da plataforma Raspberry Pi ao executar alguns dos principais algoritmos utilizados em visão computacional;
	\item Desenvolver a plataforma robótica na qual o sistema de navegação será inserido;
	\item Desenvolver uma estratégia de navegação autônoma baseada em visão monocular capaz de ser executada pela Raspberry Pi;
	\item Comparar o desempenho do sistema final desenvolvido com o dos demais sistemas descritos na literatura.
\end{itemize}

\section{Estrutura do trabalho}

A sequência deste trabalho está organizada da seguinte forma:

\begin{itemize}
	\item \autoref{cap:trabalhos_relacionados} consiste numa revisão bibliográfica acerca das principais estratégias de navegação autônoma;
	\item \autoref{cap:fluxo_optico} apresenta o conceito de fluxo óptico e discute a detecção de obstáculos com base no reconhecimento deste padrão;
	\item \autoref{cap:plataforma_proposta} apresenta os detalhes da plataforma física desenvolvida e descreve o funcionamento do sistema de navegação proposto;
	\item \autoref{cap:experimentos_e_resultados} descreve a metodologia utilizada para avaliar o sistema desenvolvido e discute os resultados obtidos.
\end{itemize}

%% file: Conteudo/TrabalhosRelacionados.tex
\chapter{Trabalhos Relacionados}
\label{cap:trabalhos_relacionados}

Diferentes estratégias de navegação autônoma baseadas em visão monocular podem ser encontradas em publicações recentes. As próximas sessões descrevem aquelas consideradas mais relevantes para este trabalho. Ao fim deste capítulo é realizada uma análise comparativa entre os sistemas de navegação desenvolvidos em cada um dos trabalhos citados.

\section{Detecção de piso por homografia}
A detecção de piso corresponde ao processo de identificação dos pontos de uma imagem que representam a superfície do ambiente capaz de ser percorrida. Esta informação é relevante para a navegação pois permite a identificação dos possíveis objetos contidos na cena. Através deste conhecimento é possível elaborar algoritmos capazes de evitar obstáculos, rastrear alvos, planejar rotas, dentre outros. O sistema proposto por \cite{Conrad2010} realiza a detecção do piso através de uma abordagem probabilística baseada em informações de homografia obtidas a partir da correlação de pontos da imagem. Mais precisamente, este sistema utiliza uma variação do algoritmo EM (\textit{Expectation Maximization}) para realizar o agrupamento não-supervisionado dos pontos, de modo a separar aqueles que representam o piso daqueles que não. Para isso, são apresentados ao EM os parâmetros da matriz de homografia obtida com base nos pontos correlacionados presentes em duas imagens subsequentes. Esta correlação é encontrada a partir da aplicação do algoritmo SIFT. Experimentalmente, o classificador gerado apresentou acurácia equivalente a 99,6\%. No entanto, a plataforma utilizada para os testes tinha como base um processador com frequência igual a 2,0GHz. Este tipo de configuração é inviável para sistemas robóticos de baixo custo e de tamanho reduzido.

\section{Detecção de piso por segmentação de linhas}
A detecção de piso também é tratada por \cite{Li2010}. Este propõe um algoritmo de detecção baseado na classificação de linhas horizontais e verticais. Estas linhas correspondem às bordas presentes nas imagens capturadas do ambiente. Após detectá-las, utiliza-se um classificador SVM para determinar se tais linhas representam de fato uma fronteira entre o piso e as demais estruturas presentes na imagem. Durante os experimentos realizados, este algoritmo apresentou acurácia igual a 89,1\%. Além disso, pelo fato do classificador desenvolvido utilizar diferentes características visuais para ponderar o valor final da classe indicada, o método proposto apresentou alto índice de acerto mesmo em ambientes com grande incidência de luz e reflexão. No entanto, a taxa de acerto foi baixa para ambientes que apresentavam pisos com texturas não-uniformes. Mais ainda, o sistema final desenvolvido foi capaz de processar apenas 5 imagens por segundo.

\section{Segmentação de fluxo óptico}
Por sua vez, \cite{Caldeira2007} sugere a segmentação de objetos com base exclusivamente na sua movimentação relativa ao observador. Esta é conhecida como fluxo óptico. O objetivo desta estratégia é agrupar pontos da imagem que apresentam o mesmo tipo de movimento. Assim, cada agrupamento pode ser associado a estruturas tridimensionais inseridas no ambiente. Após esta segmentação, utiliza-se o fluxo óptico de cada região da imagem para estimar o tempo em que possíveis objetos presentes no ambiente entrarão em contato com o robô. Dessa forma, podem-se determinar as regiões mais seguras para a navegação. Durante experimentos, o sistema desenvolvido foi capaz de realizar todo o processo de segmentação, cálculo do tempo de contato e controle do robô em cerca de 135ms. Este tempo é razoável, considerando-se que todo o processo é realizado pelo próprio robô durante a navegação.

\section{Tempo de contato}
O tempo de contato entre objetos contidos no ambiente e o robô também é calculado por \cite{Sanchez-Garcia2015}. Neste trabalho, inicialmente utiliza-se a segmentação por cor para identificar um possível obstáculo. Ao identificá-lo, calcula-se seu tamanho aparente com base nas bordas da região segmentada. Em seguida, utiliza-se a variação deste tamanho para calcular o coeficiente \textit{Tau Margin}. Este valor é proporcional ao tempo de contato entre o objeto segmentado e o robô. Quando este valor é menor que um determinado limiar deve-se alterar a direção da navegação. Para isso, calcula-se o fluxo óptico em diferentes regiões da imagem com o intuito de verificar o nível de movimentação relativa. Finalmente, o robô é desviado para a região com menor intensidade de fluxo óptico. O método proposto por este trabalho tem como vantagem o baixo custo computacional, uma vez que a maior parte das operações realizadas ocorrem apenas sobre a região segmentada da imagem. No entanto, por realizar a segmentação baseada exclusivamente em cor, este método é incapaz de identificar obstáculos com texturas não-uniformes ou que apresentem baixo contraste em relação ao ambiente.

\section{Classificação de fluxo óptico}
Finalmente, \cite{Shankar2014} propõe um sistema de navegação cuja identificação de obstáculos é realizada exclusivamente com base na detecção de padrões de fluxo óptico. Para isso, utiliza-se um classificador SVM, o qual é treinado antes da navegação a partir de amostras de fluxo óptico extraídas de ambientes semelhantes àquele no qual o robô deverá navegar autonomamente. Após a fase de treinamento, o classificador embarcado no robô é utilizado para indicar a existência ou não de obstáculos em cada imagem capturada. Nos casos em que há a indicação de obstáculos, desvia-se o robô para a direção na qual há menor índice de fluxo óptico. A intensidade do desvio é proporcional à taxa de confiança do classificador. Experimentalmente, o sistema apresentou acurácia equivalente a 88,8\%. Além disso, a taxa de processamento do robô foi de aproximadamente 7 imagens por segundo.

\section{Análise comparativa}

Com base nos dados fornecidos em cada um dos trabalhos citados anteriormente, foi construída a \autoref{tab:analise_trabalhos_relacionados}. Nela, são apresentados os principais parâmetros comparativos entre os sistemas de navegação propostos nestes trabalhos. Vale ressaltar que dentre os parâmetros listados, encontra-se o custo \textit{estimado} de aquisição dos seus componentes físicos. Além disso, como cada trabalho utilizou uma metodologia própria de avaliação, os valores de acurácia e da taxa de processamento não devem ser utilizados diretamente para estabelecerem-se relações de superioridade. Tais parâmetros, portanto, apenas podem ser considerados a título de comparação estimada.

\begin{table}[h]
	\IBGEtab{%
		\caption{Análise comparativa entre os trabalhos relacionados citados neste capítulo.\label{tab:analise_trabalhos_relacionados}}
	}{%
		\begin{tabular}{ccccc}
			\toprule
			\textbf{Estratégia} & \textbf{Acurácia} & \textbf{\begin{tabular}[c]{@{}c@{}}Frames \\ Por Segundo\end{tabular}} & \textbf{Componentes Físicos} & \textbf{\begin{tabular}[c]{@{}c@{}}Custo da \\ Plataforma (US\$)\end{tabular}} \\
			\midrule
			\begin{tabular}[c]{@{}c@{}}Detecção de piso \\ por homografia\end{tabular} & 99,60\% & - & \begin{tabular}[c]{@{}c@{}}HP Pavilion dv6; \\ P3DX Mobile Robot\end{tabular} & 7142,82 \\
			\begin{tabular}[c]{@{}c@{}}Detecção de piso por \\ segmentação de linhas\end{tabular} & 89,10\% & 5 & \begin{tabular}[c]{@{}c@{}}Logitech QuickCam Pro 4000;\\ ActivMedia Pioneer P3AT\end{tabular} & 10612,82 \\
			\begin{tabular}[c]{@{}c@{}}Segmentação \\ de fluxo óptico\end{tabular} & - & 7,41 & 
			\begin{tabular}[c]{@{}c@{}}SONY D30/D31 CCD Camera;\\ ActivMedia Pioneer 2-DX\end{tabular} & 4080,00 \\
			\begin{tabular}[c]{@{}c@{}}Tempo de \\ contato\end{tabular} & - & - & - & - \\
			\begin{tabular}[c]{@{}c@{}}Classificação \\ de fluxo óptico\end{tabular} & 88,80\% & 7 & \begin{tabular}[c]{@{}c@{}}Raspberry Pi Model B;\\ Logitech C170\end{tabular} & 50\\
			\bottomrule
		\end{tabular}
	}{%
		%sem fonte
	}
\end{table}

Ao analisar-se a \autoref{tab:analise_trabalhos_relacionados}, torna-se evidente que as técnicas de navegação baseadas na detecção de piso exigem plataformas físicas robustas e com grande poder de processamento. Este fator contribui diretamente para o alto custo financeiro do sistema de navegação. Em contrapartida, aquele baseado na segmentação de fluxo óptico apresentou a maior frequência de processamento. Além disso, o sistema baseado na classificação deste mesmo sinal apresentou o menor custo financeiro sem que sua acurácia fosse prejudicada. Sendo assim, percebe-se que a utilização do fluxo óptico como característica primária a ser considerada para a navegação demostra ser mais adequada para sistemas que devem apresentar baixo custo financeiro e computacional, como é o caso daquele pretendido por este trabalho.

%% file: Conteudo/FluxoOptico.tex
\chapter{Detecção de Obstáculos por Classificação de Fluxo Óptico}
\label{cap:fluxo_optico}

\begin{figure}[htb]
	\centering
	\begin{subfigure}[b]{0.3\linewidth}
		\includegraphics[width=\linewidth]{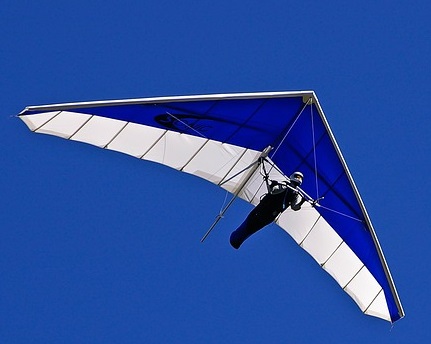}
		\caption{\label{fig:ex_fl1}}
	\end{subfigure}
	~
	\begin{subfigure}[b]{0.3\linewidth}
		\includegraphics[width=\linewidth]{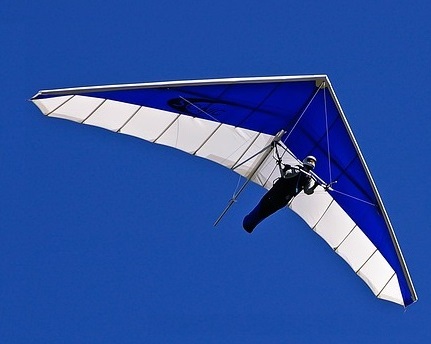}
		\caption{\label{fig:ex_fl2}}
	\end{subfigure}
	~
	\begin{subfigure}[b]{0.3\linewidth}
		\includegraphics[width=\linewidth]{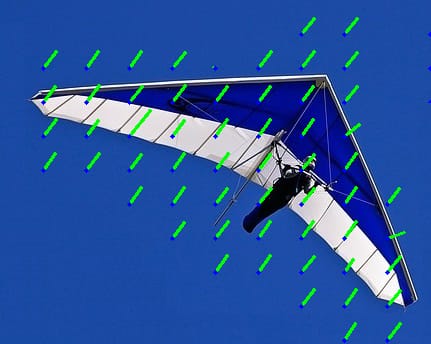}
		\caption{\label{fig:ex_fl3}}
	\end{subfigure}
	\caption{Exemplo de fluxo óptico percebido entre duas imagens em sequência. Os vetores ilustrados em \ref{fig:ex_fl3} representam a movimentação aparente percebida entre as imagens \ref{fig:ex_fl1} e \ref{fig:ex_fl2}.\label{fig:ex_fl}}
\end{figure}

Fluxo óptico corresponde ao conjunto de vetores que descrevem a movimentação aparente dos padrões de brilho de uma imagem. Esta movimentação geralmente está associada ao deslocamento relativo de objetos contidos numa sequência de imagens e a câmera (\autoref{fig:ex_fl}). Consequentemente, através deste fluxo é possível mensurar tal movimento \cite{Horn1981}.

Para o problema de navegação autônoma este tipo de informação é extremamente relevante, uma vez que a mesma está relacionada ao movimento percebido pelo robô acerca dos demais elementos contidos no ambiente. Assim, através do fluxo óptico é possível determinar suas dimensões \cite{Caldeira2007} e posições relativas \cite{Croon2015}. Ainda, tais características podem ser generalizadas e, consequentemente, transformadas em conhecimento útil acerca do ambiente \cite{Shankar2014}. Dentre as aplicações deste conhecimento está a detecção de obstáculos. Esta tarefa corresponde ao processo de identificação de elementos que bloqueiam a passagem do robô \cite{Boroujeni2012}. Devido à sua relevância, entende-se que tal processo representa o requisito mínimo para qualquer sistema de navegação autônomo. Sendo assim, buscou-se investigar sua realização com base no reconhecimento de padrões de fluxo óptico.

As seções a seguir apresentam a definição formal de fluxo óptico e discutem o problema de detecção de obstáculos a partir da classificação deste padrão.

\section{Fluxo Óptico}
Seja a intensidade do brilho de uma imagem num instante $t$ definida pela função $I(\vec{x}, t)$, onde $\vec{x} = (x, y)^T$ corresponde à posição de cada pixel. Considerando-se que em $t+1$ tal intensidade é transladada e mantem-se constante, tem-se que:

\begin{equation}\label{eq:constancia_brilho}
	I(\vec{x}, t) = I(\vec{x} + \vec{u}, t+1),
\end{equation}

\noindent onde $\vec{u} = (u_1, u_2)^T$ representa o deslocamento no plano 2D e corresponde, portanto, ao vetor de fluxo óptico.

A \autoref{eq:constancia_brilho} é geralmente utilizada como ponto de partida para o desenvolvimento de estimadores de fluxo. Muito embora a constância da intensidade de brilho aparente ser uma suposição não realista, na prática consegue-se estimar fluxos com grande precisão \cite{Fleet2005}.

A função de brilho transladada pode ser aproximada por uma série de Taylor de primeira ordem da seguinte forma:

\begin{equation} \label{eq:serie_taylor}
	I(\vec{x} + \vec{u}, t+1) \approx I(\vec{x}, t) + \vec{u}. \nabla I(\vec{x}, t) + I_t(\vec{x}, t),
\end{equation}

\noindent sendo $\nabla I \equiv (I_x, I_y)$ o vetor gradiente espacial discreto de $I(x, y, t)$, $I_t$ a derivada parcial temporal discreta de $I(x, y, t)$ e $\vec{u}. \nabla I(\vec{x}, t)$ o produto escalar.

Combinando-se a \autoref{eq:constancia_brilho} e a \autoref{eq:serie_taylor}, chega-se à seguinte expressão:

\begin{equation} \label{eq:restricao_gradiente}
	\vec{u}. \nabla I(\vec{x}, t) + I_t(\vec{x}, t) \approx 0
\end{equation} 

A \autoref{eq:restricao_gradiente} é conhecida como \textit{equação de restrição de gradiente} \cite{Anton2014}. Apenas a partir desta equação não é possível determinar os valores $u_1$ e $u_2$. É necessário que outras restrições sejam impostas. Na literatura, são listados vários métodos para a resolução dessa equação. Dentre os principais, encontra-se o método de Lucas-Kanade.

\subsection{Algoritmo de Lucas-Kanade}
Para solucionar a \autoref{eq:restricao_gradiente}, o algoritmo de Lucas-Kanade \cite{Lucas1981} propõe que o fluxo $\vec{u}$ seja considerado constante para pontos da imagem pertencentes a uma mesma vizinhança de tamanho $N$. Dessa forma, passa-se a contar com $N$ equações de restrição de gradiente através das quais pode-se determinar $\vec{u}$. Para isso, utiliza-se o \textit{método dos mínimos quadrados} \cite{Filho2016} para estimar-se o valor de $\vec{v}$ que minimize a função de erro:

\begin{equation}
	E(\vec{u}) = \sum_{\vec{x}} g(\vec{x}).[\vec{u}. \nabla I(\vec{x}, t) + I_t(\vec{x}, t)]^2,
\end{equation}

\noindent onde $g(\vec{x})$ é uma função de ponderação utilizada para dar maior relevância a restrições de pontos específicos da vizinhança.

Finalmente, o valor mínimo de $E(\vec{u})$ pode ser encontrado a partir das equações:

\begin{equation}
	\frac{\partial E(u_1, u_2)}{\partial u_1} = \sum_{\vec{x}} g(\vec{x}).[I_x^2 + I_xI_y + I_xI_t]
\end{equation}
\begin{equation}
	\frac{\partial E(u_1, u_2)}{\partial u_2} = \sum_{\vec{x}} g(\vec{x}).[I_y^2 + I_yI_x + I_yI_t].
\end{equation}

Por estimar o fluxo com base no cálculo de derivadas de primeira e segunda ordem referentes a pontos contidos numa mesma vizinhança, o algoritmo de Lucas-Kanade apresenta baixo custo computacional \cite{Barron1994}. No entanto, para que a suposição da igualdade do fluxo nas vizinhanças seja válida, é necessário que a velocidade da movimentação relativa entre a câmera e os objetos contidos na imagem seja aproximadamente constante. Além disso, é comum a utilização de filtros passa-baixa para a suavização da imagem, de modo a reduzirem-se os termos derivativos de ordem mais alta e assim tornar a aproximação da \autoref{eq:serie_taylor} válida.

\section{Reconhecimento de Padrões}
\begin{figure}[htb]
	\centering
	\begin{subfigure}[b]{0.4\linewidth}
		\includegraphics[width=\linewidth]{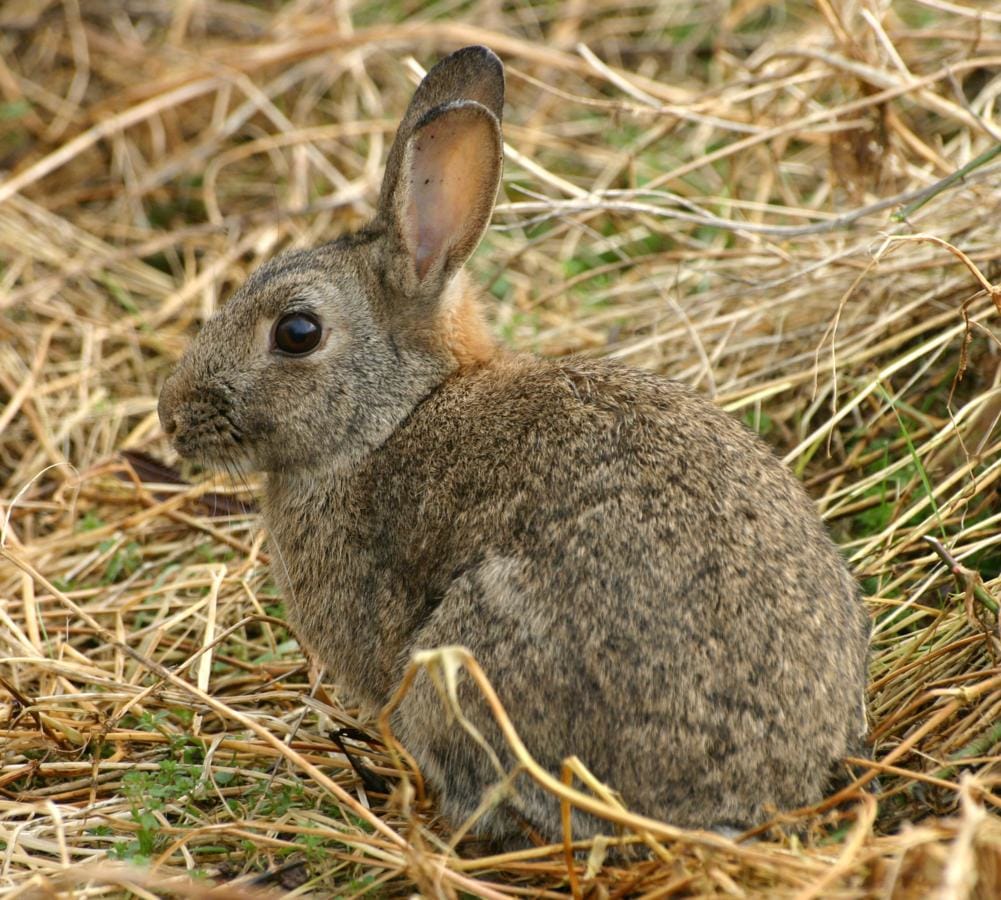}
		\caption{\textit{Oryctolagus cuniculus} (coelho europeu).\label{fig:coelho}}
	\end{subfigure}
	\quad
	\begin{subfigure}[b]{0.4\linewidth}
		\includegraphics[width=\linewidth]{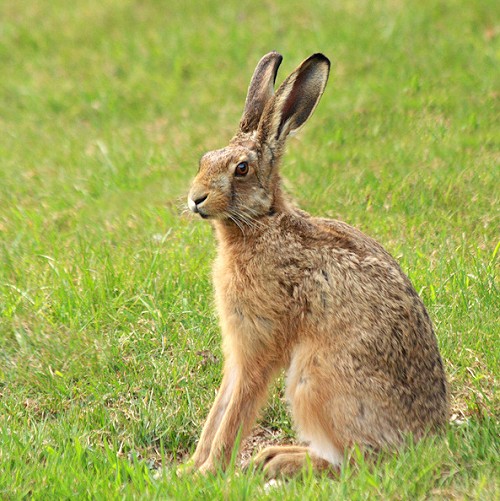}
		\caption{\textit{Lepus europaeus} (lebre europeia).\label{fig:lebre}}
	\end{subfigure}
	\caption{Exemplo de espécies semelhantes cuja principal diferença encontra-se no peso e no comprimento das orelhas. \label{fig:especies_semelhantes}}
\end{figure}

As principais definições a respeito de reconhecimento de padrões podem ser apresentadas a partir da seguinte ilustração: considerem-se as duas espécies ilustradas na \autoref{fig:especies_semelhantes}. Uma delas corresponde ao coelho europeu (\textit{Oryctolagus cuniculus}), enquanto a outra refere-se à lebre europeia (\textit{Lepus europaeus}). Ambas as espécies apresentam características físicas bem semelhantes, como as orelhas compridas e o formato da cabeça, por exemplo. No entanto, ao analisá-las com mais rigor, é possível notar que a lebre possui orelhas mais compridas e aparenta ter um corpo com dimensões maiores. De fato, sabe-se que o coelho europeu adulto pesa em torno de 2,77Kg \cite{Kraus2013} e possui orelhas com comprimento médio igual a 7,8cm \cite{Quesenberry2011}. Por sua vez, a lebre europeia pesa cerca de 3,33Kg e suas orelhas chegam a atingir em média 12,78cm \cite{Angelici2007}. Assim, percebe-se que, num primeiro momento, tais características podem ser úteis para a identificação de integrantes de cada umas destas espécies.

Com base nas informações apresentadas, suponha-se que foram coletadas medidas referentes ao peso e ao comprimento das orelha de alguns animais pertencentes às espécies citadas. Estas medidas são ilustradas na \autoref{tab:coelho} e na \autoref{tab:lebre}.

\begin{table}[h]
	\IBGEtab{%
		\caption{Características de possíveis amostras da espécie \textit{Oryctolagus cuniculus} (coelho europeu).\label{tab:coelho}}
	}{%
		\begin{tabular}{cc}
			\toprule
			\textbf{Peso (Kg)} & \textbf{Comprimento das orelhas (cm)} \\
			\midrule
			2,77 & 7,3 \\
			2,76 & 7,1 \\
			2,77 & 7,4 \\
			2,78 & 7,9 \\
			2,77 & 7,4 \\
			\bottomrule
		\end{tabular}
	}{%
		%sem fonte
	}
\end{table}

\begin{table}[h]
	\IBGEtab{%
		\caption{Características de possíveis amostras da espécie \textit{Lepus europaeus} (lebre europeia).\label{tab:lebre}}
	}{%
		\begin{tabular}{cc}
			\toprule
			\textbf{Peso (Kg)} & \textbf{Comprimento das orelhas (cm)} \\
			\midrule
			3,3 & 12,8 \\
			3,5 & 13,2 \\
			3,6 & 13,4 \\
			3,2 & 12,9 \\
			3,3 & 13,0 \\
			\bottomrule
		\end{tabular}
	}{%
		%sem fonte
	}
\end{table}

\begin{figure}[htb]
	\centerline{\includegraphics[width=0.9\linewidth]{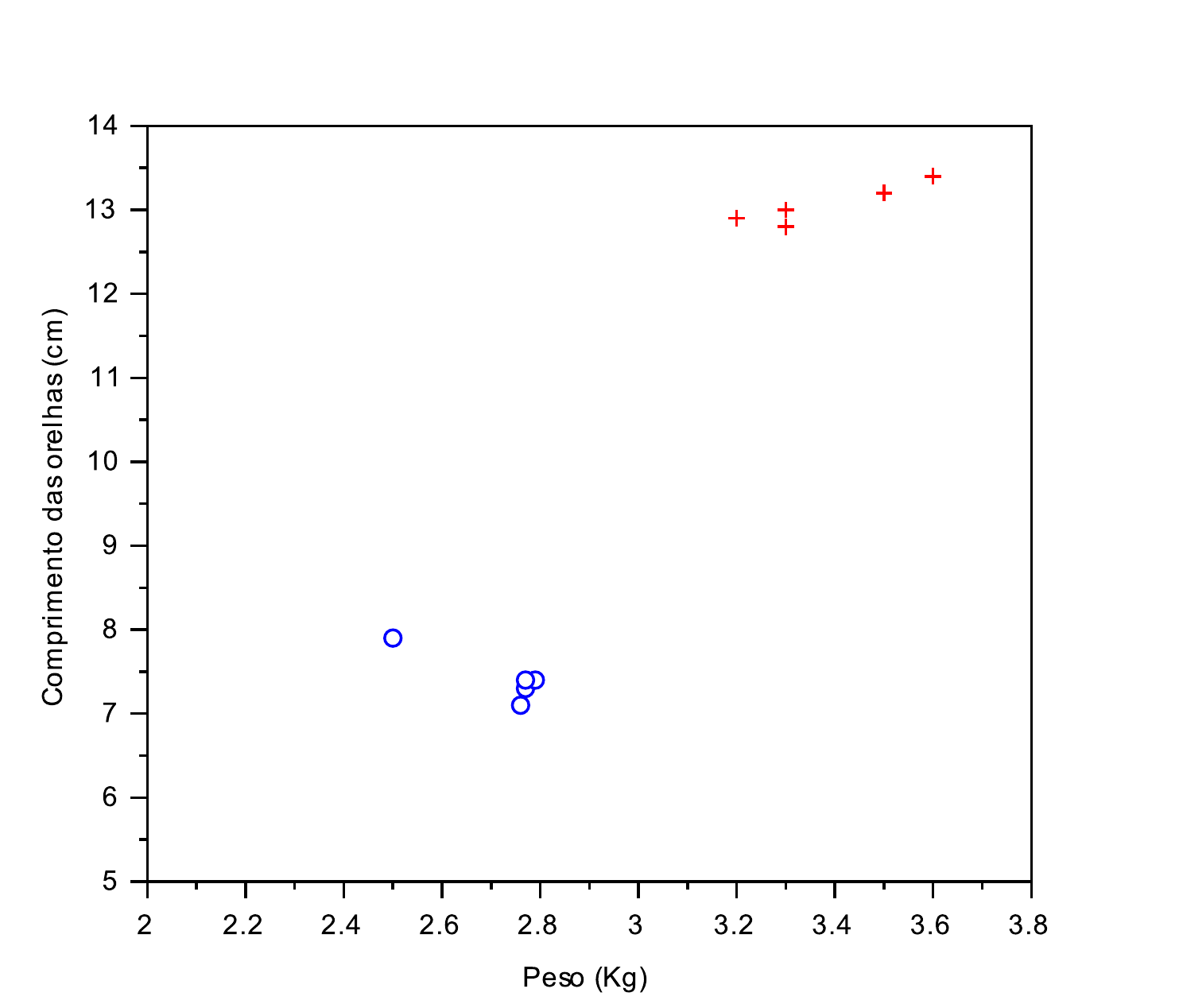}}
	\caption{Plotagem das amostras de cada espécie com base nas suas características. O eixo das abcissas corresponde ao peso e o das ordenadas representa o comprimento das orelhas. Os símbolos em azul correspondem aos coelhos europeus, enquanto que aqueles em vermelho referem-se às lebres europeias.\label{fig:distribuicao_coelhos_apriori}}
\end{figure}

Para melhor visualizar a diferença existente entre ambas as espécies, cada amostra foi plotada num gráfico bidimensional, utilizando-se como coordenadas os valores das suas respectivas características aferidas (\autoref{fig:distribuicao_coelhos_apriori}).

\begin{figure}[htb]
	\centerline{\includegraphics[width=0.9\linewidth]{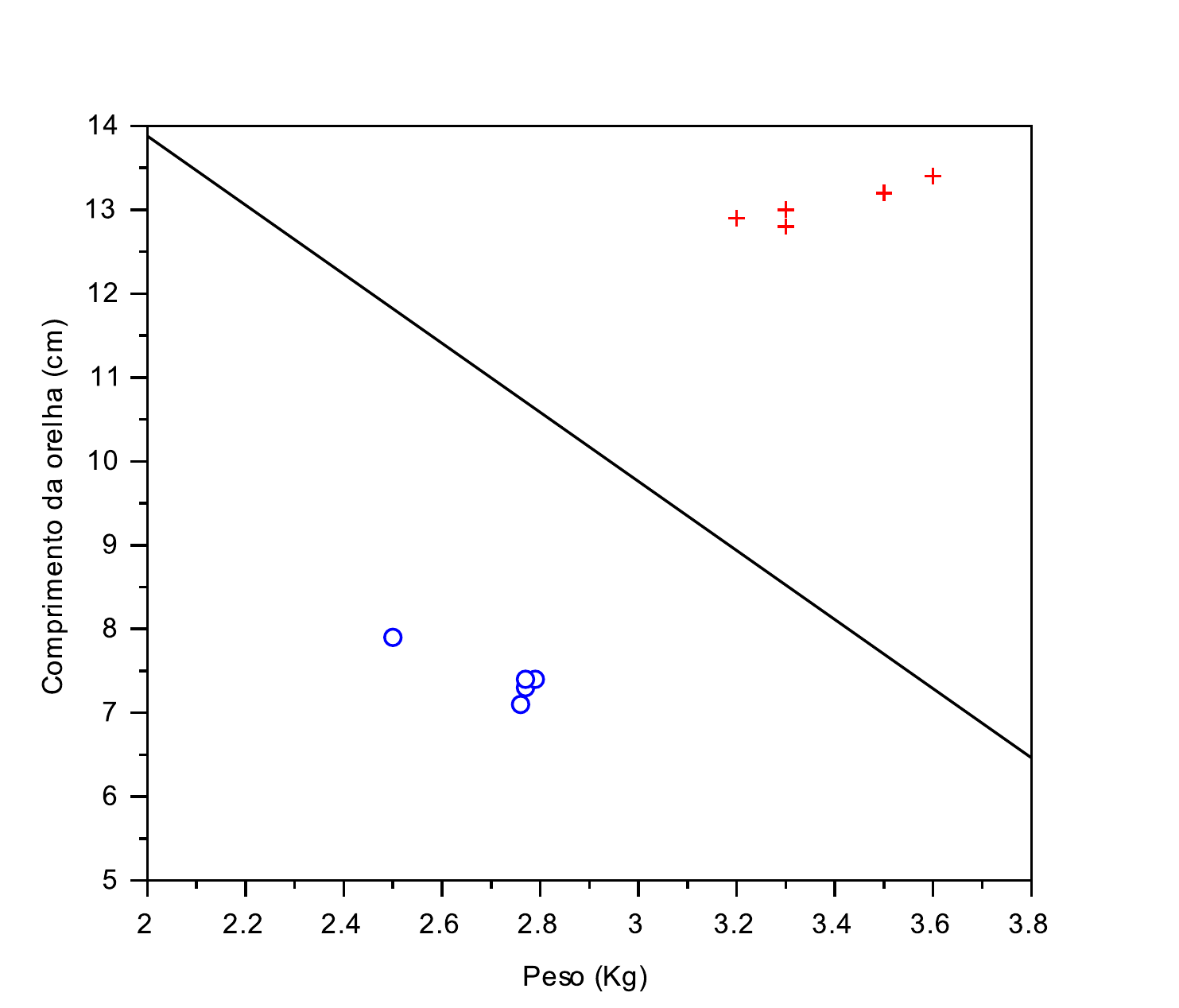}}
	\caption{Ilustração de reta que marca uma possível fronteira entre as regiões definidas por cada espécie.\label{fig:separacao_distribuicao_coelhos}}
\end{figure}

Ao analisar-se a \autoref{fig:distribuicao_coelhos_apriori}, nota-se que os pontos refentes aos membros da mesma espécie localizam-se em regiões bem definidas, de forma a ser possível identificá-los com certa facilidade. De fato, a separação entre tais regiões pode ser sinalizada através de uma reta, como ilustrado na \autoref{fig:separacao_distribuicao_coelhos}.

\begin{figure}[htb]
	\centerline{\includegraphics[width=0.9\linewidth]{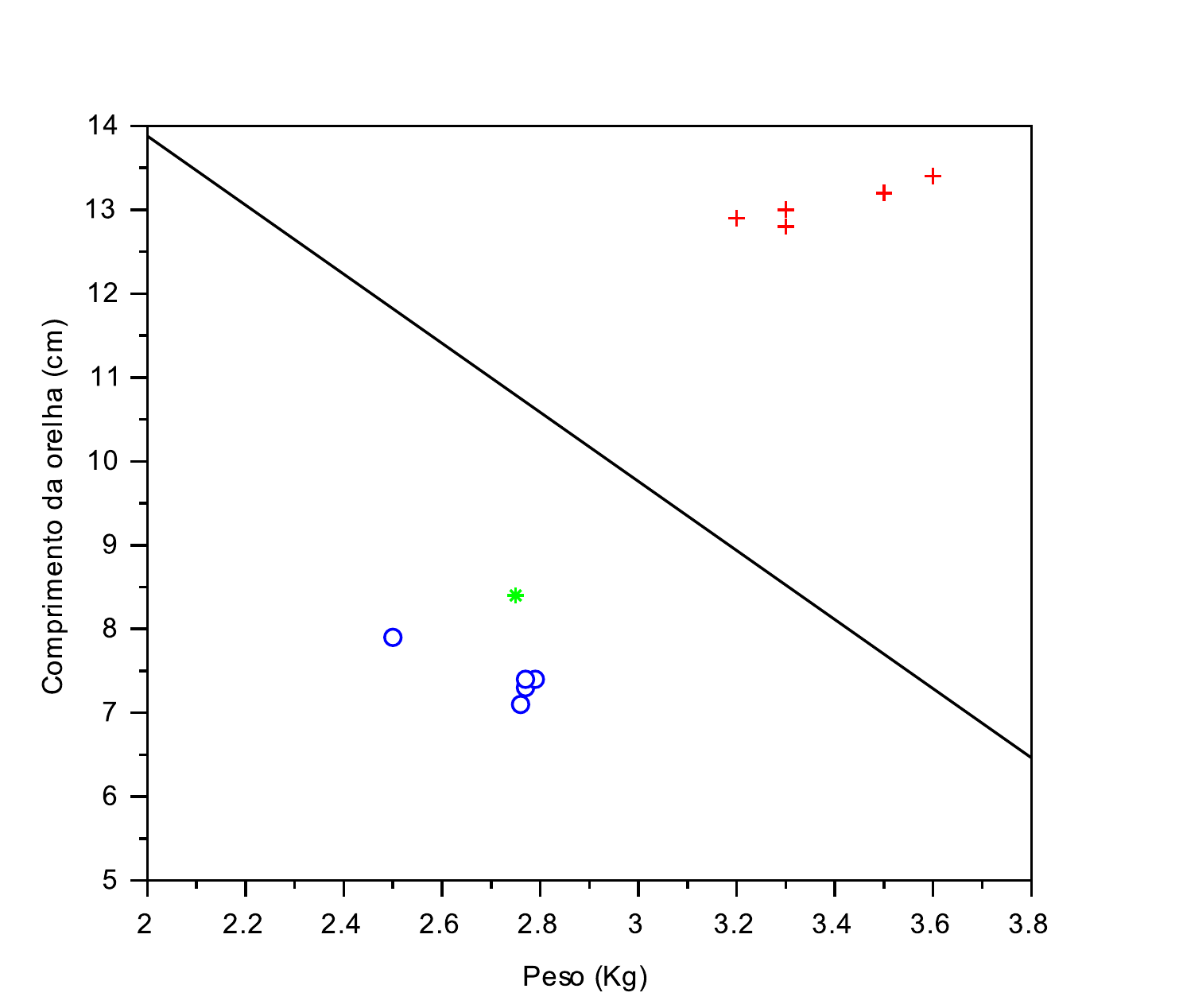}}
	\caption{Ilustração de elemento cuja espécie ainda é desconhecida (símbolo em verde). Como o mesmo encontra-se abaixo da reta separadora, pode-se estimar que tal elemento pertença à espécie \textit{Oryctolagus cuniculus}.\label{fig:distribuicao_coelhos_ponto_desconhecido}}
\end{figure}

Suponha-se, finalmente, que o peso e o comprimento das orelhas de um novo elemento tenham sido aferidos, sendo que sua espécie é desconhecida. O ponto correspondente a este elemento é plotado sobre o gráfico que contém as espécies já conhecidas. O resultado obtido é exibido na \autoref{fig:distribuicao_coelhos_ponto_desconhecido}. Ao analisar o novo ponto, percebe-se que o mesmo encontra-se próximo à região na qual estão concentrados os coelhos europeus. Sendo assim, torna-se razoável assumir que o elemento desconhecido provavelmente corresponde a um membro desta espécie.

Todo o problema ilustrado até aqui, desde o processo de identificação das características que diferenciam os membros de cada espécie até a suposição da classe à qual pertence um elemento desconhecido, corresponde a uma tarefa de reconhecimento de padrões. Um padrão nada mais é do que um objeto definido por um conjunto de características \cite{Theodoridis2003}. Na situação apresentada até então, os coelhos e as lebres correspondem a objetos. Estes são definidos por inúmeras características, tais como: tamanho, peso, cor, velocidade, dentre outras. O reconhecimento de um padrão corresponde, portanto, à tentativa de agruparem-se objetos que tenham características semelhantes. No caso do problema ilustrado, a característica desejada corresponde à espécie. Como visto, uma das possíveis estratégias de reconhecimento consiste em avaliarem-se as semelhanças existentes entre determinados objetos cujas classes já sejam conhecidas. No exemplo apresentado, estas semelhanças corresponderam ao peso do animal e ao comprimento de suas orelhas. Em seguida, desenvolve-se um modelo que represente a regra de separação entre os elementos das diferentes classes. A reta ilustrada na \autoref{fig:separacao_distribuicao_coelhos} representa a regra desenvolvida para o problema tratado. Finalmente, através deste modelo pode-se inferir a classe ao qual pertence um determinado objeto cujo rótulo ainda é desconhecido. Foi o caso do novo elemento projetado na \autoref{fig:distribuicao_coelhos_ponto_desconhecido}. Como seu respectivo ponto ficou abaixo da reta divisória, pôde-se inferir que o mesmo corresponde a um coelho europeu. Em outras palavras, foi possível identificá-lo.

O processo de reconhecimento ilustrado pode ser estendido para vários outros problemas, nos quais seja preciso levar em consideração mais do que apenas duas características de cada objeto. De forma geral, um padrão pode ser representado como um vetor de características da seguinte maneira:

\begin{equation}\label{eq:vetor_caracteristicas}
\vec{x} = (X_1, X_2, ..., X_N)
\end{equation}

Na \autoref{eq:vetor_caracteristicas}, $N$ corresponde à quantidade de características consideradas e $X_N$ ao valor da $N$-ésima característica. Como pode-se perceber, para problemas onde os vetores de características apresentam mais do que três dimensões, torna-se inviável determinar graficamente um modelo capaz de separá-los corretamente. Além disso, pode haver situações nas quais a fronteira de classificação não seja linear, de modo que a regra de separação a ser estabelecida seja mais complexa do que um reta. Finalmente, em muitas situações os padrões avaliados podem ser separados por mais de um único modelo. Nestes casos, surge a necessidade de escolher-se aquele capaz de otimizar algum parâmetro específico exigido pelo contexto em questão, como a distância entre as regiões separadas, a simplicidade da fronteira, o ajuste do modelo aos padrões conhecidos, dentre outros.

A depender do problema tratado, pode-se preferir separar os padrões analisados com base em mais de duas regiões. Independente da quantidade escolhida, é importante ressaltar que os modelos gerados durante o processo de reconhecimento de padrões são apenas capazes de \textit{estimar} a classe a qual pertencem os objetos. Tal estimativa, portanto, não garante que o reconhecimento seja correto. Sendo assim, durante a fase de criação do modelo (também conhecida como treinamento) busca-se estudar padrões que apresentem características marcantes sobre seu conjunto. Estes objetos geralmente fazem parte de uma base de treino utilizada exclusivamente para a criação do modelo. Uma vez criado, sua qualidade é aferida com base na apresentação de novos objetos, os quais fazem parte de uma base de validação. Para isso, seus rótulos são inicialmente escondidos, de modo que o modelo treinado seja utilizado para predizê-los. Ao final, as classes atribuídas através do modelo são comparadas aos rótulos reais dos padrões apresentados. A partir daí, é possível inferir a qualidade do modelo gerado. Ao analisá-la, pode-se decidir realizar novos ajustes no modelo e testá-lo novamente. Todo este processo é realizado até que a qualidade aferida seja satisfatória, de modo a minimizar a quantidade de padrões mal classificados.

Finalmente, é importante ressaltar a importância da escolha das características a serem consideradas para a criação do modelo. Em geral, os padrões estudados em problemas de reconhecimento apresentam várias características. É o caso das duas espécies tratadas na ilustração trazida anteriormente. Neste problema, diversos parâmetros poderiam ter sido considerados para caracterizá-las. No entanto, percebeu-se que o peso e o comprimento das orelhas por si só já apresentavam alta correlação com o tipo da espécie. Sendo assim, decidiu-se considerar apenas tais atributos, de modo que o modelo gerado pudesse separar todos os objetos conhecidos da forma mais simples possível. Em problemas mais complexos, como os relacionados à visão computacional, é natural que a quantidade de características a serem consideradas seja razoavelmente maior. No entanto, deve-se ter como objetivo selecionar a menor quantidade de características possível, de modo a tornar trivial a utilização do modelo para fins de classificação \cite{Duda2001}.

\subsection{Detecção de Obstáculos}
\begin{figure}[htb]
	\centering
	\begin{subfigure}[b]{0.4\linewidth}
		\includegraphics[width=\linewidth]{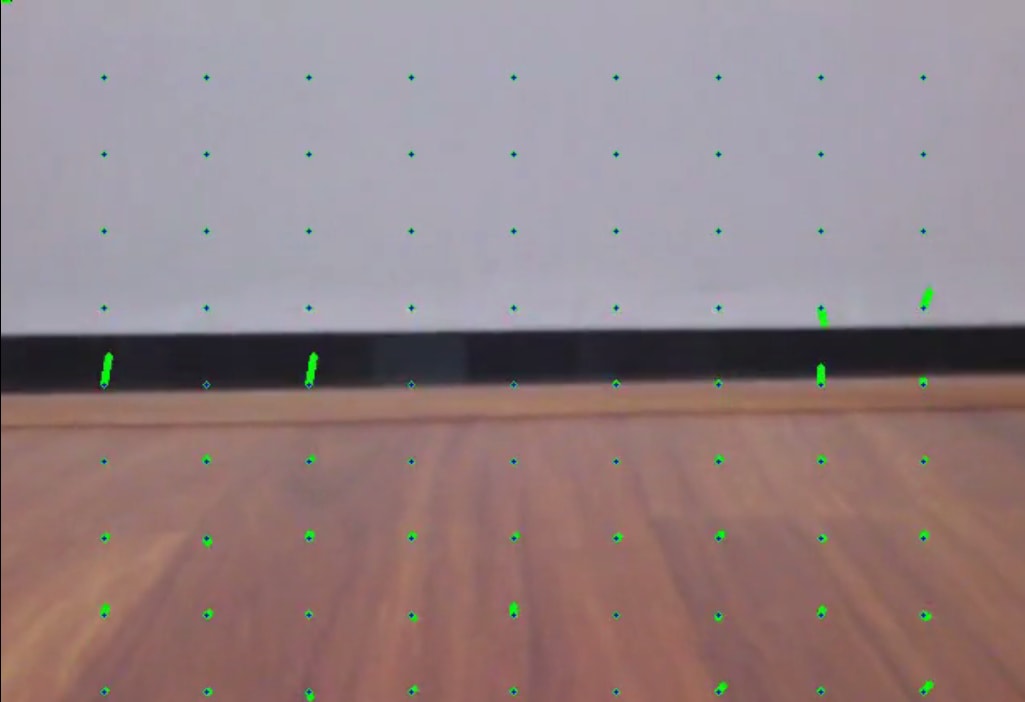}
		\caption{\label{fig:movimentacao_sem_obstaculo}}
	\end{subfigure}
	\begin{subfigure}[b]{0.4\linewidth}
		\includegraphics[width=\linewidth]{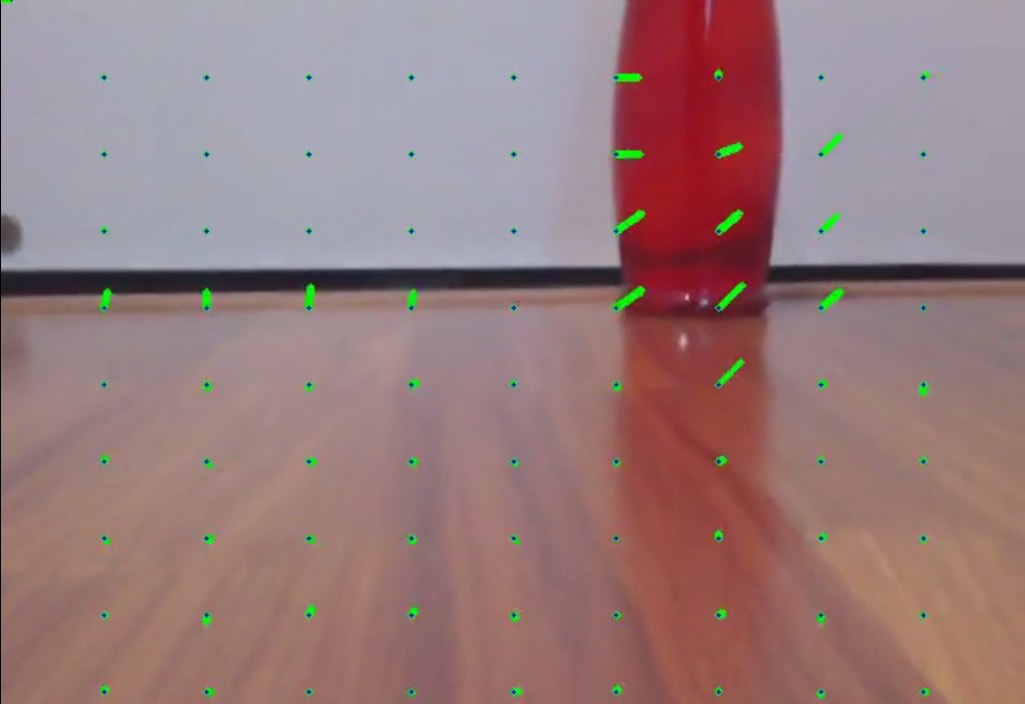}
		\caption{\label{fig:movimentacao_com_obstaculo}}
	\end{subfigure}
	\caption{Exemplo da movimentação característica de um obstáculo. \ref{fig:movimentacao_sem_obstaculo} apresenta o fluxo óptico de uma cena sem obstáculos próximos. \ref{fig:movimentacao_com_obstaculo} ilustra a movimentação contrastante de um obstáculo em relação aos demais elementos do ambiente. \label{fig:movimento_caracteristico_de_obstaculo}}
\end{figure}

A detecção de obstáculos corresponde ao processo de identificação de objetos que bloqueiam a passagem percorrida por um agente. Dentre as principais características destes objetos está a sua acentuada movimentação relativa. Tal intensidade de aproximação é contrastante àquelas apresentadas pelos demais elementos da cena, uma vez que estes encontram-se mais distantes do agente (\autoref{fig:movimento_caracteristico_de_obstaculo}).

Por sua vez, sabe-se que o movimento relativo entre os elementos de uma imagem e a câmera podem ser estimados a partir do cálculo do fluxo óptico. Sendo assim, com base nos conhecimentos acerca de reconhecimento de padrões, tem-se que a partir de amostras de fluxo óptico extraídas de diversas cenas sem e com obstáculos é possível gerar um modelo capaz de separar tais padrões. Em outras palavras, pode-se treinar um classificador com base em objetos já rotulados e aplicá-lo para indicar se uma nova cena apresenta ou não um obstáculo. Sendo assim, a detecção de obstáculos com base no reconhecimento de fluxo óptico pode ser realizada através de uma máquina classificadora, cujo aprendizado supervisionado é alcançado com base nas seguintes definições:

\begin{itemize}
	\item $\vec{x_i}=(F_1,...F_n)$ corresponde ao vetor de caraterísticas associado à $i$-ésima imagem da sequência considerada;
	\item $F_n$ corresponde ao par $(v_1, v_2)$, sendo $v_1$ e $v_2$ a norma e a fase do vetor de fluxo óptico associado ao $n$-ésimo ponto da imagem, respectivamente;
	\item Cada $\vec{x_i}$ está associado a um valor $y_i\in\{-1,+1\}$, onde os rótulos $-1$ e $+1$ indicam a ausência e a presença de obstáculos na imagem $i$, respectivamente.
\end{itemize}

\begin{figure}[htb]
	\centerline{\includegraphics[width=0.7\linewidth]{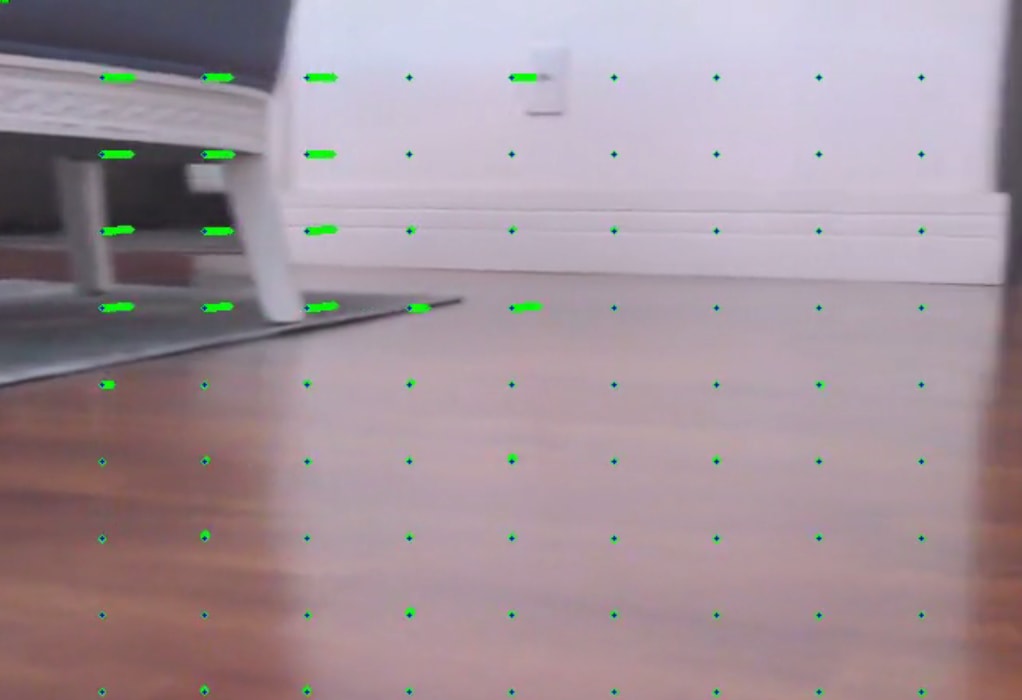}}
	\caption{Ilustração de fluxo óptico extraído de uma curva. Como visto, a intensidade do movimento é alta, porém o obstáculo desviado à esquerda não representa mais um bloqueio.\label{fig:movimentacao_curva}}
\end{figure}

Como visto, o padrão analisado corresponde ao movimento percebido entre uma sequência de imagens. Seu vetor de características é composto pelo fluxo óptico estimado em $n$ pontos distintos. Além disso, as informações utilizadas para representar o fluxo correspondem à sua norma e à sua fase. A primeira é justificada pelo contraste da intensidade de movimentação já discutido anteriormente. Por sua vez, a fase é considerada devido à possibilidade de prováveis obstáculos se moverem para fora da trajetória do agente, deixando de representar um bloqueio. É o caso de uma curva acentuada, na qual a intensidade do fluxo é alta, porém o objeto à frente do percurso já está sendo desviado pelo agente (\autoref{fig:movimentacao_curva}). Finalmente, para o problema em questão é suficiente assumirem-se a existência de apenas duas classes: a da movimentação associada a obstáculos e a do movimento referente a elementos não bloqueantes.

%% file: Conteudo/PlataformaProposta.tex
\chapter{Sistema Proposto}
\label{cap:plataforma_proposta}

Com base nas discussões realizadas nos capítulos anteriores, foi desenvolvida uma plataforma robótica cujo sistema de navegação baseia-se na detecção de obstáculos a partir da classificação de fluxo óptico. Tal sistema foi implementado sobre a placa de baixo custo Raspberry Pi. As seções a seguir descrevem o \textit{hardware} e o \textit{software} do robô desenvolvido.

\section{\textit{Hardware}}
Os componentes que formam o \textit{hardware} da plataforma projetada correspondem a uma placa Raspberry Pi, um chassi de sustentação contendo dois atuadores, uma fonte de alimentação e dois sensores: uma câmera monoscópica e um sensor ultrassônico. Os detalhes acerca das especificações técnicas e da utilização de cada um destes componentes são descritos a seguir, juntamente com a apresentação do \textit{hardware} final obtido e dos custos arcados.

\subsection{Raspberry Pi}
\begin{figure}[h]
	\centering
	\includegraphics[width=0.7\linewidth]{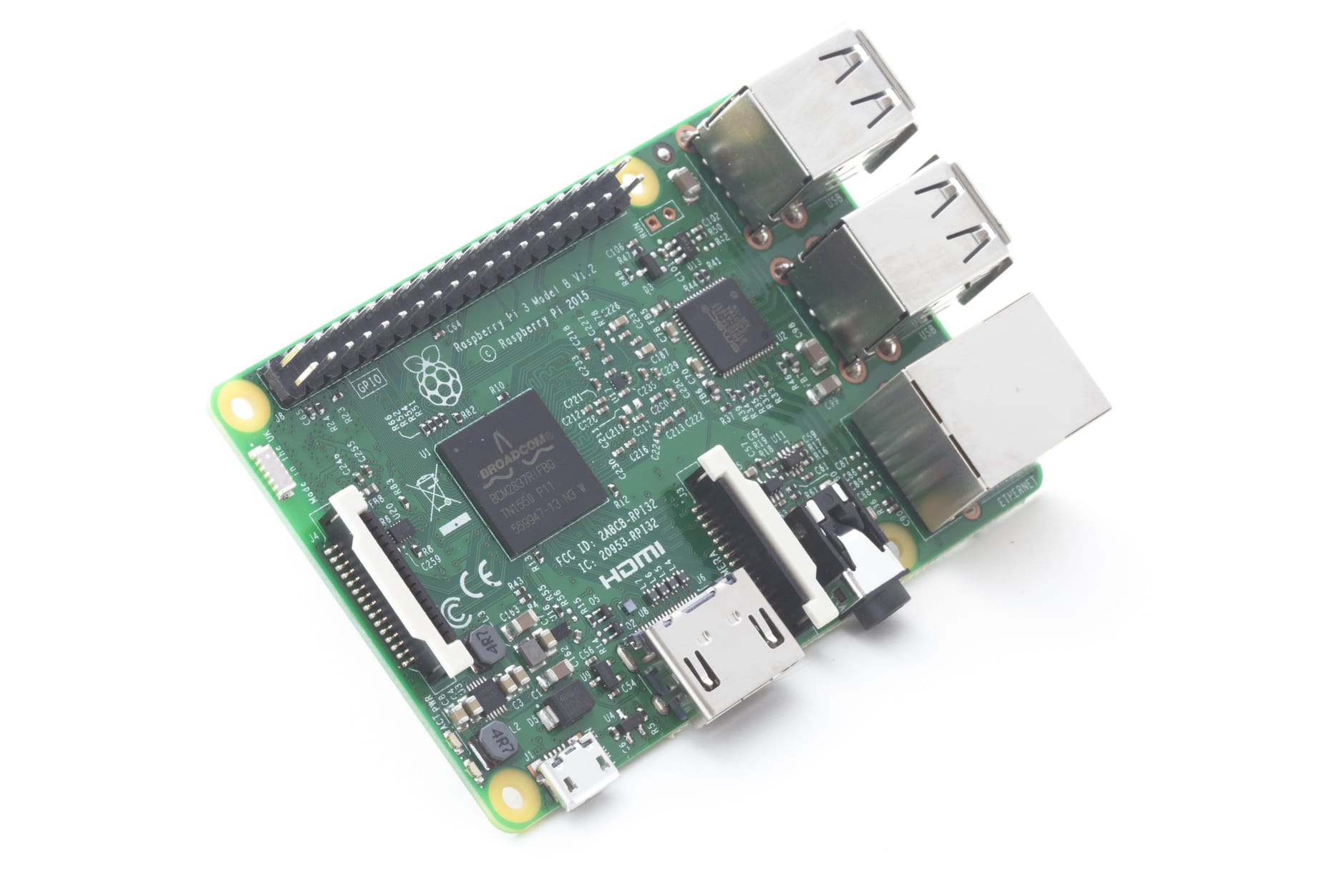}
	\caption{Raspberry Pi 3 Model B\label{rasp3b}}
\end{figure}

A plataforma Raspberry Pi \cite{Raspberry} teve seu primeiro modelo lançado mundialmente no ano de 2012 e desde então tornou-se amplamente difundida entre educadores, instituições de ensino e entusiastas da tecnologia. Com o objetivo de estimular o interesse no aprendizado de Computação, este pequeno computador tem chamado a atenção da comunidade acadêmica devido às suas dimensões extremamente reduzidas, ao seu baixo custo de aquisição e à sua alta performance. Baseada na arquitetura ARM \cite{ARM}, a Raspberry Pi comporta, além de seu processador, uma unidade de processamento gráfico (GPU), memória RAM, entradas USB e interface Ethernet integrados numa única placa, cujas dimensões podem ser comparadas às de um cartão de crédito. A \autoref{rasp3b} apresenta a Raspberry Pi 3 Model B utilizada neste projeto. Suas principais especificações são \cite{Raspberry3}:

\begin{itemize}
	\item CPU ARMv8 quad-core 64-bit 1,2GHz;
	\item RAM 1GB;
	\item GPU VideoCore IV 3D;
	\item 1 leitor de cartão Micro SD;
	\item 4 portas USB;
	\item LAN Wireless 802.11n;
	\item Bluetooth 4.1;
	\item 1 porta Ethernet;
	\item 1 porta HDMI.
\end{itemize}

\begin{figure}[htb]
	\centering
	\begin{subfigure}[b]{0.7\linewidth}
		\includegraphics[width=\linewidth]{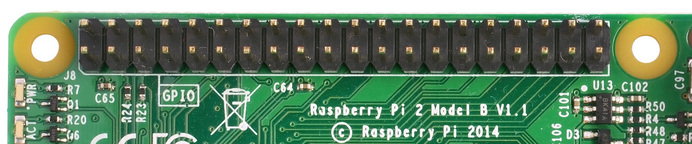}
		\caption{Ilustração da GPIO.\label{fig:gpio_pinos}}
	\end{subfigure}
	\begin{subfigure}[b]{0.7\linewidth}
		\includegraphics[width=\linewidth]{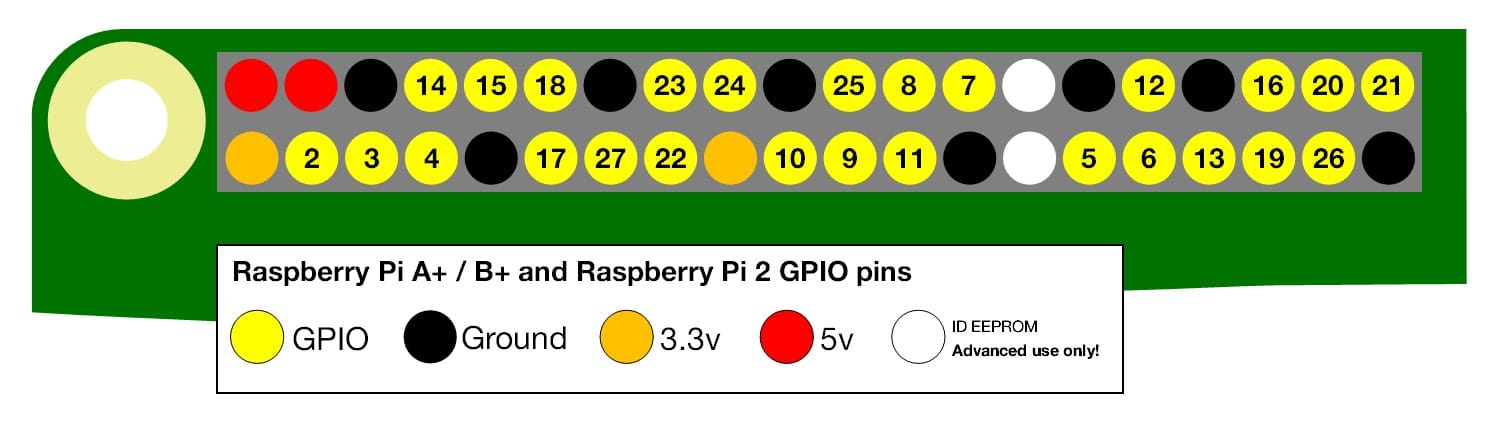}
		\caption{Numeração da GPIO.\label{fig:gpio_numeros}}
	\end{subfigure}
	\caption{Pinos de propósito geral de uma Raspberry Pi 2. Estes são os mesmos presentes na Raspberry Pi 3 Model B.\cite{RaspberryGPIO}.\label{fig:gpio}}
\end{figure}

Como visto nas especificações citadas, a Raspberry Pi 3 Model B conta com um adaptador \textit{Wireless} embutido, tornando-a ainda mais adequada para aplicações em sistemas autônomos, como o proposto neste trabalho. Além disso, um dos grandes diferenciais da Raspberry Pi é o seu conjunto de pinos de propósito geral, conhecido como GPIO (\autoref{fig:gpio}). Estes pinos permitem que a Raspberry Pi se comunique através de sinais digitais com  dispositivos externos de entrada e saída, como sensores, sinalizadores e motores \cite{Corporation2012}. A tensão máxima de entrada e saída suportada por cada pino corresponde a 3,3V. Já a corrente máxima suportada por cada um é igual a 16mA, sendo que todo o conjunto é capaz de fornecer ao máximo 50mA simultaneamente. Por fim, é importante ressaltar que apesar de todos esses recursos e especificações, a Raspberry Pi exige uma fonte de alimentação de apenas 5,1V capaz de fornecer cerca de 2,5A, sendo seu consumo médio inferior a 5,1W \cite{RaspberryPower}.

\subsection{Chassi}
\begin{figure}[htb]
	\centering
	\includegraphics[width=0.5\linewidth]{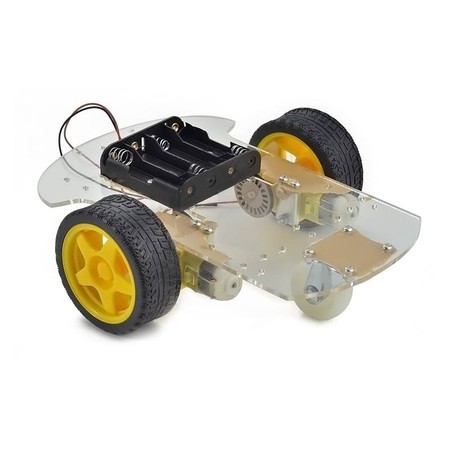}
	\caption{Chassi utilizado como base para a plataforma robótica\label{fig:chassi}}
\end{figure}

A \autoref{fig:chassi} ilustra o chassi utilizado como base para a construção da plataforma robótica, o qual contém os atuadores a serem controlados pelo sistema de navegação. Tal escolha deve-se ao fácil controle das rodas do chassi e à variedade de superfícies em que o mesmo pode ser utilizado sem que haja perda de estabilidade. Além disso, considerou-se adequada a escolha de um chassi que evidenciasse a simplicidade, o baixo custo e a eficiência do sistema final a ser desenvolvido.

Este chassi conta com as seguintes especificações:

\begin{itemize}
	\item Dimensões: $21,3x15,3 cm$;
	\item Diâmetro das rodas: $6,5 cm$;
	\item Espessura das rodas: $2,7 cm$.
\end{itemize}

\begin{figure}[htb]
	\centering
	\includegraphics[width=0.5\linewidth]{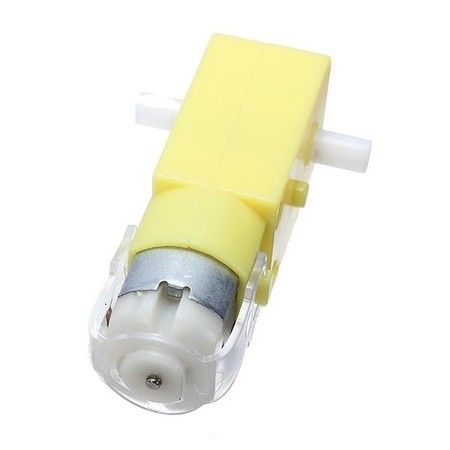}
	\caption{Modelo do motor DC utilizado para movimentar as rodas do chassi.\label{fig:motor_dc}}
\end{figure}

\begin{figure}[htb]
	\centering
	\begin{subfigure}[b]{0.5\linewidth}
		\includegraphics[width=\linewidth]{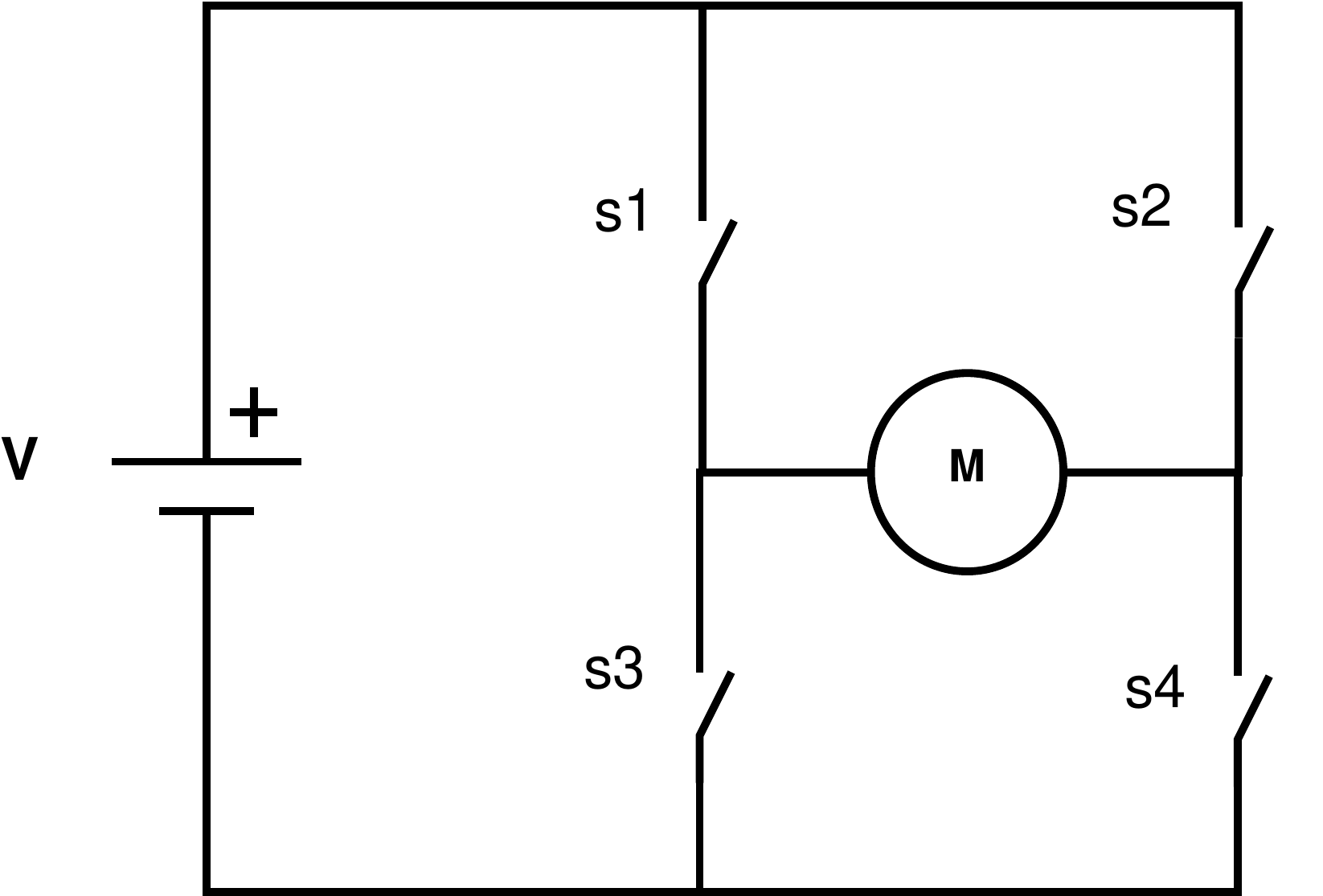}
		\caption{\label{fig:ponteh_esquema_eletrico}}
	\end{subfigure}
	\begin{subfigure}[b]{0.7\linewidth}
		\includegraphics[width=\linewidth]{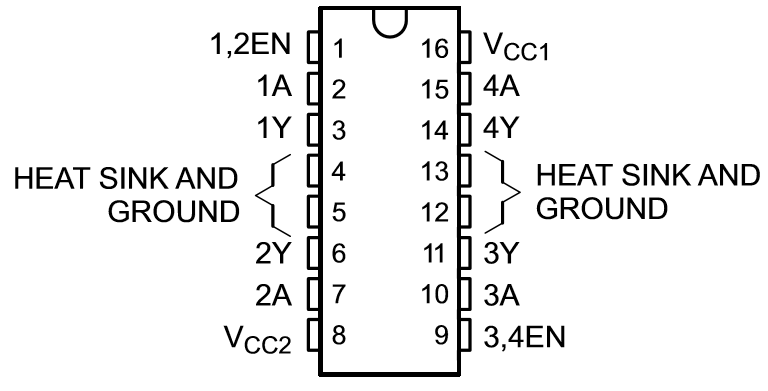}
		\caption{\label{fig:l293d_pinagem}}
	\end{subfigure}
	\caption{Ponte H. \ref{fig:ponteh_esquema_eletrico} apresenta o esquema elétrico deste circuito. \ref{fig:l293d_pinagem} exibe a pinagem do circuito integrado L293D.\label{fig:ponteh_ilustracoes}}
\end{figure}

Para movimentar as rodas do chassi, foram utilizados dois motores DC com caixa de redução e eixo duplo (\autoref{fig:motor_dc}). A relação entre a tensão, a corrente e a velocidade de rotação destes motores são apresentadas na \autoref{tab:motores-espec}. Para acionar estes motores foi utilizado o circuito integrado L293D \cite{TII2016}, o qual possui internamente dois circuitos denominados \textit{Ponte H}. Como visto na \autoref{fig:ponteh_esquema_eletrico}, neste tipo de circuito o acionamento do motor \textit{M} é realizado através de quatro chaves, as quais definem o sentido com que a corrente gerada pela fonte \textit{V} atravessará o motor. Assim, a utilização deste tipo de circuito provê duas vantagens: é possível alterar facilmente o sentido de rotação do motor e o seu controle pode ser realizado de maneira isolada do fornecimento de alimentação. Esta última vantagem é especialmente importante quando deseja-se utilizar um controlador incapaz de fornecer a alimentação necessária para o funcionamento do motor, como é o caso da Raspberry Pi utilizada neste projeto.

\begin{table}[htb]
	\IBGEtab{%
		\caption{Especificações dos motores DC\label{tab:motores-espec}}
	}{%
		\begin{tabular}{ccc}
			\toprule
			\textbf{Tensão} & \textbf{Corrente} & \textbf{RPM} \\
			\midrule
			3V     & 100mA    & 120                 \\
			5V     & 100mA    & 208                 \\
			6V     & 120mA    & 240                 \\
			\bottomrule                
		\end{tabular}
	}{%
		%sem fonte
	}
\end{table}

Na \autoref{fig:l293d_pinagem} é possível observar a pinagem do L293D. Os pinos \textit{1Y} e \textit{2Y} correspondem aos terminais da primeira ponte \textit{H} nos quais seria ligado o motor \textit{M} da \autoref{fig:ponteh_esquema_eletrico}. Da mesma forma, os pinos \textit{3Y} e \textit{4Y} correspondem aos mesmos terminais presentes na segunda ponte \textit{H}. O valor de saída destes pinos é determinado pelos valores de entrada nos respectivos pinos \textit{A} de mesma numeração. Já os pinos \textit{1,2E} e \textit{3,4E} habilitam os pinos \textit{1Y}, \textit{2Y} e \textit{3Y}, \textit{4Y}, respectivamente. Por fim, a tensão destinada ao pino \textit{VCC1} é utilizada para alimentar os circuitos lógicos deste circuito integrado, enquanto que a tesão aplicada no pino \textit{VCC2} é destinada exclusivamente aos pinos de saída \textit{Y}. Sendo assim, um dos motores instalados no chassi foi ligado aos pinos \textit{Y1} e \textit{Y2} do LD293D, enquanto que o outro motor foi ligado aos pinos \textit{Y3} e \textit{Y4}. Já os demais pinos de controle (do tipo \textit{E} e \textit{A}) foram conectados aos pinos da interface GPIO da Raspberry Pi. Além disso, esta também foi utilizada para fornecer uma tensão de 5V para o circuito lógico do LD293D. 

\subsection{Alimentação}
\begin{figure}[htb]
	\centering
	\includegraphics[width=0.5\linewidth]{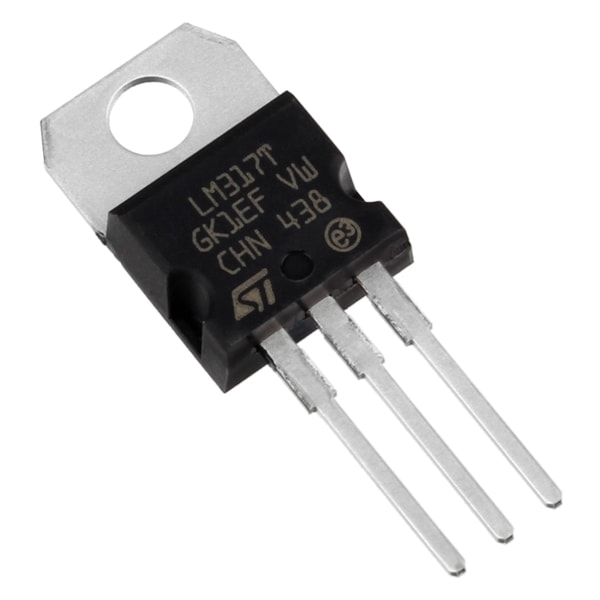}
	\caption{Regulador de tensão LM317 utilizado inicialmente para estabilizar a alimentação da Raspberry Pi. No entanto, o mesmo demonstrou-se inviável, já que sua corrente máxima fornecida sem que haja necessidade de um dissipador de calor é menor do que a exigia pela placa \cite{Incorporated2002}.\label{fig:lm317}}
\end{figure}

Para alimentar os dois motores foram utilizadas inicialmente quatro pilhas de 1,5V. Já para alimentar a Raspberry Pi de maneira estável, tentou-se utilizar algumas pilhas recarregáveis de 1,25V em conjunto com o regulador linear de tensão LM317 (\autoref{fig:lm317}). Esta escolha foi feita em virtude da fácil manipulação e do baixo custo de aquisição apresentados por este regulador. No entanto, por ser linear, o LM317 exige uma tensão de entrada 3V maior que aquela desejada na saída, o que ocasiona seu rápido aquecimento devido à alta dissipação de potência. Além disso, a corrente de saída recomendada pelo fabricante deste regulador para que não haja a necessidade de conectá-lo a um dissipador de calor corresponde a somente 1,5A, inferior à exigida pela Raspberry Pi. Sendo assim, ao tentar alimentar esta placa utilizando o LM317, percebeu-se que este passou a esquentar muito rapidamente, de modo que seu circuito de proteção limitou a tensão de saída para um valor muito inferior ao exigido pela Raspberry Pi.

\begin{figure}[htb]
	\centering
	\includegraphics[width=0.5\linewidth]{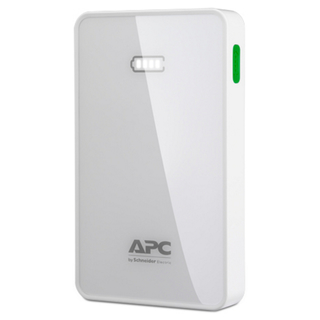}
	\caption{Carregador portátil Power Pack APC utilizado para alimentar a Raspberry Pi e os dois motores DC do chassi \cite{PowerAPC}.\label{fig:powerbank}}
\end{figure}

Devido às dimensões reduzidas do chassi, tornava-se inviável a instalação de um dissipador de calor. Pelo mesmo motivo desconsiderou-se a utilização de um regulador do tipo \textit{Step-Down} \cite{Incorporated2016}. Finalmente, decidiu-se utilizar o carregador portátil Power Pack APC (\autoref{fig:powerbank}). Este é capaz de fornecer 5000mAH já regulados, possuindo duas saídas de 5V, sendo que uma delas fornece no máximo 1A, enquanto que a outra é capaz de prover até 2,4A. Sendo assim, foi possível alimentar tanto a Raspberry Pi quanto os dois motores DC, conseguindo-se também reduzir o peso total sobre o chassi e aumentar o espaço disponível, já que foram retiradas as pilhas anteriormente utilizadas para alimentar os motores.

\subsection{Sensores}
\begin{figure}[htb]
	\centering
	\includegraphics[width=0.5\linewidth]{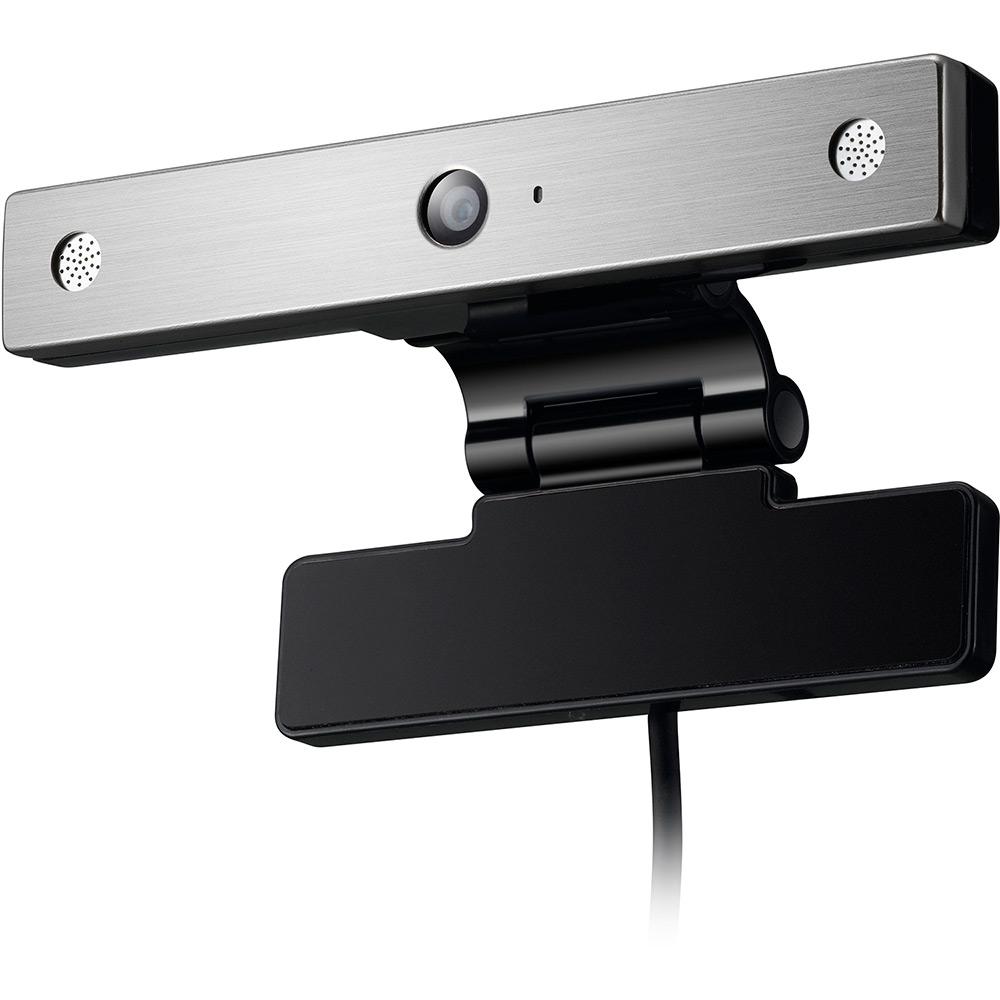}
	\caption{Câmera LG AN-VC500 instalada no chassi \cite{LG2016}.\label{fig:camera}}
\end{figure}

Com o maior espaço disponível sobre o chassi, foi possível instalar os dois sensores a serem utilizados pelo sistema de navegação. O primeiro deles foi a câmera LG AN-VC500 \autoref{fig:camera}, a qual possui as seguintes especificações:

\begin{itemize}
	\item Resolução máxima: $1920x1080$ \textit{pixels};
	\item Taxa de captura: 30 FPS;
	\item Formatos de saída: H.264 e YUY2;
	\item Interface: USB 2.0.
\end{itemize}

A mesma foi posicionada na parte frontal do chassi, de modo que sua inclinação em relação ao solo fosse de aproximadamente $0^\circ$. Em seguida, foi preciso apenas conectá-la a uma das portas USB da Raspberry Pi.

\begin{figure}[htb]
	\centering
	\includegraphics[width=0.5\linewidth]{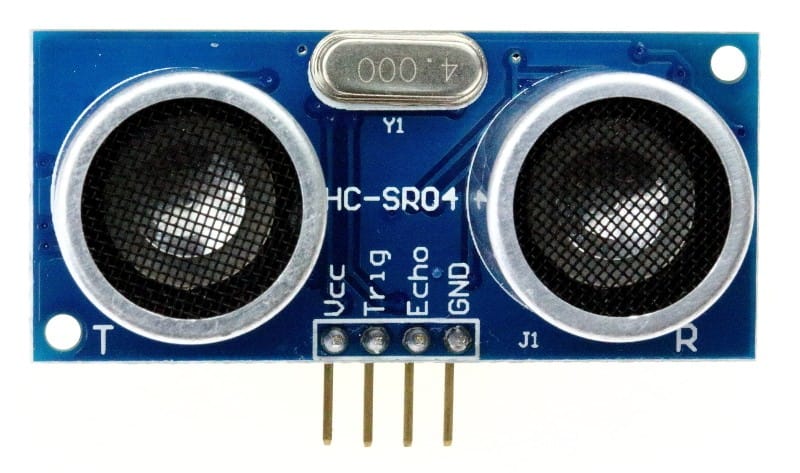}
	\caption{Sensor ultrassônico HC-SR04 instalado no chassi para utilização durante o treinamento do sistema. \cite{Indoware2013}.\label{fig:sensor_ultrassonico}}
\end{figure}

\begin{figure}[htb]
	\centering
	\includegraphics[width=0.5\linewidth]{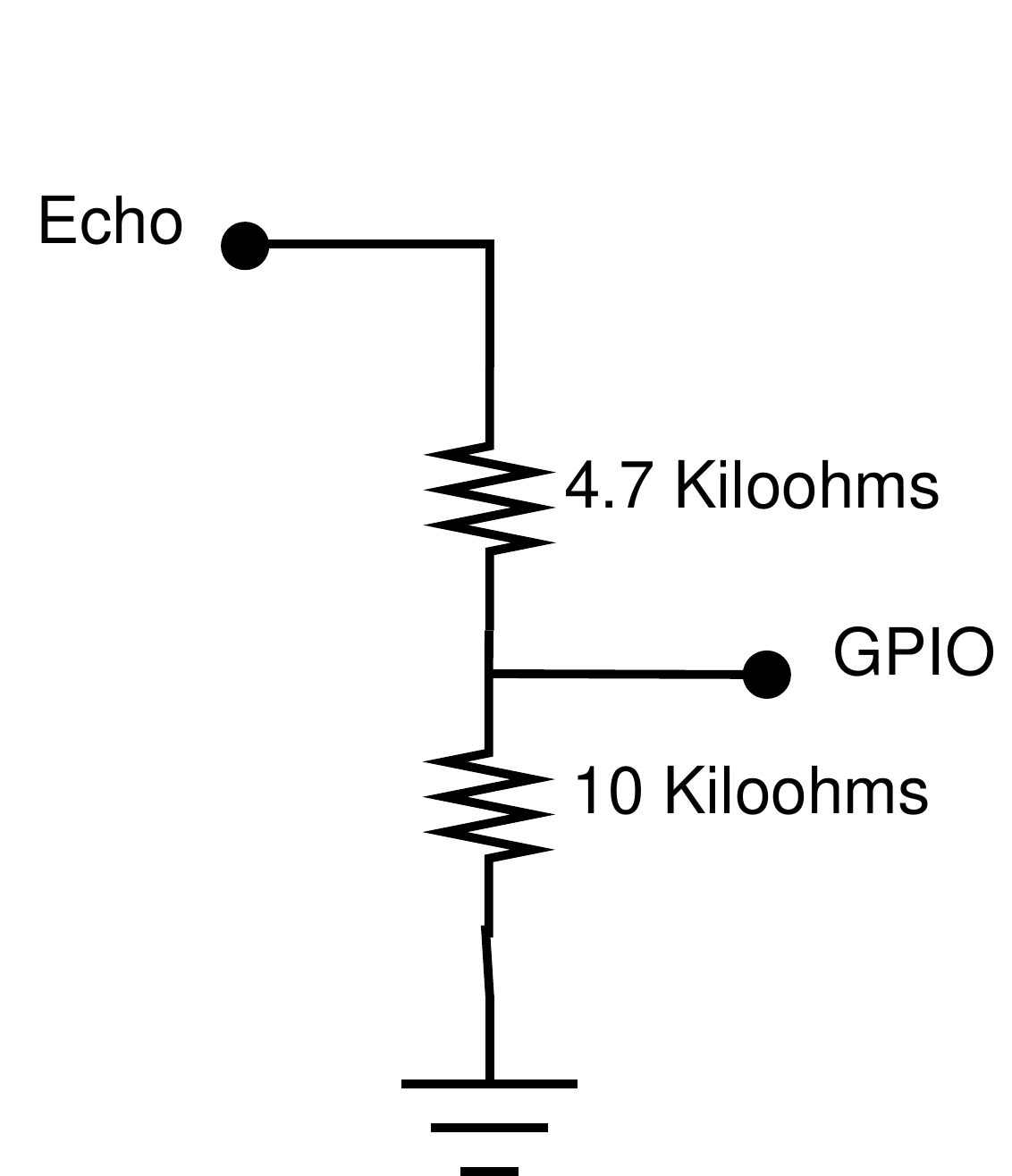}
	\caption{Divisor de tensão utilizado para conectar a saída \textit{Echo} do sensor ultrassônico à interface GPIO da Raspberry Pi.\label{fig:divisor_tensao}}
\end{figure}

Em seguida, foi instalado o sensor ultrassônico HC-SR04 (\autoref{fig:sensor_ultrassonico}). O mesmo funciona com base na emissão de uma onda sonora, cuja frequência é igual a 40KHz, e na medição do tempo que a mesma leva para ser refletida pelo ambiente. Este sensor foi instalado na plataforma com a intenção de ser utilizado durante o processo de treinamento do sistema de navegação. Através deste instrumento é possível inferir a distância do robô a obstáculos rígidos, lisos e reflexivos. Seu alcance varia entre 2cm a 4m, com precisão de 3mm, segundo o fabricante. Além disso, o mesmo requer 5V de alimentação. Seus 4 pinos são nomeados como: \textit{VCC}, \textit{GRD}, \textit{Trigger} e \textit{Echo}. Os dois primeiros correspondem aos pinos de alimentação e de referência (terra), respectivamente. Já o terceiro corresponde ao pino de entrada utilizado para efetuar o disparo da onda sonora. Para isso, deve-se fornecer um pulso cuja duração seja igual a 10$\mu$s. Por fim, o último corresponde ao pino de saída, o qual retorna uma onda cujo tempo em nível lógico alto é proporcional ao tempo de ida e volta do sinal emitido. Os três primeiros pinos foram conectados diretamente à interface GPIO da Raspberry Pi. O mesmo não foi possível para o pino \textit{Echo}. Como sua tensão de saída corresponde a 5V, foi preciso utilizar um circuito divisor de tensão para conectá-lo à Raspberry Pi. O mesmo foi formado por resistores de $4,7K\Omega$ e $10K\Omega$ (\autoref{fig:divisor_tensao}).

\subsection{\textit{Hardware} Final}

\begin{figure}[htb]
	\centering
	\begin{subfigure}[b]{0.7\linewidth}
		\includegraphics[width=\linewidth]{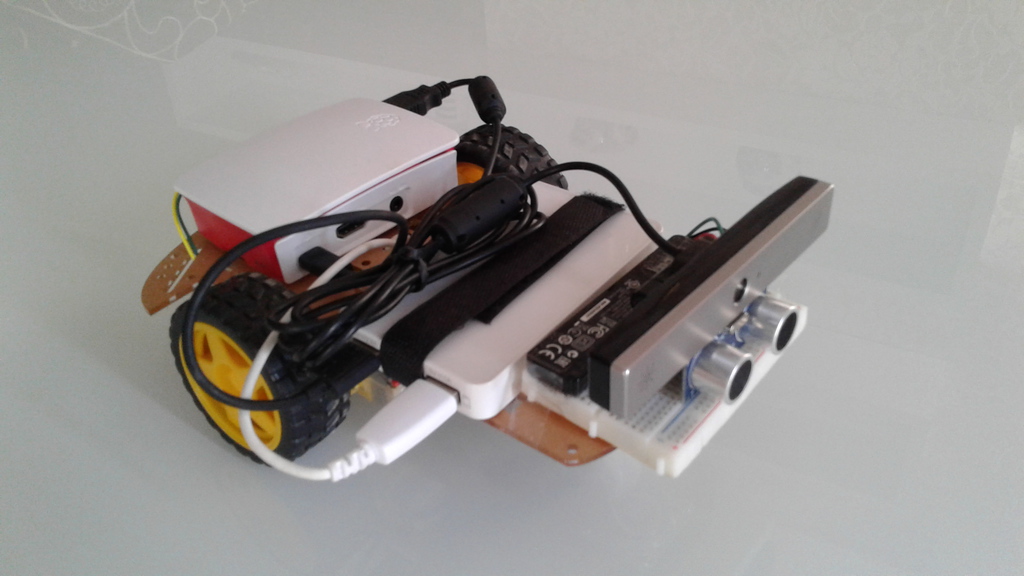}
		\caption{\label{fig:plataforma_perspectiva}}
	\end{subfigure}
	\qquad
	\begin{subfigure}[b]{0.7\linewidth}
		\includegraphics[width=\linewidth]{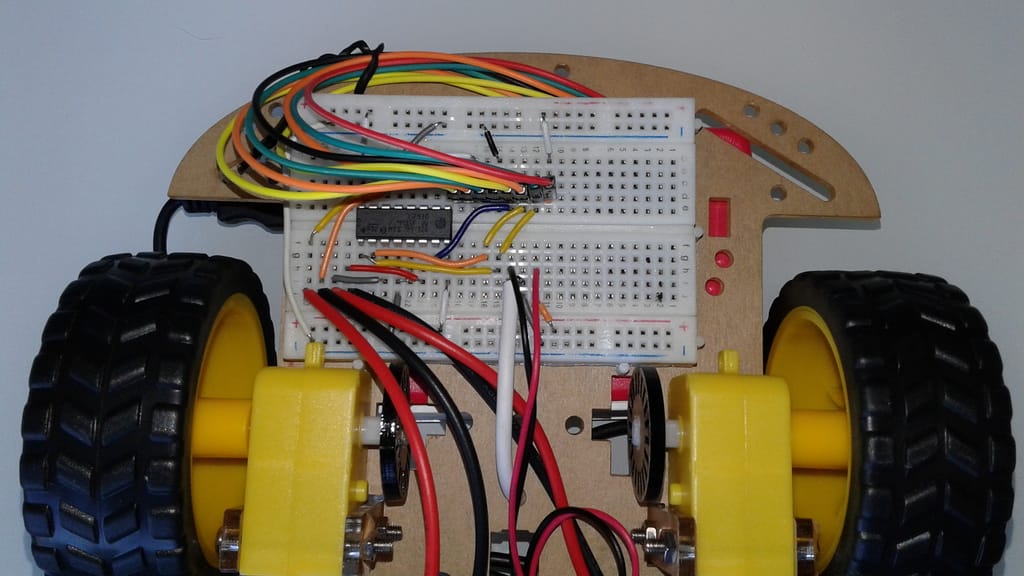}
		\caption{\label{fig:plataforma_inferior}}
	\end{subfigure}
	\caption{Plataforma robótica construída. \ref{fig:plataforma_perspectiva} apresenta a visão em perspectiva da plataforma completa. \ref{fig:plataforma_inferior} exibe a parte inferior da plataforma, com foco sobre a ligação dos motores ao L293D.\label{plataforma_construida}}
\end{figure}

A \autoref{plataforma_construida} apresenta a plataforma final construída. Já o seu custo financeiro é descrito na \autoref{tab:custo_plataforma}.

\begin{table}[H]
	\IBGEtab{%
		\caption{Custo da plataforma desenvolvida.\label{tab:custo_plataforma}}
	}{%
		\begin{tabular}{cc}
			\toprule
			\textbf{Componente} & \textbf{Custo (US\$)} \\
			\midrule
			Raspberry Pi 3 & 30,00 \\
			Câmera LG AN-VC500 & 89,99 \\
			Chassi & 21,55 \\
			Sensor HC-SR04 & 2,00 \\
			L293D & 1,90 \\
			Power Bank APC M5WH & 25,38 \\
			\midrule
			\textbf{Total} & 170,82 \\
			\bottomrule
		\end{tabular}
	}{%
		%sem fonte
	}
\end{table}

Ao compararem-se os custos apresentados na \autoref{tab:custo_plataforma} com os dos trabalhos relacionados exibidos na \autoref{tab:analise_trabalhos_relacionados}, nota-se que a plataforma desenvolvida é mais barata que a da maioria dos sistemas citados. Além disso, é importante ressaltar que o trabalho que propôs a plataforma mais barata da \autoref{tab:analise_trabalhos_relacionados} não inclui os custos do chassi e da fonte de alimentação utilizados.

\section{\textit{Software}}
 
\begin{figure}[htb]
	\centerline{\includegraphics[width=\linewidth]{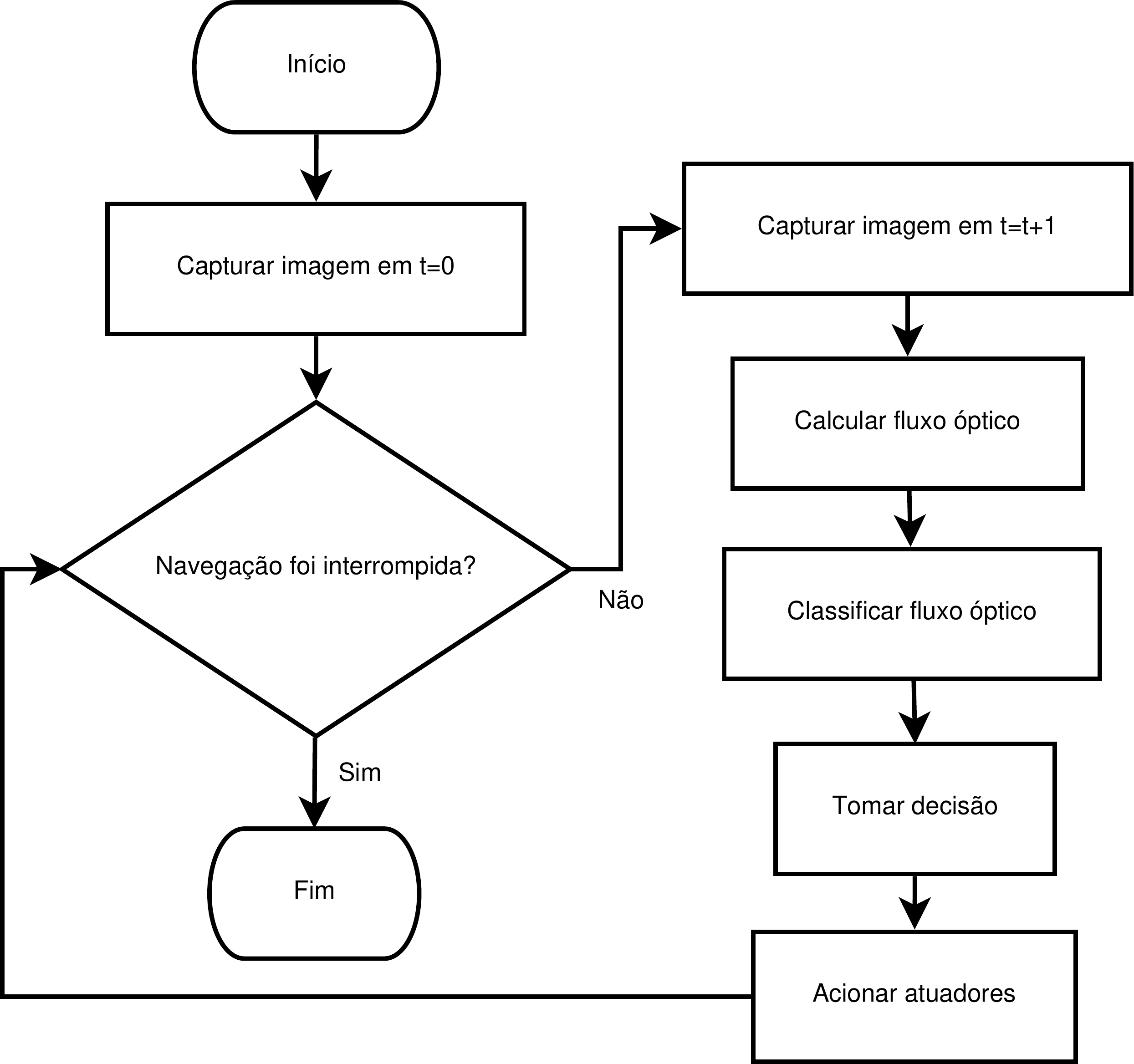}}
	\caption{Fluxograma do sistema de navegação proposto.\label{fig:fluxograma_sistema}}
\end{figure}

O sistema de navegação proposto foi desenvolvido sobre o sistema operacional Raspbian \cite{Raspbian} com base nas bibliotecas de visão computacional OpenCV \cite{OpenCV3-2} e de aprendizado de máquina Scikit-Learn \cite{scikit-learn} (\autoref{ap:config_rasp}). Seu ciclo de funcionamento é resumidamente descrito pelo fluxograma da \autoref{fig:fluxograma_sistema}. Como visto em tal diagrama, o sistema inicialmente captura uma imagem de referência e inicia o ciclo de navegação. Neste, captura-se uma nova imagem do ambiente e calcula-se o fluxo óptico entre as duas imagens registradas. O fluxo estimado é apresentado ao classificador, o qual indicará se há algum obstáculo no caminho percorrido. Com base nesta indicação, toma-se uma decisão quanto à atualização da direção do robô, de modo a realizar-se alguma manobra de desvio ou manter-se a trajetória do robô em linha reta.

Os detalhes acerca do funcionamento e da implementação de cada etapa da navegação são melhor descritos nas subseções apresentadas a seguir.

\subsection{Cálculo do fluxo óptico}
\begin{figure}[htb]
	\centerline{\includegraphics[width=0.7\linewidth]{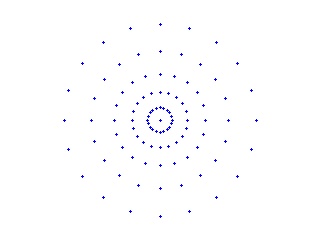}}
	\caption{Distribuição de pontos considerados para a estimação de fluxo óptico.\label{fig:distribuicao_pontos_monitorados}}
\end{figure}

Para estimar o fluxo óptico entre as duas imagens do ambiente capturadas em sequência, optou-se pela utilização do algoritmo de Lucas-Kanade, cuja implementação encontra-se disponível na OpenCV. Para utilizá-lo, considerou-se uma distribuição circular e simétrica de pontos, a qual foi definida a partir de experimentações visuais realizadas neste trabalho (\autoref{ap:selecao_pontos}). Esta consiste em 1 ponto central e 5 anéis concêntricos, cada um contendo 20 pontos igualmente espaçados. Estes anéis foram posicionados de modo que suas distâncias ao centro aumentassem de maneira exponencial. Tal distribuição foi gerada através de um \textit{script} em \textit{R} \cite{RCT2013} e salva num arquivo \textit{.dat}, de forma a ser carregada pelo sistema de navegação. Ao carregá-la, o sistema a projeta sobre o centro da imagem a ser utilizada para a estimação do fluxo. Esta projeção foi configurada para ocupar 80\% do tamanho da imagem. A \autoref{fig:distribuicao_pontos_monitorados} ilustra tal distribuição já projetada.

\begin{figure}[htb]
	\centerline{\includegraphics[width=0.4\linewidth]{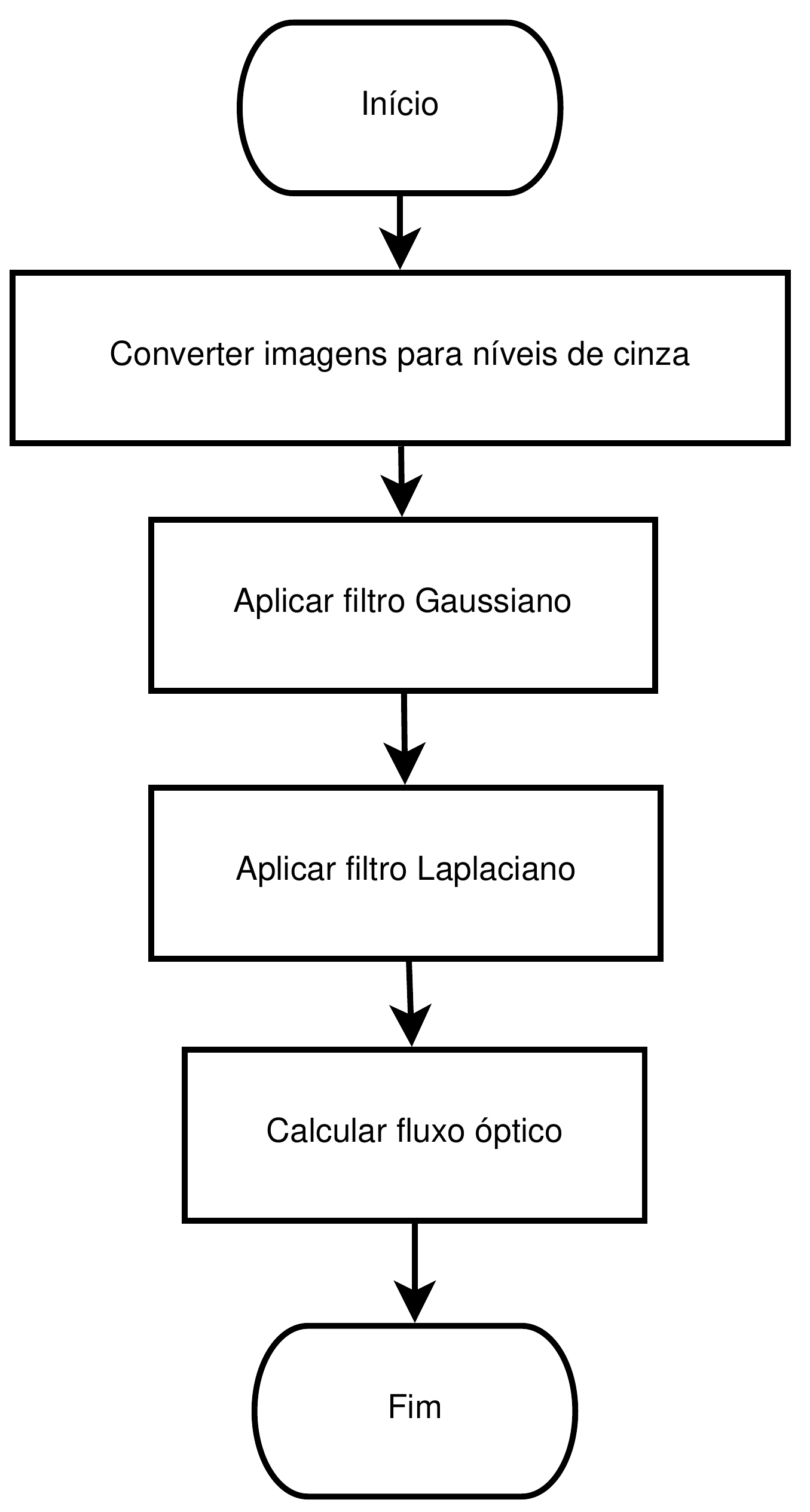}}
	\caption{Fluxograma da etapa de cálculo do fluxo óptico.\label{fig:fluxograma_fluxo_optico}}
\end{figure}

Uma vez definidos os pontos a serem monitorados e com base nas definições apresentadas no \autoref{cap:fluxo_optico}, considerou-se a aplicação de alguns filtros sobre as imagens utilizadas para a estimação do fluxo. A \autoref{fig:fluxograma_fluxo_optico} ilustra todo o processo realizado para o cálculo do fluxo óptico.

\begin{figure}[htb]
	\centering
	\begin{subfigure}[b]{0.3\linewidth}
		\includegraphics[width=\linewidth]{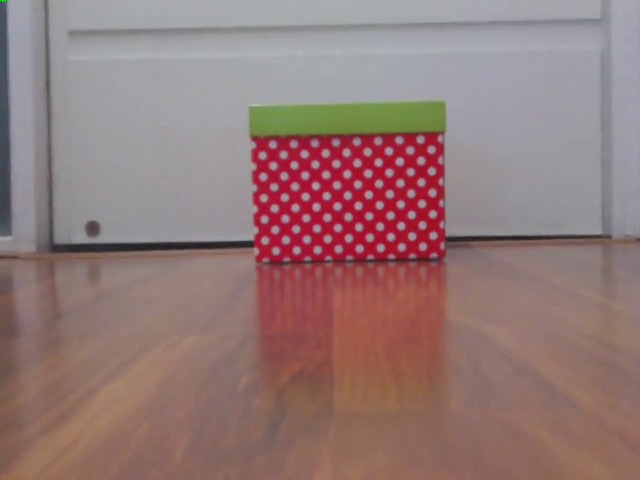}
		\caption{\label{fig:imagem_original_1}}
	\end{subfigure}
	\qquad
	\begin{subfigure}[b]{0.3\linewidth}
		\includegraphics[width=\linewidth]{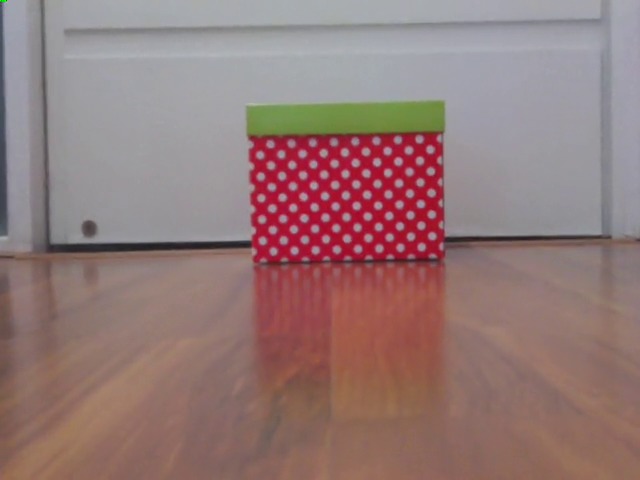}
		\caption{\label{fig:imagem_original_2}}
	\end{subfigure}
	\\
	\begin{subfigure}[b]{0.3\linewidth}
		\includegraphics[width=\linewidth]{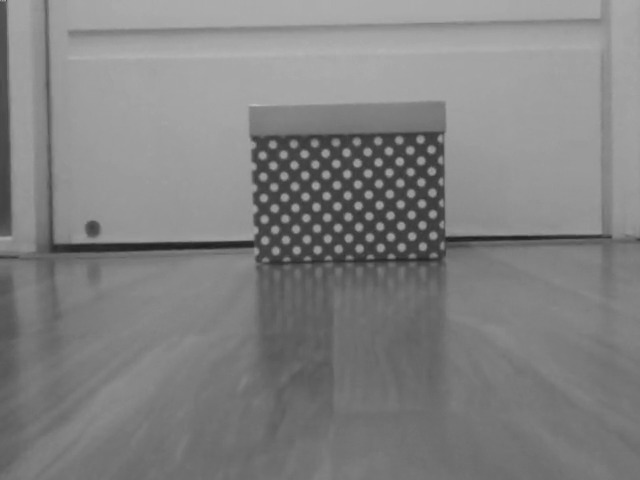}
		\caption{\label{fig:imagem_escala_cinza_1}}
	\end{subfigure}
	\qquad
	\begin{subfigure}[b]{0.3\linewidth}
		\includegraphics[width=\linewidth]{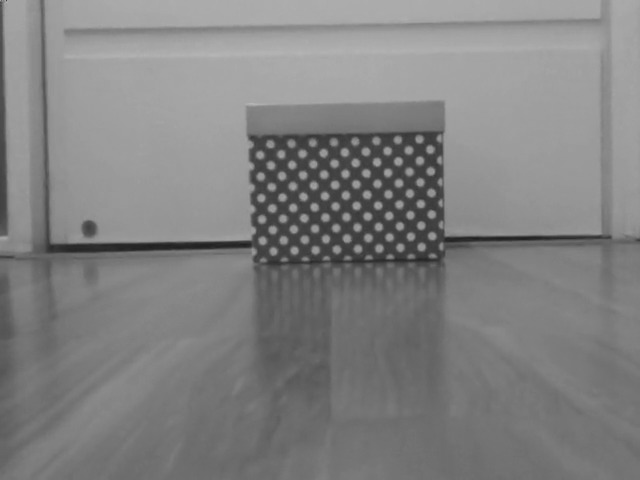}
		\caption{\label{fig:imagem_escala_cinza_2}}
	\end{subfigure}
	\\
	\begin{subfigure}[b]{0.3\linewidth}
		\includegraphics[width=\linewidth]{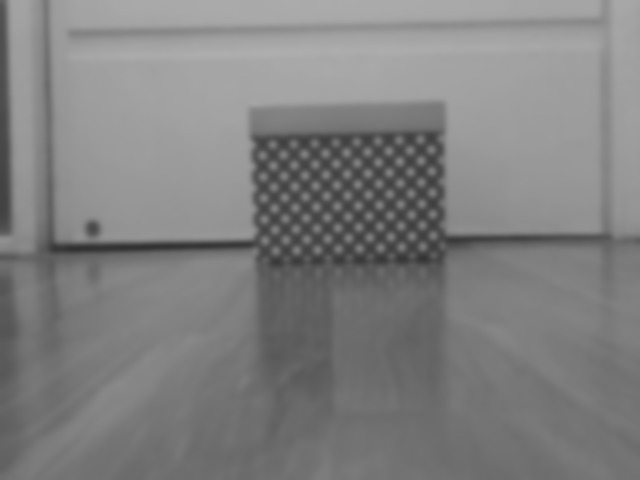}
		\caption{\label{fig:imagem_filtro_gaussiano_1}}
	\end{subfigure}
	\qquad
	\begin{subfigure}[b]{0.3\linewidth}
		\includegraphics[width=\linewidth]{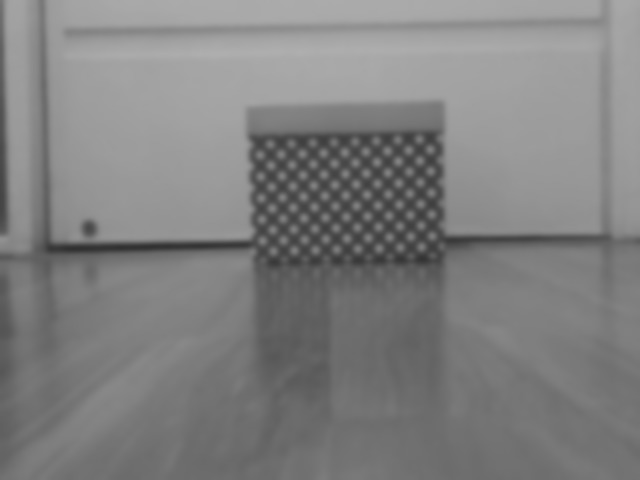}
		\caption{\label{fig:imagem_filtro_gaussiano_2}}
	\end{subfigure}
	\\
	\begin{subfigure}[b]{0.3\linewidth}
	\includegraphics[width=\linewidth]{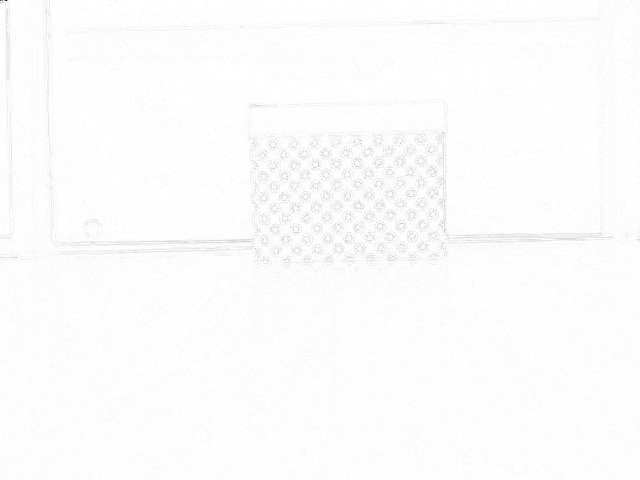}
	\caption{\label{fig:imagem_filtro_laplaciano_1}}
	\end{subfigure}
	\qquad
	\begin{subfigure}[b]{0.3\linewidth}
	\includegraphics[width=\linewidth]{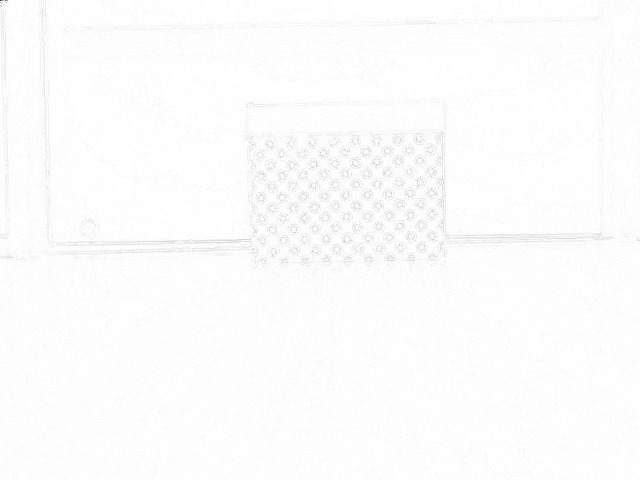}
	\caption{\label{fig:imagem_filtro_laplaciano_2}}
	\end{subfigure}
	\\
	\begin{subfigure}[b]{0.3\linewidth}
	\includegraphics[width=\linewidth]{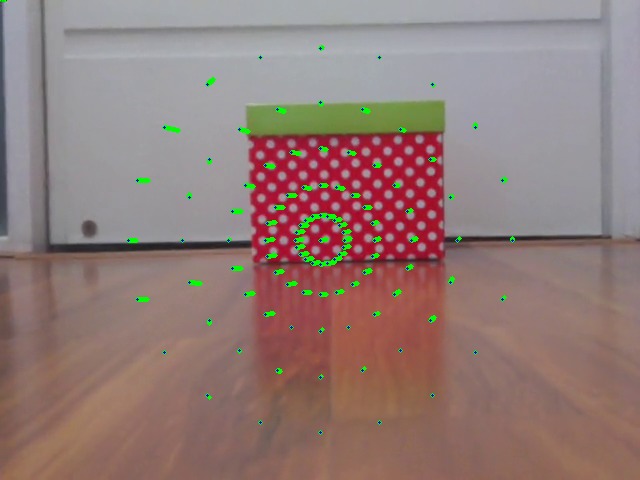}
	\caption{\label{fig:imagem_fluxo_optico}}
	\end{subfigure}
	\caption{Sequência de pré-processamento para o cálculo do fluxo óptico. \ref{fig:imagem_original_1} e \ref{fig:imagem_original_2} correspondem a capturas realizadas em $t$ e $t+1$, respectivamente. Já \ref{fig:imagem_escala_cinza_1} e \ref{fig:imagem_escala_cinza_2} apresentam a conversão de tais capturas para níveis de cinza. A aplicação de um filtro Gaussiano sobre estas conversões é ilustrada em \ref{fig:imagem_filtro_gaussiano_1} e \ref{fig:imagem_filtro_gaussiano_2}. Por sua vez, \ref{fig:imagem_filtro_laplaciano_1} e \ref{fig:imagem_filtro_laplaciano_2} ilustram a aplicação de um filtro Laplaciano sobre os resultados do filtro Gaussiano. Finalmente, \ref{fig:imagem_fluxo_optico} apresenta o fluxo óptico obtido.\label{fig:etapas_fluxo_optico}}
\end{figure}

Inicialmente, as imagens consideradas são convertidas para níveis de cinza, tendo em vista que a estimação do fluxo é feita com base na intensidade de seu brilho. Em seguida, aplica-se um filtro Gaussiano com tamanho de janela igual a $3x3$ \textit{pixels}. Este corresponde a um filtro passa-baixa capaz de suavizar as imagens de forma ponderada e simétrica \cite{Stringhini2011}. Assim, consegue-se atenuar termos derivativos de ordem mais alta e, consequentemente, tornar a aproximação da \autoref{eq:serie_taylor} válida. Logo após, as imagens suavizadas são submetidas a um filtro passa-alta Laplaciano. O objetivo de tal aplicação é realçar suas bordas, uma vez que estas correspondem às regiões com maior variação de brilho. Vale notar que neste trabalho deu-se preferência ao filtro Laplaciano em virtude do mesmo gerar como saída uma imagem em níveis de cinza, ao contrário de filtros como o de Canny \cite{Canny_1986}, o qual produz uma imagem binária. Dessa forma, consegue-se fornecer mais detalhes sobre as bordas realçadas. Finalmente, as imagens filtradas são utilizadas como entrada para o algoritmo de Lucas-Kanade. Este foi configurado com tamanho de vizinhança igual a $31x31$ \textit{pixels} e critérios de parada correspondentes ao valor máximo de 10 iterações e ao erro mínimo de 0,03. Além disso, utilizou-se a implementação piramidal deste algoritmo, considerando-se a quantidade total de 3 camadas \cite{Bouguet2001}. Esta implementação consiste na estimação recursiva do fluxo, de modo que cada camada está associada a um nível de resolução menor da imagem. Assim, através da implementação piramidal, consegue-se lidar com deslocamentos de padrões de brilho maiores que o tamanho da vizinhança considerada, os quais são gerados a partir de movimentações acentuadas. A \autoref{fig:etapas_fluxo_optico} ilustra a realização de cada uma destas etapas.

\clearpage

\subsection{Classificação do fluxo óptico}
\begin{figure}[htb]
	\centerline{\includegraphics[width=0.4\linewidth]{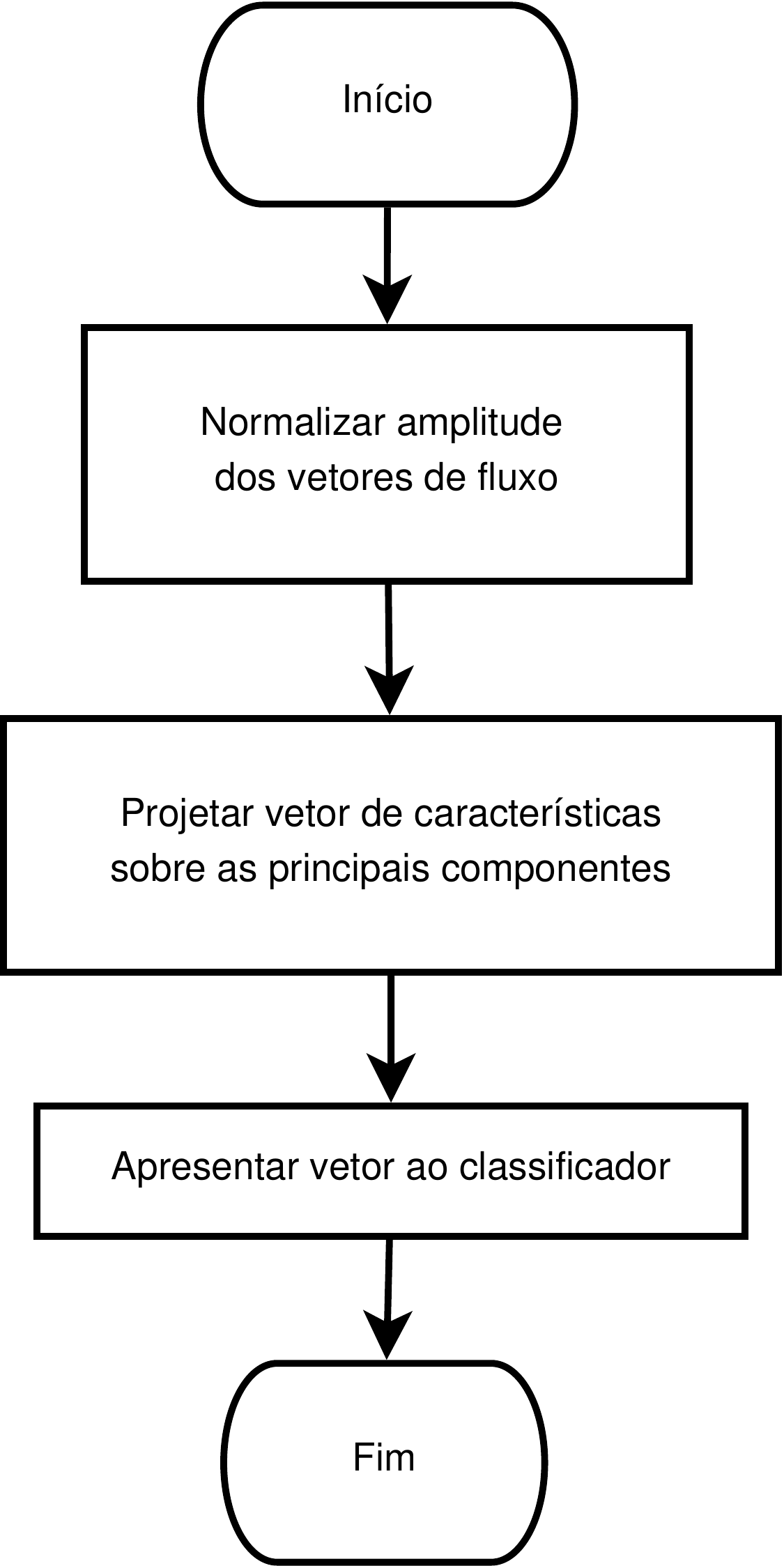}}
	\caption{Fluxograma da etapa de pré-processamento e classificação do fluxo óptico.\label{fig:fluxograma_classificacao_fluxo_optico}}
\end{figure}

Para classificar os padrões de fluxo óptico, utilizou-se a implementação de SVM (\autoref{ap:svm}) com \textit{kernel} BRF fornecida pela Scikit-Learn. Tal escolha justifica-se pelo alto grau de generalização alcançado por essa máquina, uma vez que a mesma é capaz de maximizar a margem de separação entre exemplos de classes distintas. Além disso, foi levada em conta a boa performance alcançada por classificadores SVM quando utilizados para o reconhecimento de padrões localizados em espaços com grande número de dimensões \cite{Duda2001}. É o caso do vetor de características formado pelos fluxos ópticos, o qual apresenta dezenas de atributos. Também é o caso de vetores de características formados a partir de padrões sequenciais de fluxo, através dos quais seria possível analisar a variação temporal da movimentação percebida. Apesar desta análise não ter sido realizada neste trabalho, a mesma pode ser investigada em trabalhos futuros. Uma vez treinada, esta máquina seria utilizada para apontar se o padrão de fluxo apresentado indica ou não a existência de obstáculos. Porém, antes que tal apresentação fosse feita, seria preciso condicionar o vetor de características, de modo a facilitar o processo de classificação. A sequência de passos executados nesta etapa é ilustrada na \autoref{fig:fluxograma_classificacao_fluxo_optico}.

Primeiramente, a amplitude do fluxo é normalizada linearmente entre o intervalo $[0,1]$. O objetivo desta normalização é homogeneizar a representação de padrões que encontram-se em escalas diferentes mas possuem a mesma relevância. \cite{Theodoridis2003}. Em seguida, o vetor normalizado é projetado para um espaço de dimensões reduzidas. Este é definido pelos vetores obtidos através do algoritmo PCA (\textit{Principal Component Analysis} \cite{hotelling1933analysis}) durante a fase de treinamento da SVM. O objetivo deste redimensionamento é otimizar o processo de classificação, tendo em vista a redução da quantidade de atributos a serem apresentados à SVM, além da atenuação de ruídos. Finalmente, o vetor obtido é apresentado à SVM, a qual sinalizará se o mesmo indica ou não a presença de obstáculos.

\subsection{Tomada de decisão}
\begin{figure}[htb]
	\centerline{\includegraphics[width=\linewidth]{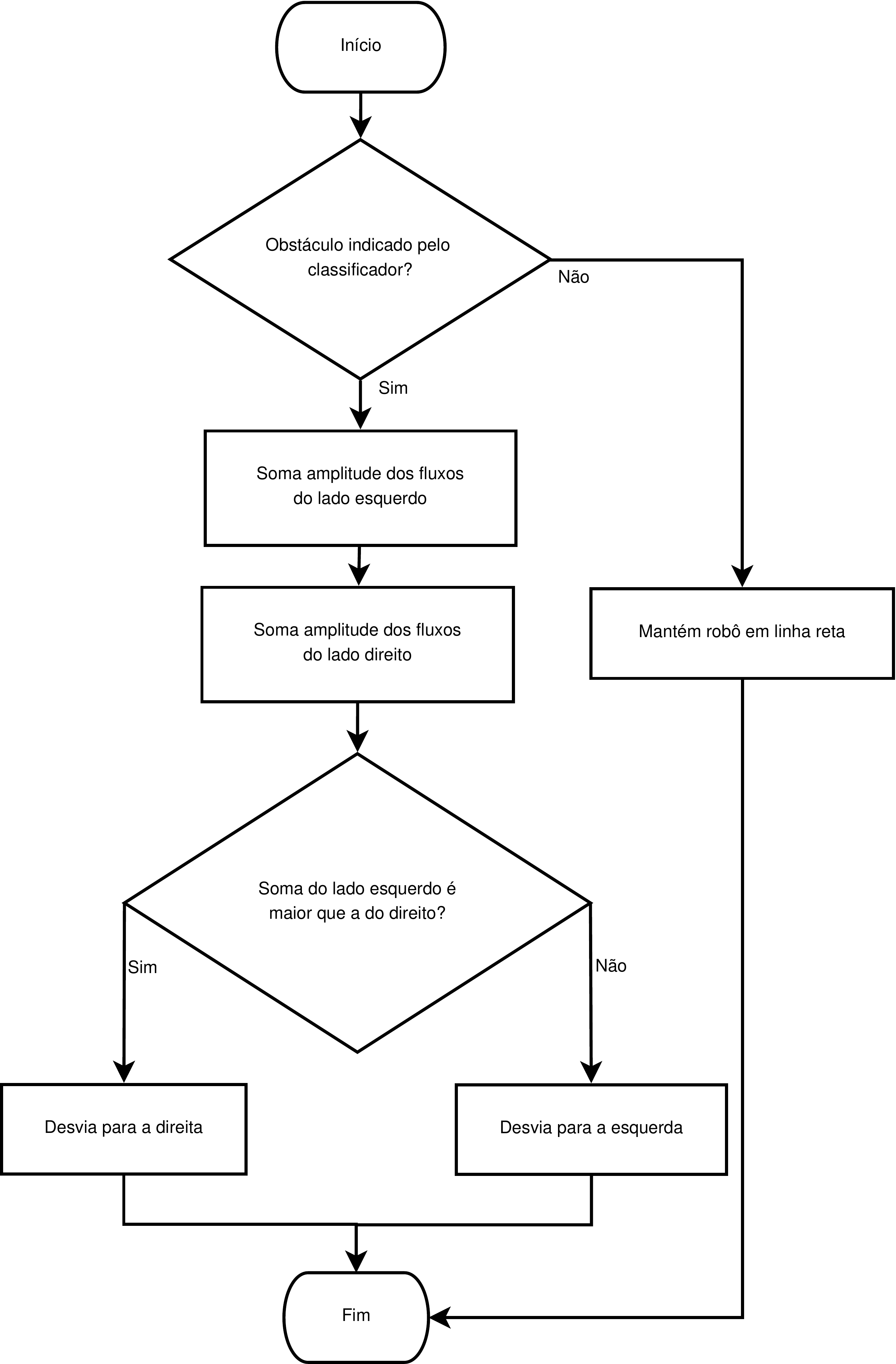}}
	\caption{Fluxograma da etapa de tomada de decisão.\label{fig:fluxograma_tomada_de_decisao}}
\end{figure}

\begin{figure}[htb]
\centering
\begin{subfigure}[b]{0.4\linewidth}
	\includegraphics[width=\linewidth]{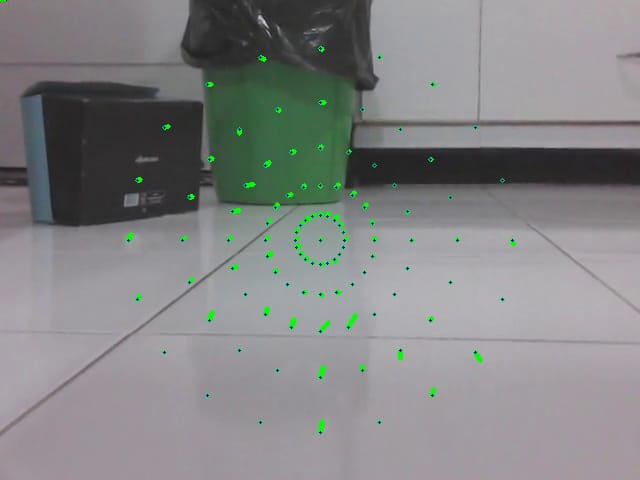}
	\caption{\label{fig:fluxo_optico_tomada_de_decisao_vetores}}
\end{subfigure}
\qquad
\begin{subfigure}[b]{0.4\linewidth}
	\includegraphics[width=\linewidth]{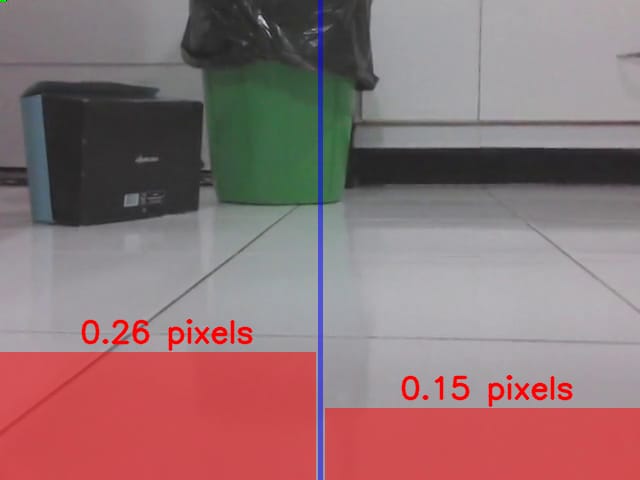}
	\caption{\label{fig:fluxo_optico_tomada_de_decisao_media}}
\end{subfigure}
\caption{Comparação de fluxo óptico para tomada de decisão. \ref{fig:fluxo_optico_tomada_de_decisao_vetores} apresenta o fluxo óptico calculado sobre uma determinada cena. \ref{fig:fluxo_optico_tomada_de_decisao_media} ilustra a amplitude média dos vetores de fluxo associados a cada partição da imagem.\label{fig:fluxo_optico_tomada_de_decisao}}
\end{figure}

A \autoref{fig:fluxograma_tomada_de_decisao} ilustra o processo de tomada de decisão com base na indicação do classificador. Caso este sinalize a ausência de obstáculos, deve-se conduzir o robô em linha reta. Caso contrário, realiza-se uma manobra de desvio. Neste caso, é preciso definir qual será a direção tomada. Para isso, optou-se em dividir a última imagem capturada em duas partições simétricas. Em seguida, escolhe-se a direção da partição na qual o valor médio das amplitudes do fluxo óptico seja menor. Esta estratégia considera que a direção com menor intensidade de fluxo tem menos chance de apresentar obstáculos próximos, uma vez que a movimentação relativa percebida entre o robô e os elementos contidos no ambiente é menor. É o caso da cena apresentada na \autoref{fig:fluxo_optico_tomada_de_decisao}, na qual os objetos contidos no lado esquerdo contribuem para que a média da amplitude dos vetores de fluxo nesta partição da imagem seja maior que a média da parte oposta. Nesta situação, o robô seria direcionado para o lado direito.

\subsection{Acionamento dos atuadores}
\begin{figure}[htb]
	\centerline{\includegraphics[width=0.9\linewidth]{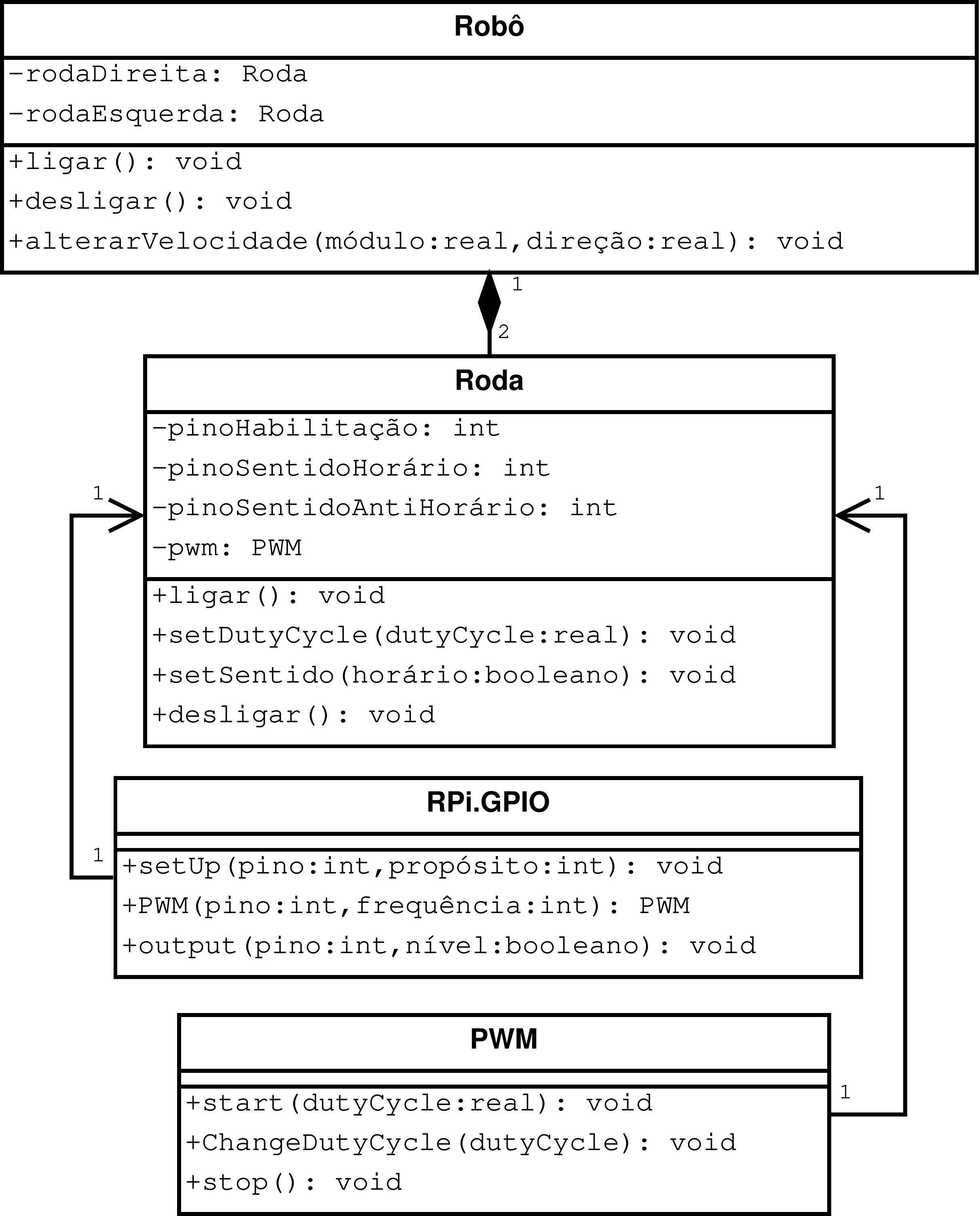}}
	\caption{Diagrama de classes de controle dos atuadores.\label{fig:classes_controle_atuadores}}
\end{figure}

Para acionar os atuadores do sistema de navegação, utilizou-se como base o módulo Python RPi.GPIO \cite{Croston2017}. Este corresponde a uma classe que disponibiliza funções para acessar e controlar a interface GPIO da Raspberry Pi. Para utilizá-la, foram projetadas duas novas classes: \textit{Roda} e \textit{Robô}. A primeira teria como finalidade modelar o comportamento dos atuadores a serem controlados através da interface GPIO. Já a segunda teria como papel oferecer ao sistema de navegação uma interface de controle de alto nível. A \autoref{fig:classes_controle_atuadores} apresenta o diagrama UML que descreve a relação entre as classes utilizadas.

Como visto no diagrama, a classe \textit{Roda} possui 4 atributos, dentre os quais 3 representam pinos presentes em circuitos \textit{Ponte H}, como os contidos no CI L293D (\autoref{fig:ponteh_ilustracoes}) utilizado para conectar os motores do robô à interface GPIO. O atributo \textit{pinoHabilitação} refere-se ao pino que habilita a ponte. Este é análogo ao pino \textit{1,2EN} do L29D3D. Já os atributos \textit{pinoSentidoHorário} e \textit{pinoSentidoAntiHorário} referem-se às chaves de controle da ponte. Estas são análogas aos pinos \textit{1A} e \textit{2A} do L29D3D. Ao combiná-las é possível alterar o sentido da corrente que passa pela ponte, de modo a configurar o sentido de rotação da roda associada. Cada pino de um objeto \textit{Roda} é inicializado utilizando-se o método \textit{setUp} disponibilizado por um objeto \textit{RPi.GPIO}. Este recebe como parâmetros a numeração do pino real da interface GPIO que será considerado e a indicação do propósito do pino, ou seja, se o mesmo será utilizado como entrada ou saída de sinal. No caso da classe \textit{Roda}, todos os pinos são configurados como saídas. Já o último atributo corresponde a um objeto capaz de controlar a aplicação da modulação PWM sobre qualquer pino de propósito geral da Raspberry Pi. Este objeto é instanciado por meio do método \textit{PWM}, fornecido pela classe \textit{RPi.GPIO}. Como parâmetros para sua instanciação foram passados uma referência ao atributo \textit{pinoHabilitação} e o valor 2000Hz, referente à frequência da onda gerada.

Sendo assim, a partir destes atributos, foi possível modelar o comportamento desejado para cada roda do robô com base na elaboração de 4 métodos: \textit{ligar}, \textit{setDutyCycle}, \textit{setSentido} e \textit{desligar}. O primeiro refere-se à inicialização da geração do sinal modulado sobre o pino de habilitação da roda modelada. Sua implementação consiste na invocação do método \textit{start}, implementado pela classe \textit{PWM}, passando-se como parâmetro o valor referente à largura de pulso nula. Por sua vez, o segundo método refere-se à alteração da largura do pulso gerado e aplicado sobre o pino de habilitação da roda. Esta largura é diretamente proporcional ao torque da roda, sendo que sua alteração é realizada através do método \textit{ChangeDutyCycle}, fornecido pela classe \textit{PWM}. Já o método \textit{setSentido} refere-se à configuração do sentido de rotação da roda. Sua implementação é baseada na habilitação das chaves de controle da ponte H associada à roda modelada, as quais são representadas pelos atributos \textit{pinoSentidoHorário} e \textit{pinoSentidoAntiHorário}. Para o caso do sentido desejado ser o horário, coloca-se em nível lógico alto o primeiro atributo e em nível lógico baixo o segundo. Já no caso do sentido oposto, o primeiro atributo é colocado em nível lógico baixo enquanto que o segundo é colocado em nível lógico alto. Tais atribuições de nível lógico são realizadas pela classe \textit{Roda} através do método \textit{output} fornecido por um objeto \textit{RPi.GPIO}. Finalmente, o terceiro método corresponde ao encerramento do trabalho realizado pela roda modelada. Sua implementação consiste em cessar-se o fornecimento do sinal modulado para o pino de habilitação da roda. Para isso, utiliza-se o método \textit{stop} fornecido pela classe \textit{PWM}. 

Para modelar o comportamento físico do robô projetado e fornecer uma interface de alto nível para o controle dos atuadores, projetou-se a classe \textit{Robô}. Os métodos disponibilizados pela mesma são implementados com base na manipulação de dois objetos \textit{Roda}, os quais a compõem. Tais métodos são: \textit{ligar}, \textit{desligar} e \textit{alterarVelocidade}. Os dois primeiros consistem na invocação dos métodos de mesmo nome implementados pelos objetos do tipo \textit{Roda}. Por sua vez, o terceiro método consiste na configuração das velocidades linear e angular do robô. Seus parâmetros correspondem ao módulo e à fase do vetor $\vec{V_R}$ da velocidade resultante desejada (\autoref{fig:vetor_velocidade}). O valor do primeiro parâmetro é dado em termos percentuais de ciclo de trabalho, enquanto que o segundo é dado em radianos. As componentes do vetor resultante foram modeladas da seguinte forma:

\begin{figure}[htb]
	\centerline{\includegraphics[width=0.9\linewidth]{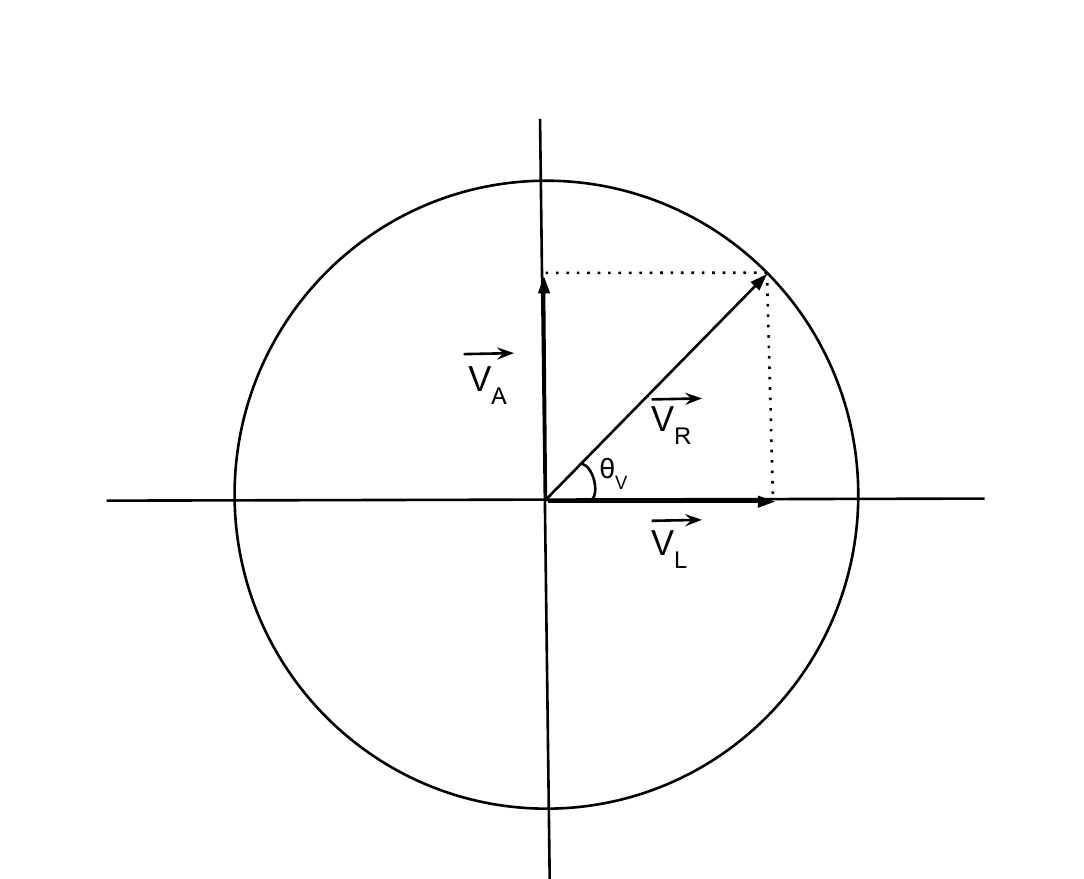}}
	\caption{Ilustração do vetor velocidade e suas componentes.\label{fig:vetor_velocidade}}
\end{figure}

\begin{equation}
|\vec{V_L}| = |\vec{V_R}|cos(\theta_V)
\end{equation}
\begin{equation}
|\vec{V_A}| = |\vec{V_R}|sen(\theta_V)
\end{equation}

\noindent , sendo $|\vec{V_L}|$ e $|\vec{V_A}|$ o módulo das componentes proporcionais às velocidades linear e angular do robô, respectivamente, e $\theta_V$ a inclinação do vetor $\vec{V_R}$.

Inicialmente, o método \textit{alterarVelocidade} utiliza as expressões anteriores para calcular o módulo das componentes da velocidade resultante desejada. Em seguida, o valor $|\vec{V_A}|$ é utilizado para configurar a velocidade angular do robô. Este valor é interpretado como sendo a diferença entre os ciclos de trabalho de cada roda. Sendo assim, o mesmo é passado como argumento para o método \textit{setDutyCycle} da roda mais distante ao centro de rotação do robô. Em seguida, configura-se o sentido de rotação das rodas com base no cosseno do ângulo $\theta_V$. Caso este valor seja positivo, o sentido das rodas é configurado de modo a fazer com que o robô se locomova para frente. Caso seja negativo, configuram-se as rodas para girarem no sentido contrário. Esta configuração é realizada a partir do método \textit{setSentido} da classe \textit{Roda}. Finalmente, o valor $|\vec{V_L}|$ é adicionado ao ciclo de trabalho de ambas as rodas através do método \textit{setDutyCycle}, respeitando-se seu limite máximo, uma vez que tal ciclo é definido de maneira percentual. Vale ressaltar que o método \textit{alterarVelocidade} dá prioridade à aplicação da velocidade angular em detrimento da linear. Esta é interpretada pela classe \textit{Robô} como sendo a velocidade apresentada pela roda mais lenta. Sendo assim, há a possibilidade da velocidade linear desejada não ser atingida devido à saturação do valor de ciclo de trabalho de alguma das rodas.

Por fim, a interface fornecida pela classe \textit{Robô} foi utilizada pelo sistema de navegação para efetuar o acionamento dos atuadores após a tomada de decisão. Nos casos em que a trajetória do robô deveria ser mantida em linha reta, utilizou-se o método \textit{alterarVelocidade} com os argumentos velocidade percentual igual a 50\% e direção igual a 0rad. Já para a efetuação dos desvios, utilizou-se, inicialmente, o mesmo método com os argumentos velocidade percentual igual a 60\% e direção igual a $\frac{\pi}{2}$rad (desvio para a direita) e $\frac{3\pi}{2}$rad (desvio para a esquerda). Após 200ms, utilizou-se novamente o método \textit{alterarVelocidade}, passando-se como argumentos os mesmos valores empregados para a trajetória em linha reta.

%% file: Conteudo/ExperimentosResultados.tex
\chapter{Avaliação da Plataforma}
\label{cap:experimentos_e_resultados}

A plataforma desenvolvida foi avaliada com base em dois experimentos: um focado exclusivamente na performance \textit{offline} do seu classificador e outro com ênfase no seu desempenho \textit{online}. As próximas seções descrevem as metodologias utilizadas para cada tipo de avaliação.

\section{Avaliação \textit{Offline}}
\label{sec:avaliacao_offline}

\subsection{Metodologia}
A metodologia \textit{offline} empregada para avaliar a qualidade do sistema consistiu na apresentação de determinados exemplos previamente rotulados e no posterior registro da quantidade de indicações corretas e incorretas realizadas pelo seu classificador, juntamente com o tempo levado para realizar cada indicação. Mais especificamente, utilizou-se o método de validação cruzada \textit{k-fold} \cite{kohavi1995study} para particionar uma determinada base de dados em $k$ divisões mutuamente exclusivas. Tal base foi formada a partir de padrões de fluxo óptico já rotulados. Durante $k$ iterações, os exemplos de $(k-1)$ partições foram utilizados para treinar o classificador. Ao fim do treino, os exemplos da partição restante foram apresentados a este último, o qual deveria prever a classe correspondente a cada um deles. Finalmente, as predições feitas pelo classificador foram comparadas aos valores dos rótulos previamente conhecidos, de modo a serem registradas as quantidades de previsões corretas e incorretas. É importante notar que durante cada iteração deste processo, a partição não utilizada para treino foi alternada. Desse modo, garante-se que ao fim do processo de validação o classificador tenha predito a classe de todos os exemplos contidos nas $k$ partições.

\begin{figure}[htb]
	\centerline{\includegraphics[width=0.7\linewidth]{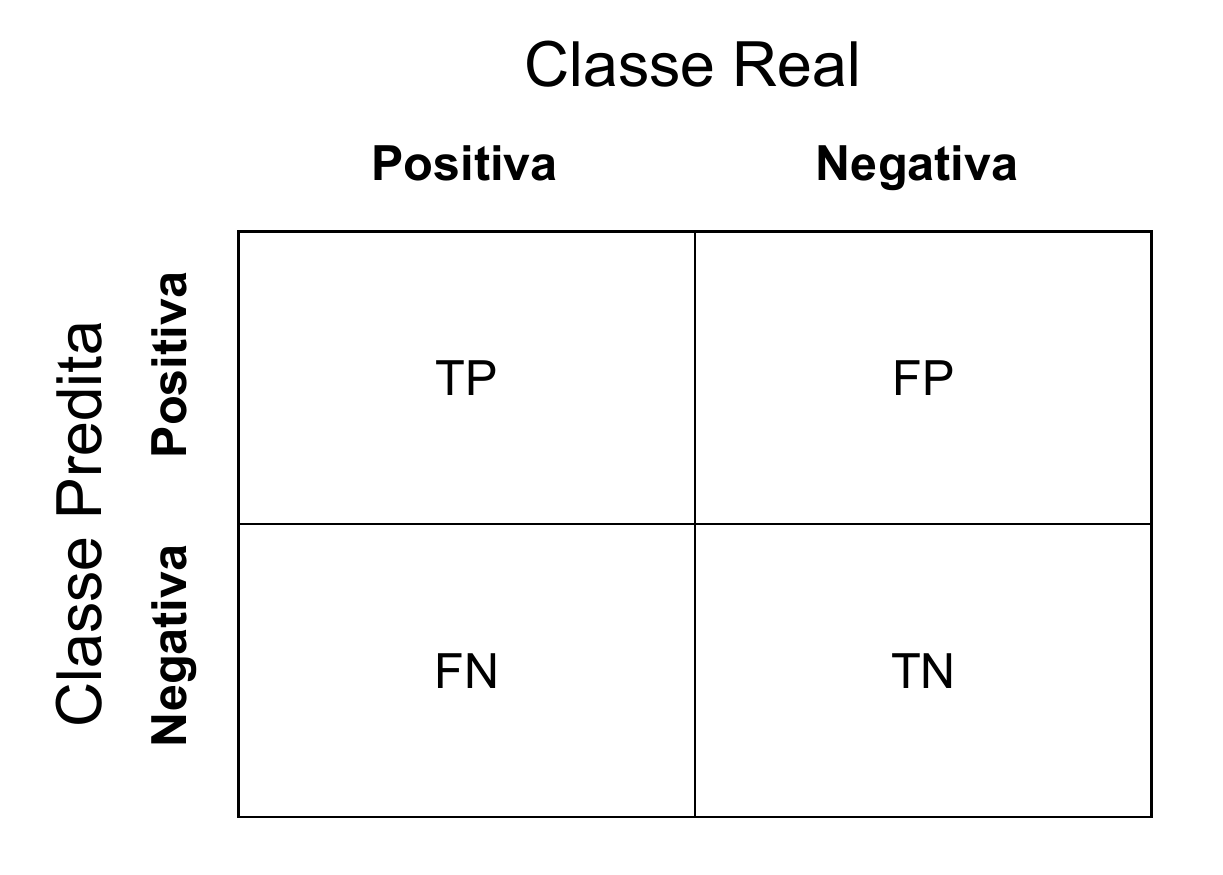}}
	\caption{Ilustração de uma matriz de confusão. Suas linhas são definidas pela previsão do classificador, enquanto que suas colunas são definidas pelo valor da classe a qual um determinado exemplo realmente pertence.\label{fig:matriz_confusao}}
\end{figure}

Os erros e acertos do classificador ao longo de todo o processo de validação foram registrados para ambas as classes, de modo a ser preenchida sua matriz de confusão. Esta é ilustrada na \autoref{fig:matriz_confusao}. Como pode ser visto, seu preenchimento é realizado de modo que a célula TP (\textit{True Positive}) contenha a quantidade de exemplos corretamente preditos pelo classificador como sendo da classe positiva, FP (\textit{False Positive}) contenha a quantidade de exemplos classificados incorretamente como positivos, TN (\textit{True Negative}) contenha a quantidade de exemplos corretamente apontados como negativos e FN (\textit{False Negative}) contenha a quantidade de exemplos incorretamente preditos como negativos.

A partir da matriz de confusão, foram extraídas medidas que expressam de maneira mais clara as relações entre os erros e acertos do classificador para ambas as classes. Tais medidas correspondem à precisão (\autoref{eq:precisao}), à cobertura (\autoref{eq:cobertura}), à media F (\autoref{eq:medida_f}) e à acurácia (\autoref{eq:acurácia}). A precisão refere-se à qualidade do classificador em acertar previsões para a classe positiva. Já a cobertura mede a quantidade de exemplos pertencentes à classe positiva que foram corretamente preditos pelo classificador. Por sua vez, a medida F representa uma média harmônica entre a precisão e a cobertura. Finalmente, a acurácia informa o total de previsões corretas realizadas pelo classificador independentemente da classe apontada.

\begin{equation} \label{eq:precisao}
	\textnormal{Precisão} = \frac{TP}{TP+FP}
\end{equation}

\begin{equation} \label{eq:cobertura}
	\textnormal{Cobertura} = \frac{TP}{TP+FN}
\end{equation}

\begin{equation} \label{eq:medida_f}
	\textnormal{Medida F} = \frac{2*\textnormal{Precisão}*\textnormal{Cobertura}}{\textnormal{Precisão}+\textnormal{Cobertura}}
\end{equation}

\begin{equation} \label{eq:acurácia}
	\textnormal{Acurácia} = \frac{TP+TN}{TP+FP+TN+FN}
\end{equation}

Os valores de tais medidas, juntamente com aqueles referentes ao tempo de execução, foram registrados ao fim do processo de validação cruzada, de modo a serem analisados e comparados aos obtidos pelos trabalhos relacionados.

\subsection{Base de Dados}
\begin{figure}[htb]
	\centerline{\includegraphics[width=0.7\linewidth]{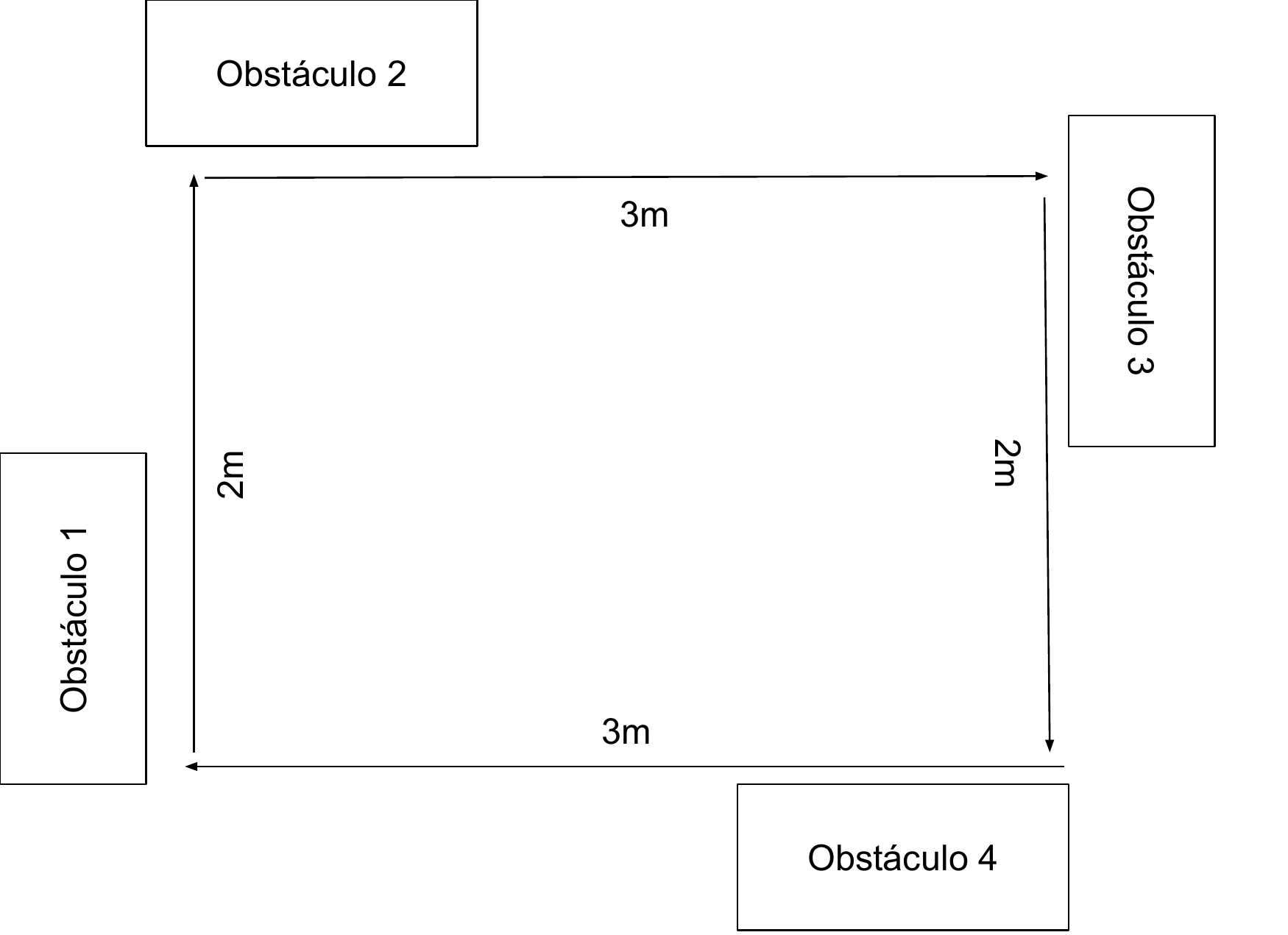}}
	\caption{Ilustração do circuito de teste \textit{offline} elaborado.\label{fig:diagrama_circuito_teste}}
\end{figure}

\begin{figure}[htb]
	\centerline{\includegraphics[width=0.7\linewidth]{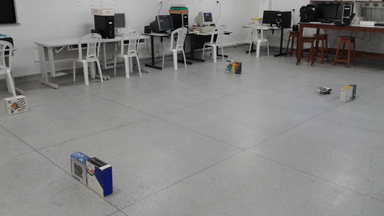}}
	\caption{Ambiente no qual foram gravados os vídeos da base de dados.\label{fig:ambiente_teste}}
\end{figure}

A base de dados utilizada para o processo de validação do classificador foi formada por padrões de fluxo óptico extraídos de 8 vídeos. Estes foram gravados com o auxílio da própria plataforma robótica, de modo que as imagens capturadas representassem o mesmo ponto de vista percebido pelo robô durante a navegação autônoma. Para isso, o robô foi guiado remotamente por um controlador humano ao longo de um circuito contendo diferentes obstáculos. O objetivo foi registrar o movimento de aproximação de cada obstáculo ao observador da imagem. Ao aproximar-se suficientemente de um destes, o robô foi sempre desviado para o próximo e assim sucessivamente. Este comportamento é ilustrado na \autoref{fig:diagrama_circuito_teste}. Nela é possível notar a distribuição espacial dos 4 obstáculos considerados, além do trajeto percorrido pelo robô. Dessa forma, a partir dessa estratégia foi possível gravar sequências contínuas de navegação. A \autoref{fig:ambiente_teste} apresenta o ambiente no qual os vídeos foram gravados.

Os vídeos foram gravados com a resolução $320x240$ \textit{pixels}, sendo a velocidade linear do robô fixada em 0,1m/s, aproximadamente. Cada vídeo consistiu em capturas realizadas durante 3 voltas completas do robô ao longo do circuito. Além disso, para avaliar posteriormente o nível de fixação e aprendizado do sistema de navegação, pensou-se em alterar as características do ambiente à medida em que novos vídeos fossem gravados. Para isso, foi definido que entre cada uma das quatro primeiras gravações seria alternado o posicionamento dos obstáculos. Por exemplo, considerando que inicialmente os obstáculos encontram-se distribuídos segundo ilustrado na \autoref{fig:diagrama_circuito_teste}, após a gravação do vídeo inicial o primeiro obstáculo foi colocado na posição do segundo, o qual foi posicionado no lugar do terceiro, que por sua vez foi colocado no lugar do quarto, o qual foi deslocado para a posição do primeiro. A partir deste novo arranjo, grava-se o segundo vídeo. Em seguida, repete-se o mesmo processo de redistribuição sequencial de obstáculos antes de gravar-se o terceiro vídeo. Por fim, o mesmo foi realizado para a gravação do quarto vídeo.

De forma semelhante foi realizado o processo de alteração do posicionamento dos obstáculos para a gravação dos 4 últimos vídeos. A distribuição do vídeo 5 é análoga à do vídeo 1, a do vídeo 6 assemelha-se à do vídeo 2 e assim sucessivamente. A diferença neste caso é que o sentido da navegação corresponde ao oposto daquele considerado durante a gravação dos 4 vídeos iniciais. Com essa alteração, conseguiu-se não apenas registrar os mesmos obstáculos em frente a planos de fundo diferentes, mas também capturar movimentações de curvas para direções até então inéditas. Dessa forma, esta alteração foi relevante na medida em que pôde ser utilizada para ensinar ao robô qual é o padrão de movimento de um obstáculo sendo desviado nos dois sentidos possíveis.

Finalmente, a base de dados a ser formada deveria contar não apenas com os padrões de fluxo óptico, mas também com seus respectivos rótulos. Neste trabalho, entendeu-se que não seria viável realizar a atribuição de rótulos manualmente, uma vez que tal processo é de natureza subjetiva. Ou seja, a depender do supervisor humano os padrões poderiam ser rotulados de forma diferente, fato este que inviabilizaria a reprodução deste mesmo processo por diferentes trabalhos. Além disso, como a base formada tende a possuir milhares de exemplos, sua rotulação manual corresponde a um processo extremamente exaustivo, de certo modo inviável para aplicações práticas. Sendo assim, com a intenção de automatizar este processo, utilizou-se o sensor ultrassônico HC-SR04, o qual já havia sido instalado na plataforma. A partir deste sensor foi possível registrar a distância entre o robô e os obstáculos. Estas foram posteriormente armazenadas de modo que pudessem ser facilmente associadas aos seus respectivos quadros contidos nos vídeos capturados.

\begin{figure}[htb]
	\centering
	\begin{subfigure}[b]{0.4\linewidth}
		\includegraphics[width=\linewidth]{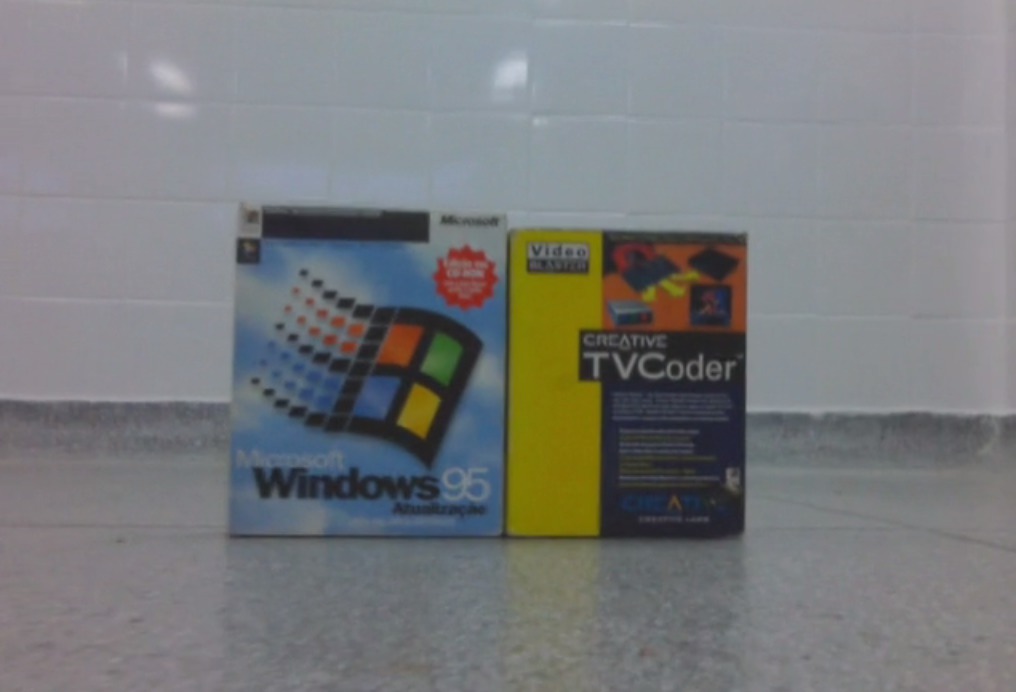}
		\caption{\label{fig:obstaculo1}}
	\end{subfigure}
	\qquad
	\begin{subfigure}[b]{0.4\linewidth}
		\includegraphics[width=\linewidth]{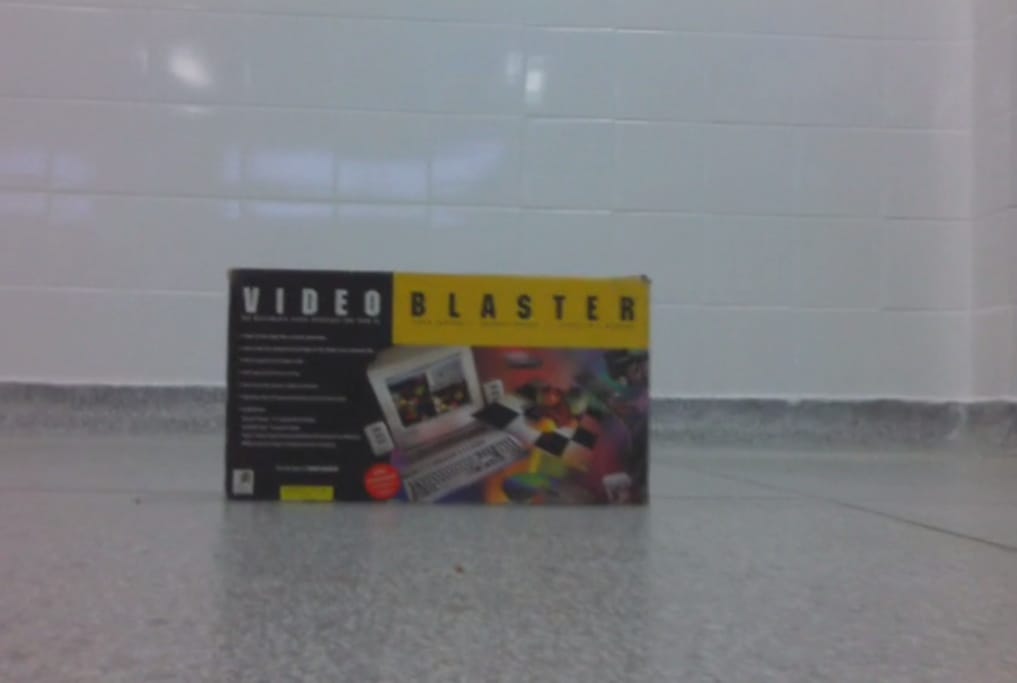}
		\caption{\label{fig:obstaculo2}}
	\end{subfigure}
	\qquad
	\begin{subfigure}[b]{0.4\linewidth}
		\includegraphics[width=\linewidth]{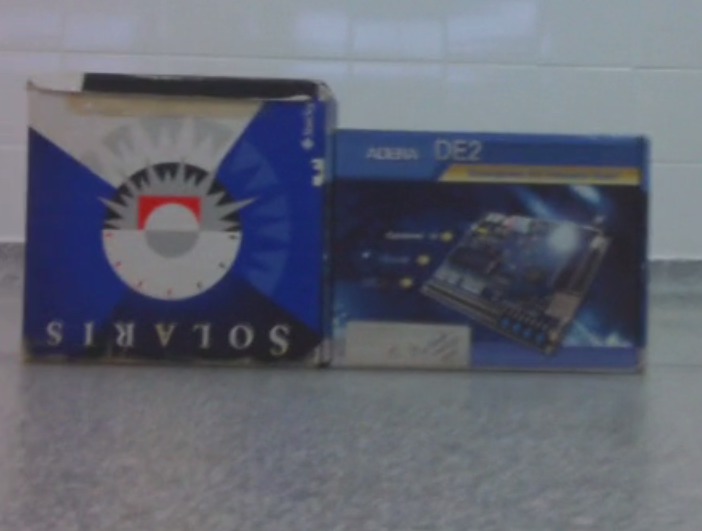}
		\caption{\label{fig:obstaculo3}}
	\end{subfigure}
	\qquad
	\begin{subfigure}[b]{0.4\linewidth}
		\includegraphics[width=\linewidth]{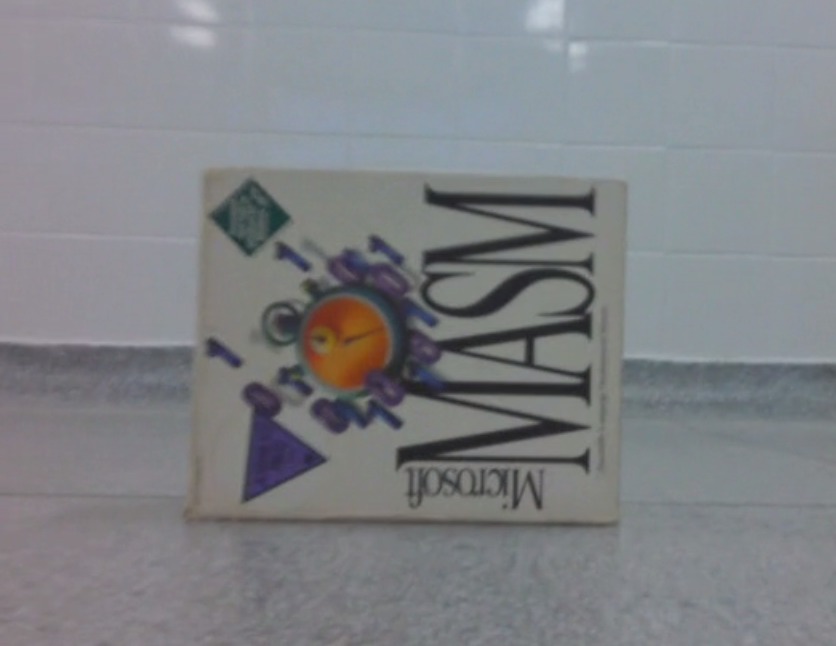}
		\caption{\label{fig:obstaculo4}}
	\end{subfigure}
	\caption{Diferentes obstáculos inseridos no circuito de teste \textit{offline}.\label{fig:obstaculos}}
\end{figure}

\begin{figure}[htb]
	\centerline{\includegraphics[width=0.7\linewidth]{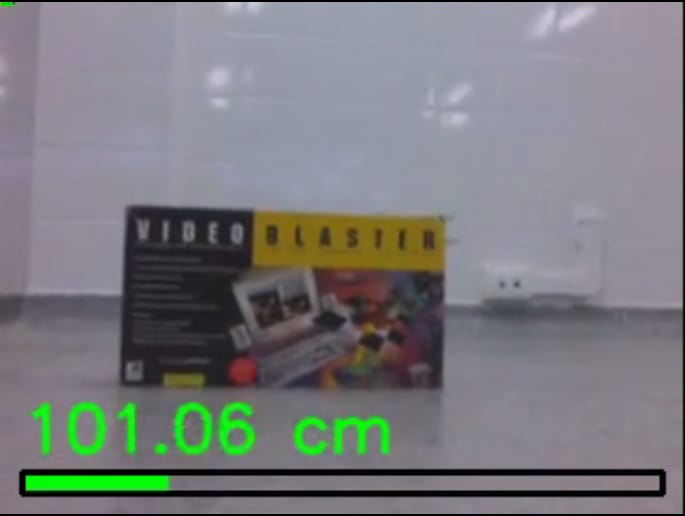}}
	\caption{Exemplo de quadro capturado durante gravação da base de dados. Sobre o quadro foi ilustrada a medição do sensor ultrassônico.\label{fig:exemplo_captura_teste}}
\end{figure}

Tendo em vista que o sensor ultrassônico deveria ser capaz de registrar com precisão e exatidão a distância do robô aos obstáculos, os mesmos foram escolhidos de modo a maximizar a qualidade da sua medição. Sendo assim, foram escolhidos os obstáculos exibidos na \autoref{fig:obstaculos}. Os mesmos apresentam superfície lisa e reflexiva, sendo apropriados para a utilização deste tipo de sensor de alcance. A \autoref{fig:exemplo_captura_teste} apresenta um dos quadros capturados durante a navegação do robô, juntamente com a medida registrada pelo sensor ultrassônico. Por fim, vale ressaltar que estes mesmos obstáculos também foram selecionados devido ao seu alto nível de textura, em contraste ao dos demais elementos presentes no cenário considerado. Através deste grau de textura torna-se possível estimar padrões de fluxo óptico mais bem definidos.

Finalmente, uma vez capturados os vídeos, calculou-se o fluxo óptico associado aos mesmos. Para isso, utilizou-se a implementação do algoritmo de Lucas-Kanade disponibilizada pela OpenCV. Além disso, consideram-se os mesmos parâmetros utilizados durante a classificação, os quais foram detalhados no \autoref{cap:plataforma_proposta}. Ao fim, foram gerados 8 arquivos contendo amostras de fluxo. Por sua vez, as medidas obtidas pelo ultrassom foram convertidas em rótulos \textit{-1} e \textit{1} da seguinte forma: 

\begin{equation} \label{eq:criacao_rotulos}
f(m) =
\left \{
\begin{array}{cc}
1, & \textnormal{se } L_{inf} \leq m \leq L_{sup} \\
-1, & \textnormal{caso contrário} \\
\end{array}
\right.
\end{equation}

\noindent , onde $m$ é o valor da medida do sensor ultrassônico, $L_{sup}$ representa a distância máxima para que um objeto seja considerado obstáculo e $L_{inf}$ refere-se ao valor de medida máximo considerado ruído. Neste trabalho, consideram-se $L_{sup}=$ 70cm e $L_{inf}=$ 10cm. 

As características da base final gerada são listadas a seguir:

\begin{itemize}
	\item Quantidade de atributos: 202;
	\item Quantidade de exemplos da classe positiva: 6533;
	\item Quantidade de exemplos da classe negativa: 31627;
	\item Quantidade total de exemplos: 38160;
\end{itemize}

\subsection{Treinamento}
O processo de treinamento do classificador ao longo de cada iteração da validação cruzada consistiu, inicialmente, na extração das principais características dos padrões a serem ensinados. Considerou-se que tal extração seria necessária para facilitar o processo de aprendizado do classificador, uma vez que os exemplos a serem ensinados continham centenas de atributos. Sendo assim, optou-se pela aplicação da PCA (\textit{Principal Component Analysis}) sobre a base de treinamento, através da implementação fornecida pela biblioteca Scikit-Learn \cite{scikit-learn}. Para utilizá-la foi preciso apenas informar a base de dados a ser analisada e a quantidade percentual de informação que deveria ser agregada no conjunto de componentes extraídas. Tal valor foi definido como 90\% devido ao entendimento de que esta porcentagem seria suficiente para conter as informações mais relevantes para a classificação, reduzir drasticamente o número de dimensões dos padrões a serem ensinados ao sistema e, por fim, reduzir-se o nível de dados ruidosos. Ao final da aplicação da PCA, foram armazenadas as principais componentes extraídas. A intenção foi utilizá-las tanto durante o treinamento quanto ao longo dos testes. Vale ressaltar que ao longo de todo o processo de validação conseguiu-se reduzir o número de dimensões dos padrões para, em média, 43.

As principais componentes extraídas da base de treinamento foram utilizadas para projetar os exemplos desta base em seu espaço vetorial de dimensões reduzidas. Os vetores projetados foram apresentados ao classificador, o qual consistiu numa SVM (\textit{Support Vector Machine}). Para isso, foi utilizada uma implementação de SVM também fornecida pela biblioteca Scikit-Learn. As principais configurações realizadas foram relacionadas ao \textit{kernel} considerado e ao ponderamento de cada classe durante o aprendizado. Para a primeira, selecionou-se o \textit{kernel} RBF (\textit{Radial Basis Function}) \cite{Chih-WeiHsuChih-ChungChang2008}). Através deste, a SVM é capaz de mapear de modo não-linear os vetores de características para espaços com dimensões mais altas, permitindo assim que sejam tratados casos nos quais a fronteira de separação dos padrões não pode ser linear. Já a segunda configuração consistiu na utilização do parâmetro \textit{balanced} para utilizar o recurso de ponderação balanceada fornecido pela implementação da Scikit-Learn. Esta configuração foi feita em virtude do desbalanceamento da base de dados utilizada, um vez que a mesma apresenta aproximadamente 5 vezes mais exemplos da classe negativa do que da classe positiva. Este recurso altera o nível de contribuição de cada exemplo para o aprendizado da SVM, de modo que os exemplos da classe minoritária têm seu nível aumentado como forma de compensação. Por fim, os demais parâmetros disponíveis foram configurados com seus valores padrões.

Ao final do processo de treinamento, foram salvos os vetores correspondentes às principais componentes extraídas a partir da PCA e os vetores de suporte definidos durante o treinamento da SVM. O objetivo foi disponibilizá-los para a execução da rotina de classificação.

\subsection{\textit{Baseline}}

A metodologia empregada para avaliar o sistema com base no classificador SVM também foi utilizada considerando-se outros dois modelos de aprendizado, cujas implementações também foram fornecidas pela biblioteca Scikit-Learn: um classificador \textit{Perceptron} \cite{Rosenblatt1958} e um SVR (\textit{Support Vector Regressor}) \cite{Drucker1997}. O primeiro consiste num modelo de classificação linear, sendo que seus principais parâmetros de treinamento consistiram em 100 iterações máximas e na ponderação balanceada dos exemplos de cada classe. O objetivo de avaliar-se a performance do sistema com base neste modelo consistiu em utilizar os resultados obtidos como referência para analisarem-se aqueles alcançados através da SVM. Já o segundo consiste num modelo de regressão, a partir do qual é possível estimar a distância associada a cada padrão de fluxo. O objetivo em mensurar-se o desempenho do sistema com base neste modelo consiste na investigação da relevância do processo de normalização dos vetores de características. Sendo assim, para utilizá-lo, foram apresentados os padrões não normalizados, juntamente com as medidas originais obtidas através do sensor de ultrassom. Após treinado e durante os testes, as medidas estimadas por este modelo foram convertidas em rótulos, de maneira análoga ao processo realizado durante a validação com base nos demais modelos. Assim, a partir destes rótulos foi possível realizar a classificação dos padrões. Vale ressaltar que, assim como para a SVM, utilizou-se o SVR com \textit{kernel} RBF.

\subsection{Resultados}
A \autoref{tab:matriz_confusao} apresenta a matriz de confusão do classificador SVM obtida durante cada iteração da validação cruzada.

\begin{table}[htb]
	\IBGEtab{%
		\caption{Matriz de confusão do classificador SVM durante cada iteração da validação cruzada\label{tab:matriz_confusao}}
	}{%
		\begin{tabular}{ccccc}
			\toprule
			\textbf{Iteração (K)} & \textbf{TP} & \textbf{FP} & \textbf{TN} & \textbf{FN} \\
			\midrule
			1 & 610 & 131 & 3827 & 329 \\
			2 & 432 & 213 & 2876 & 250 \\
			3 & 485 & 102 & 3947 & 309 \\
			4 & 501 & 230 & 4060 & 336 \\
			5 & 541 & 144 & 4012 & 266 \\
			6 & 574 & 249 & 3740 & 285 \\
			7 & 435 & 130 & 4045 & 373 \\
			8 & 458 & 133 & 3788 & 349 \\
			\bottomrule
		\end{tabular}
	}{%
		%sem fonte
	}
\end{table}

A partir da \autoref{tab:matriz_confusao}, foram calculadas a precisão, a cobertura, a medida F e a acurácia médias do classificador. Tais medidas são listadas a seguir:

\begin{itemize}
	\item Precisão = $75,46\pm6,21\%$;
	\item Cobertura = $61,71\pm4,75\%$;
	\item Medida F = $68,00\pm3,75\%$;
	\item Acurácia = $89,90\pm1,36\%$.
\end{itemize}

\begin{figure}[htb]
	\centering
	\begin{subfigure}[b]{0.7\linewidth}
		\includegraphics[width=\linewidth]{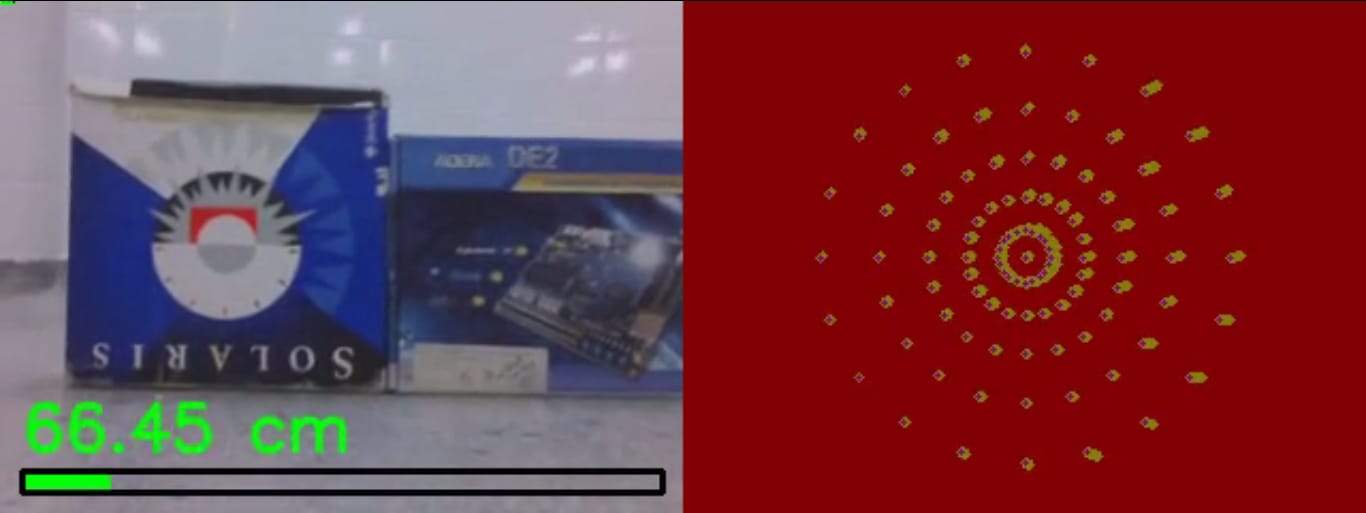}
		\caption{\label{fig:TP}}
	\end{subfigure}
	~
	\begin{subfigure}[b]{0.7\linewidth}
		\includegraphics[width=\linewidth]{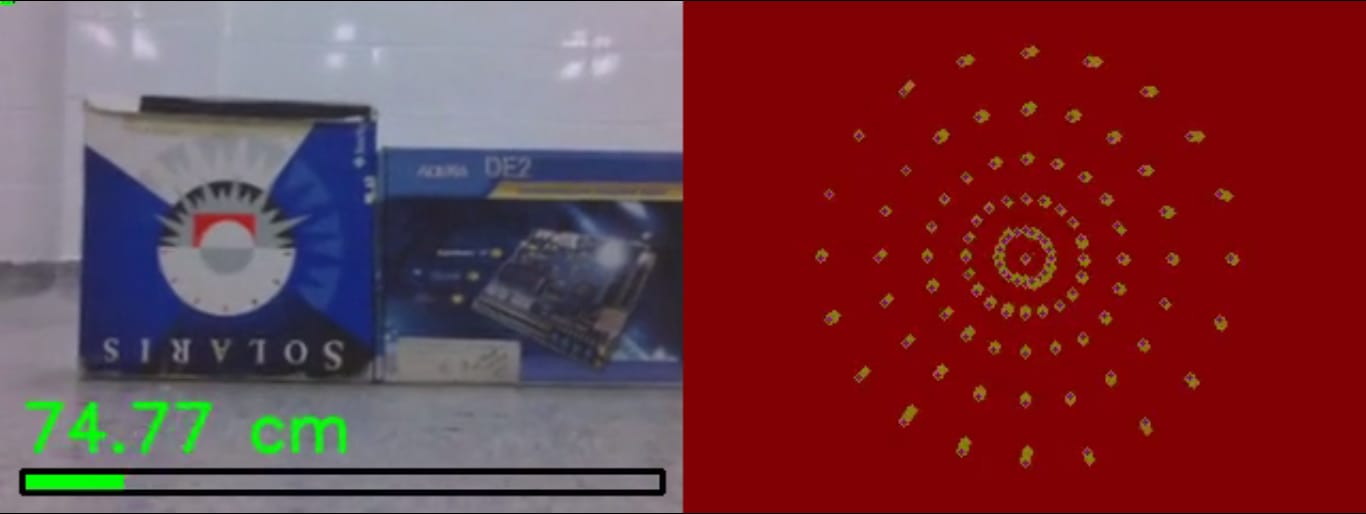}
		\caption{\label{fig:FP}}
	\end{subfigure}
	\caption{Exemplo de padrões de fluxo rotulados pelo classificador. A imagem à esquerda corresponde ao quadro original capturado pelo robô. Sobre o mesmo foi ilustrada a medida obtida pelo sensor ultrassônico naquele instante. Já a imagem à direita ilustra o fluxo óptico estimado. A cor vermelha representa a indicação de obstáculo por parte do sistema. \ref{fig:TP} apresenta um padrão corretamente identificado pelo classificador. Já \ref{fig:FP} ilustra um padrão mal classificado, já que a distância associada encontra-se acima do limiar $L_\textnormal{sup}$.\label{fig:FP_toleravel}}
\end{figure}

Ao analisarem-se as medidas obtidas, percebe-se que o valor da cobertura foi consideravelmente baixo. Isso indica que muitos dos padrões de fluxo associados à movimentação de obstáculos não foram corretamente percebidos pelo classificador. No entanto, vale lembrar que numa situação real é necessário que apenas uma amostra de fluxo seja reconhecida como pertencente a um obstáculo para que seja realizada a manobra de desvio. Ou seja, o valor da cobertura não equivale necessariamente à taxa de colisões do robô. Por sua vez, a precisão do classificador possui maior relevância, já que quanto menor for esta medida maior tende a ser o número de desvios equivocados realizados pelo sistema. Ao analisar-se a precisão média obtida, é possível notar que seu valor foi consideravelmente superior ao da cobertura, estando de acordo com o esperado. Além disso, boa parte dos padrões incorretamente classificados como pertencentes à classe positiva foram capturados poucos centímetros acima do limiar $L_\textnormal{sup}$ considerado para gerar os rótulos, como ilustra a figura \autoref{fig:FP_toleravel}. Assim, entende-se que numa situação real o desvio realizado devido a estas indicações poderia ser tolerado. Finalmente, o valor da acurácia média do classificador encontra-se acima daqueles apresentados pela maioria dos sistemas discutidos no \autoref{cap:trabalhos_relacionados}.

A \autoref{tab:matriz_confusao_baselines} apresenta tanto as medidas de desempenho obtidas pelo classificador SVM quanto aquelas associadas aos demais modelos utilizados como \textit{baseline}. Ao analisá-las, percebe-se que o classificador linear \textit{Perceptron} apresentou o menor valor de acurácia. Assim, pode-se supor inadequada a utilização de uma fronteira linear para separar os padrões de fluxo óptico. Por sua vez, o SVR não foi capaz de reconhecer nenhum exemplo da classe positiva. De fato, durante sua avaliação percebeu-se que todas as medidas estimadas pertenceram ao intervalo $[135cm, 150cm]$, mesmo com os vetores de fluxo não sendo normalizados em etapa alguma da validação. Dessa forma, com base nas medidas associadas a cada modelo torna-se justificável a utilização da SVM com \textit{kernel} RBF somada à normalização do fluxo óptico.

\begin{table}[htb]
	\IBGEtab{%
		\caption{Medidas médias aferidas com base nos diferentes modelos utilizados.\label{tab:matriz_confusao_baselines}}
	}{%
		\begin{tabular}{ccccc}
			\toprule
			Modelo & Precisão & Cobertura & Medida F & Acurácia \\
			\midrule
			SVM & $75,46\pm6,21\%$  & $61,71\pm4,75\%$ & $68,00\pm3,75\%$ & $89,90\pm1,36\%$\\
			\textit{Perceptron} & $16,37\pm2,36\%$  & $45,92\pm9,89\%$ & $24,08\pm3,81\%$ & $50,83\pm3,09\%$\\
			SVR & -  & $0,00\pm0,00\%$ & - & $82,84\pm1,08\%$\\
			\bottomrule
		\end{tabular}
	}{%
		%sem fonte
	}
\end{table}

Por sua vez, a \autoref{tab:matriz_tempo} apresenta o tempo médio levado por cada etapa da classificação durante as iterações da validação cruzada. $T_{op}$ corresponde ao tempo médio levado para se estimar o fluxo óptico, $T_{pca}$ é o tempo médio utilizado para a projeção do vetor de características através da PCA e $T_{svm}$ é o tempo médio gasto pela SVM para classificar o padrão de fluxo.

\begin{table}[htb]
\IBGEtab{%
	\caption{Tempo médio de processamento utilizado para cada etapa da classificação durante a validação cruzada.\label{tab:matriz_tempo}}
}{%
	\begin{tabular}{cccc}
		\toprule
		\textbf{Iteração (K)} & \textbf{$\mathbf{T_{op}}$ (ms)} & \textbf{$\mathbf{T_{pca}}$ (ms)} & \textbf{$\mathbf{T_{svm}}$ (ms)} \\
		\midrule
		1 & 53,66 & 0,76 & 17,16 \\
		2 & 51,99 & 0,74 & 16,76 \\
		3 & 48,93 & 0,70 & 15,72 \\
		4 & 50,53 & 0,72 & 15,77 \\
		5 & 50,58 & 0,72 & 16,04 \\
		6 & 50,53 & 0,71 & 15,33 \\
		7 & 49,01 & 0,68 & 15,23 \\
		8 & 49,08 & 0,70 & 15,70 \\
		\bottomrule
	\end{tabular}
}{%
	%sem fonte
}
\end{table}

A partir da \autoref{tab:matriz_tempo}, foi calculado o tempo médio levado pelo sistema para executar cada etapa da classificação. Tais valores são listados a seguir:

\begin{itemize}
	\item Tempo para estimação do fluxo óptico: $50,50\pm1,64ms$;
	\item Tempo para projeção através da PCA: $0,72\pm0,02ms$;
	\item Tempo para previsão da SVM: $15,97\pm0,66ms$.
\end{itemize}

Ao analisarem-se tais medidas, nota-se que o processo mais custoso corresponde à estimação do fluxo óptico. No entanto, ao somar-se o tempo gasto com todas as etapas da classificação, percebe-se que a frequência de processamento do sistema, sem contar com o tempo de captura das imagens, é igual a 14,88FPS. No entanto, ao considerar-se a frequência média de captura do sistema como sendo igual a 25,28FPS (\autoref{ap:avaliacao_rasp}), chega-se à conclusão de que sua taxa média total de processamento corresponde a 9,4FPS. Este valor pode ser considerado satisfatório, uma vez que é consideravelmente maior que aqueles apresentados pelos trabalhos citados no \autoref{cap:trabalhos_relacionados}. Além disso, vale ressaltar que a estimação da frequência de captura foi realizada com base em imagens de tamanho $640x480$ \textit{pixels}. Como a avaliação do classificador foi realizada sobre imagens cujos tamanhos correspondem à metade deste valor, pode-se supor que numa situação real a taxa de captura seja ainda maior. Logo, pode-se estimar que a frequência de processamento média do sistema desenvolvido é maior que 9,4FPS.

Sendo assim, a \autoref{tab:analise_sistema_vs_trabalhos_relacionados} apresenta uma comparação final entre o sistema desenvolvido e aqueles propostos pelos trabalhos citados no \autoref{cap:trabalhos_relacionados}. Como pode-se perceber, o sistema de navegação desenvolvido neste trabalho apresentou a maior frequência de processamento. Além disso, seu custo final é superior a apenas um dos sistemas citados. Finalmente, o valor de sua acurácia média encontra-se acima dos apresentados pela maioria dos demais sistemas. Sendo assim, com base nos parâmetros comparativos apresentados, pode-se concluir que o sistema de navegação desenvolvido é de baixo custo, apresenta alta frequência de processamento e possui acurácia satisfatória.

\begin{table}[h]
	\IBGEtab{%
		\caption{Análise comparativa entre o sistema desenvolvido e os trabalhos relacionados citados no \autoref{cap:trabalhos_relacionados}.\label{tab:analise_sistema_vs_trabalhos_relacionados}}
	}{%
		\begin{tabular}{cccc}
			\toprule
			\textbf{Sistema} & \textbf{Acurácia} & \textbf{\begin{tabular}[c]{@{}c@{}}Frames \\ Por Segundo\end{tabular}} & \textbf{\begin{tabular}[c]{@{}c@{}}Custo da \\ Plataforma (US\$)\end{tabular}} \\
			\midrule
			\begin{tabular}[c]{@{}c@{}}Detecção de piso por homografia \\\cite{Conrad2010}\end{tabular} & 99,60\% & - & 7142,82 \\
			\begin{tabular}[c]{@{}c@{}}Detecção de piso por segmentação de linhas \\\cite{Li2010}\end{tabular} & 89,10\% & 5 & 10612,82 \\
			\begin{tabular}[c]{@{}c@{}}Segmentação de fluxo óptico \\\cite{Caldeira2007}\end{tabular} & - & 7,41 & 4080,00 \\
			\begin{tabular}[c]{@{}c@{}}Tempo de contato \\\cite{Sanchez-Garcia2015}\end{tabular} & - & - & - \\
			\begin{tabular}[c]{@{}c@{}}Classificação de fluxo óptico \\\cite{Shankar2014}\end{tabular} & 88,80\% & 7 & 50\\
			\begin{tabular}[c]{@{}c@{}}\textbf{Desenvolvido}\end{tabular} & \textbf{89,90\%} & \textbf{9,4} & \textbf{170,82} \\
			\bottomrule
		\end{tabular}
	}{%
		%sem fonte
	}
\end{table}

\section{Avaliação \textit{Online}}

\begin{figure}[htb]
	\centerline{\includegraphics[width=0.7\linewidth]{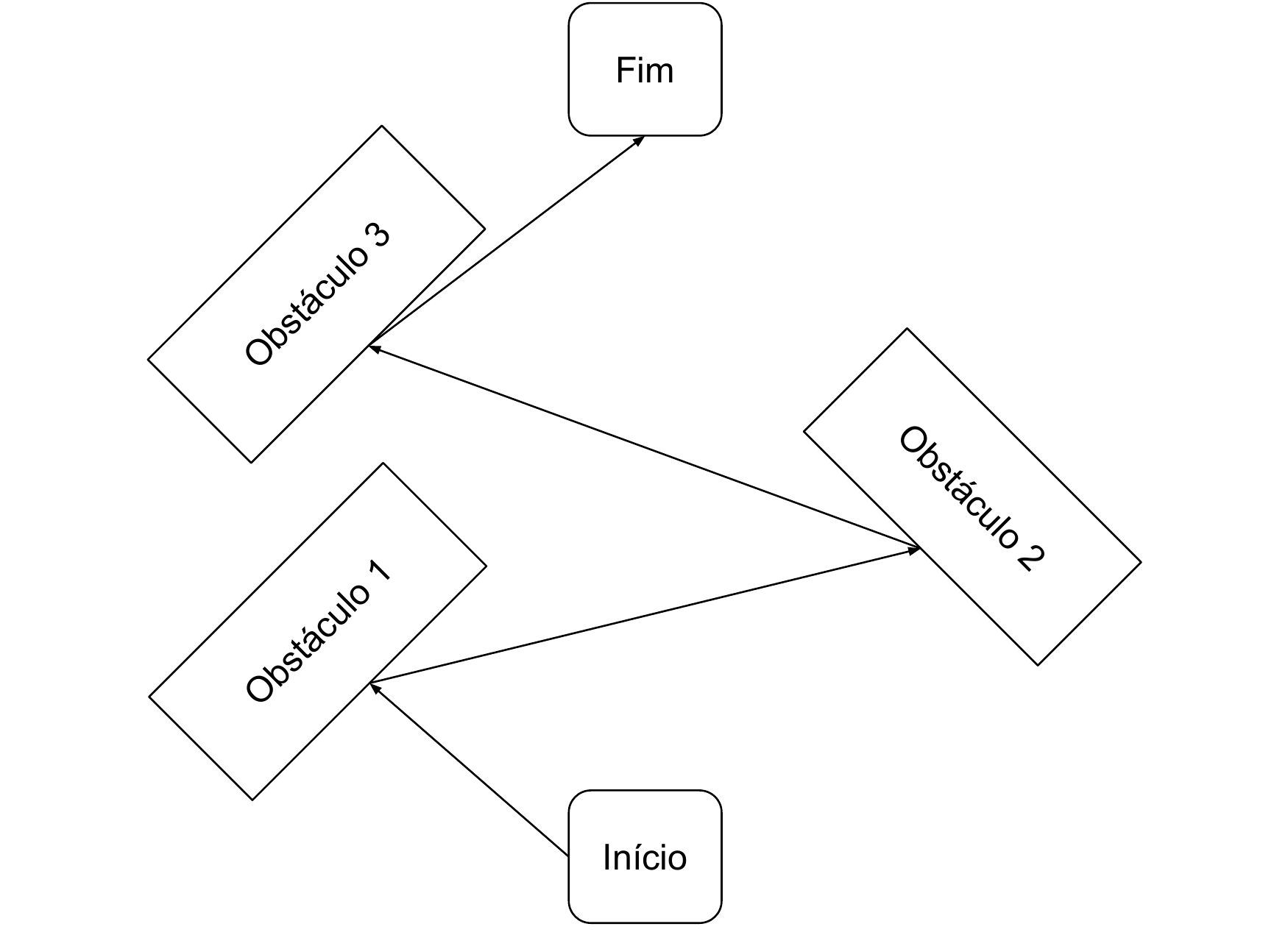}}
	\caption{Ilustração do circuito de teste \textit{online}.\label{fig:circuito_teste_online}}
\end{figure}

\begin{figure}[htb]
	\centerline{\includegraphics[width=0.7\linewidth]{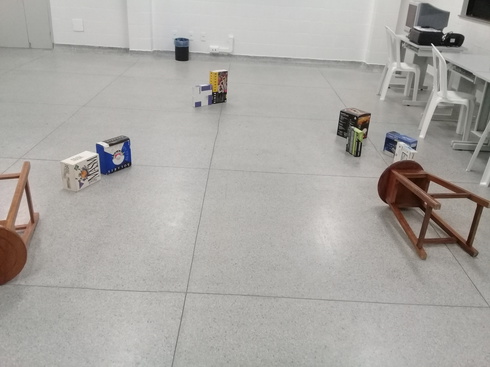}}
	\caption{Ambiente de navegação utilizado para a avaliação \textit{online}.\label{fig:ambiente_teste_online}}
\end{figure}

A avaliação \textit{online} da plataforma consistiu em inserir-se o robô num determinado ambiente e observá-lo navegar de maneira autônoma. Para isso, considerou-se o circuito ilustrado na \autoref{fig:circuito_teste_online}. Como pode-se perceber, no mesmo foram distribuídos três obstáculos diferentes. Através do arranjo considerado, os desvios realizados pelo robô o direcionariam ao longo de todo o circuito da seguinte forma: inicialmente o mesmo já encontraria-se posicionado na direção do primeiro obstáculo. Ao aproximar-se deste, o desvio esperado deveria direcioná-lo para o segundo obstáculo. Uma vez próximo do mesmo, um novo desvio deveria ser capaz de posicioná-lo na frente do terceiro obstáculo. Finalmente, ao desviar deste último, o robô alcançaria o fim do circuito. A \autoref{fig:ambiente_teste_online} apresenta uma imagem do ambiente de navegação considerado.

\begin{figure}[htb]
	\centering
	\begin{subfigure}[b]{0.3\linewidth}
		\includegraphics[width=\linewidth]{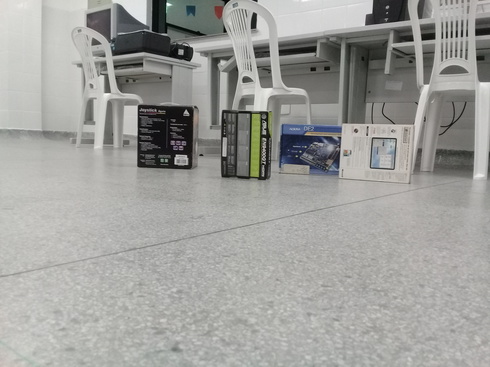}
		\caption{Obstáculo 1.\label{fig:obstaculo1_online}}
	\end{subfigure}
	\qquad
	\begin{subfigure}[b]{0.3\linewidth}
		\includegraphics[width=\linewidth]{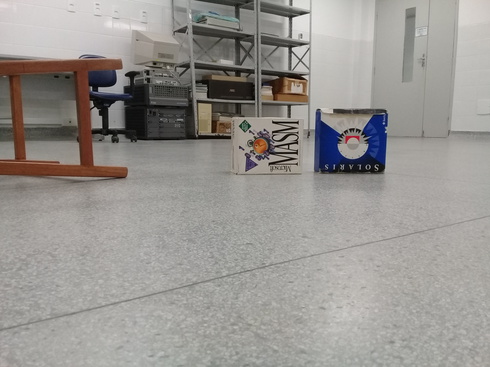}
		\caption{Obstáculo 2.\label{fig:obstaculo2_online}}
	\end{subfigure}
	\qquad
	\begin{subfigure}[b]{0.3\linewidth}
		\includegraphics[width=\linewidth]{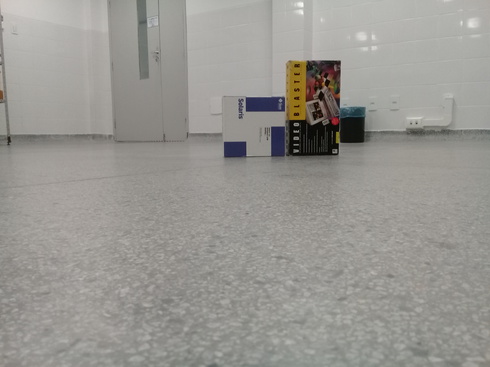}
		\caption{Obstáculo 3.\label{fig:obstaculo3_online}}
	\end{subfigure}
	\caption{Diferentes obstáculos inseridos no circuito de teste \textit{online}.\label{fig:obstaculos_online}}
\end{figure}

Para que o robô fosse capaz de reconhecer os obstáculos distribuídos no circuito, embarcou-se na plataforma uma SVM treinada a partir de todos os padrões presentes na base criada durante o experimento \textit{offline} (\autoref{sec:avaliacao_offline}). Além disso, tal máquina foi treinada de forma idêntica à empregada naquele experimento. Quanto às configurações do robô, consideraram-se a mesma velocidade linear e a mesma angulação da câmera utilizados durante a captura dos padrões de treino. Finalmente, é importante destacar que alguns dos obstáculos contidos no circuito de teste \textit{online} não foram utilizados para a formação da base de treinamento. Dessa forma, o robô deveria ser capaz de generalizar o conhecimento adquirido naquela fase. A \autoref{fig:obstaculos_online} apresenta os obstáculos considerados.

\begin{figure}[htb]
	\centering
	\begin{subfigure}[b]{0.4\linewidth}
		\includegraphics[width=\linewidth]{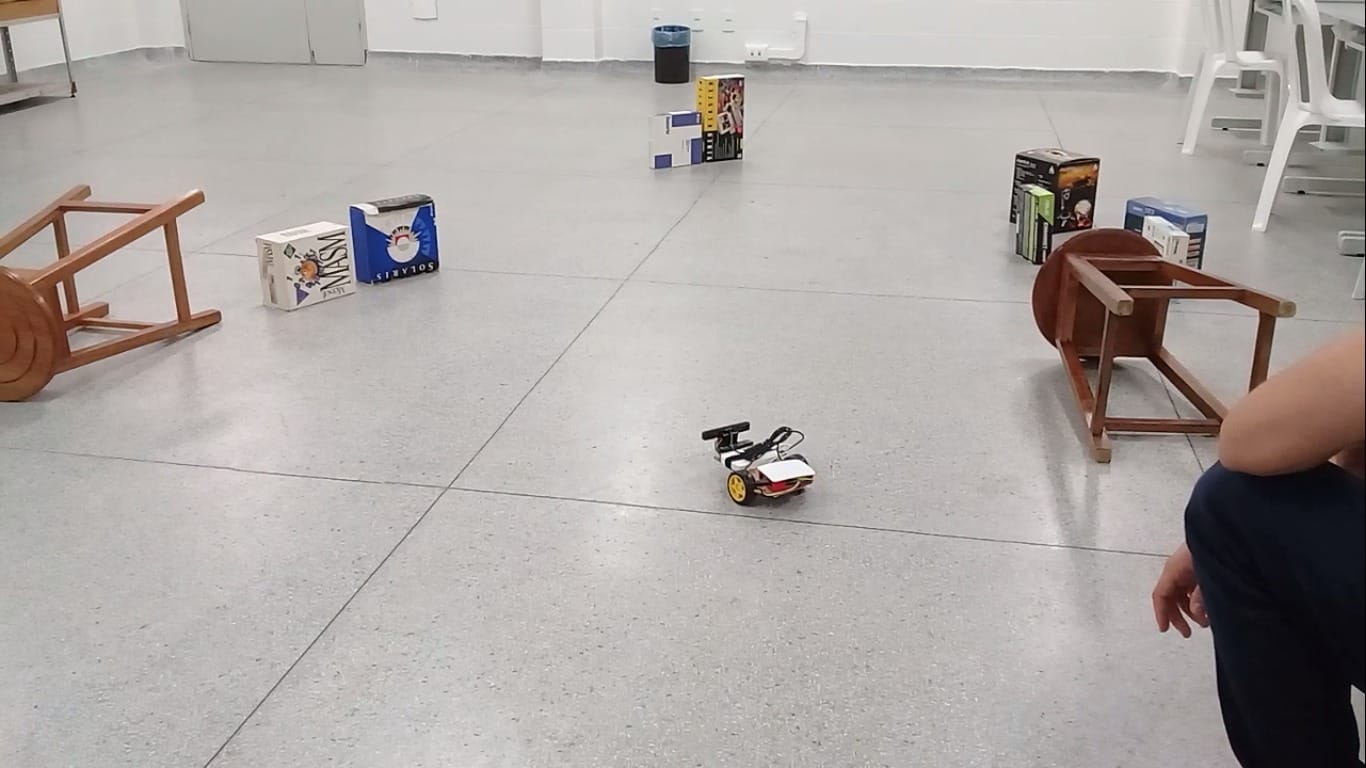}
		\caption{\label{fig:teste_online_cena_1}}
	\end{subfigure}
	\qquad
	\begin{subfigure}[b]{0.4\linewidth}
		\includegraphics[width=\linewidth]{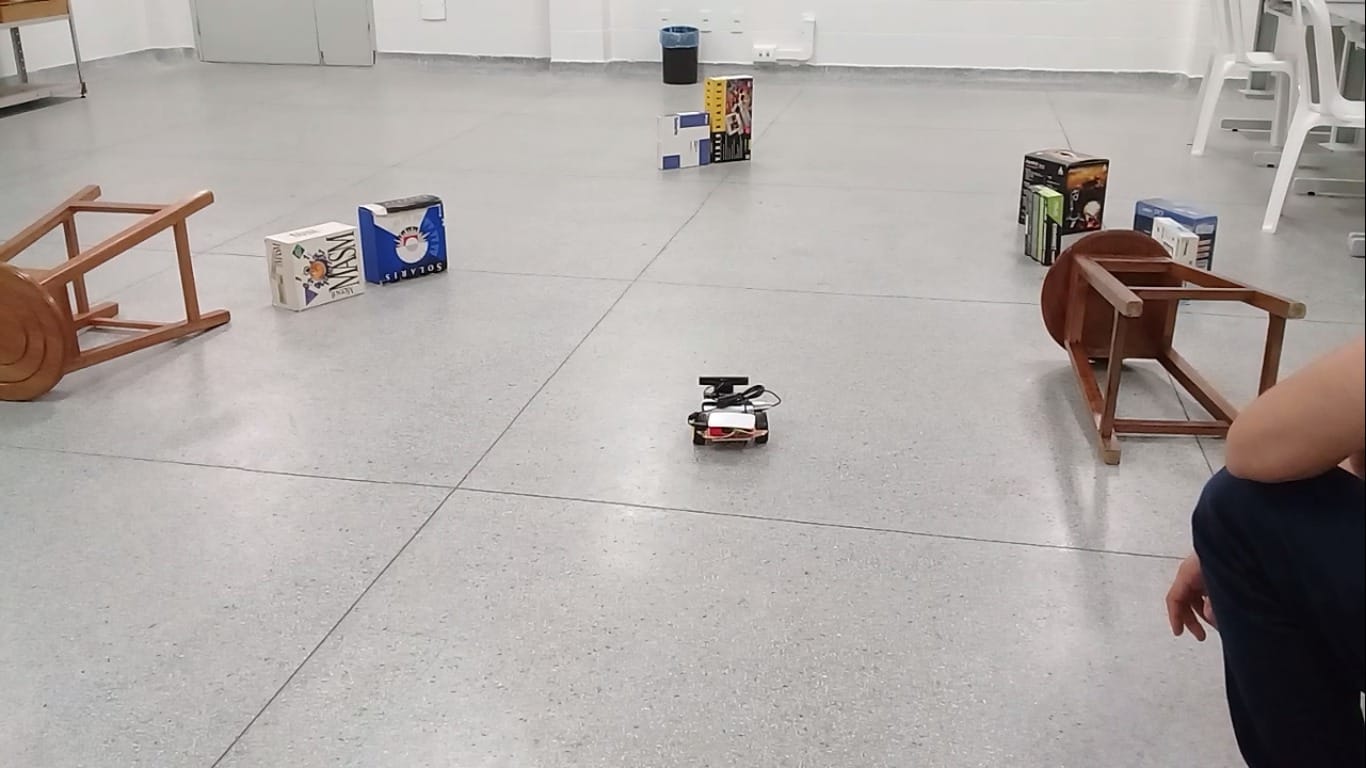}
		\caption{\label{fig:teste_online_cena_2}}
	\end{subfigure}
	\qquad
	\begin{subfigure}[b]{0.4\linewidth}
		\includegraphics[width=\linewidth]{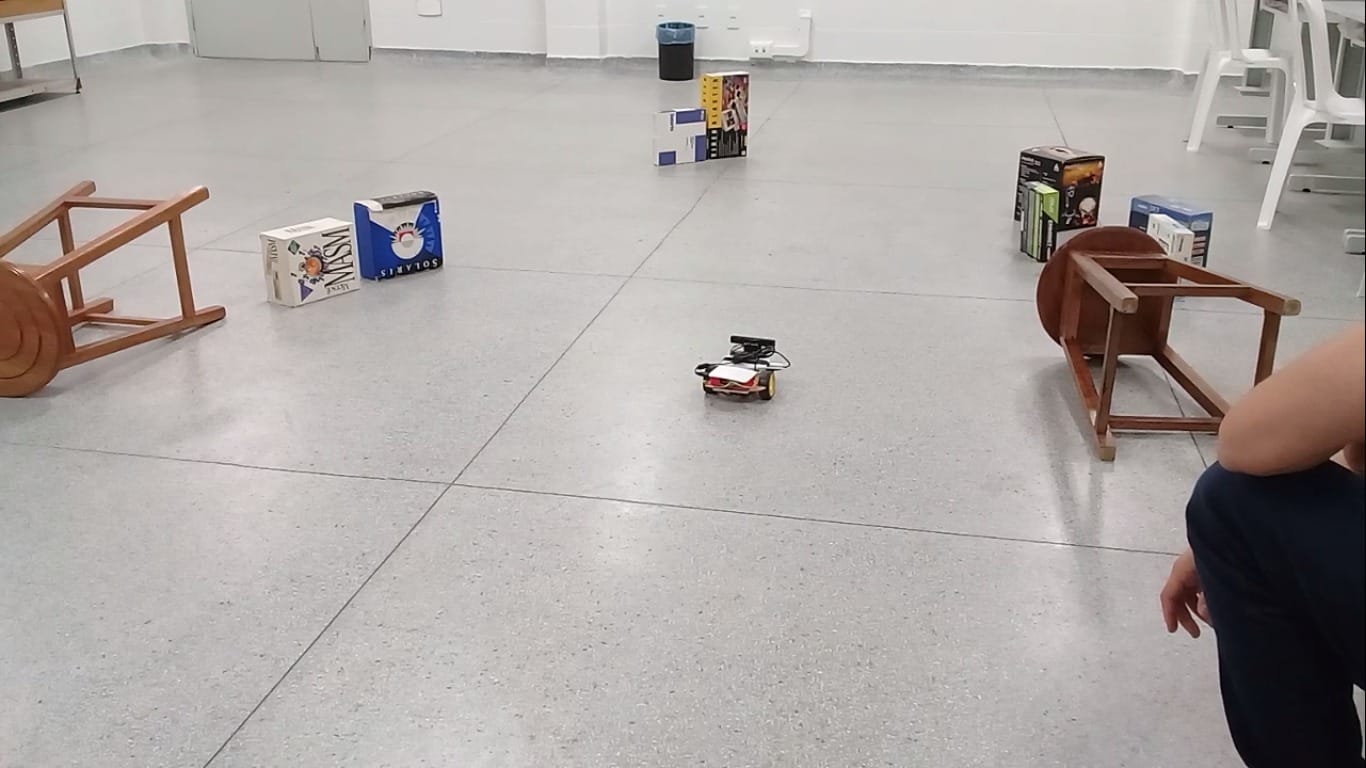}
		\caption{\label{fig:teste_online_cena_3}}
	\end{subfigure}
	\qquad
	\begin{subfigure}[b]{0.4\linewidth}
		\includegraphics[width=\linewidth]{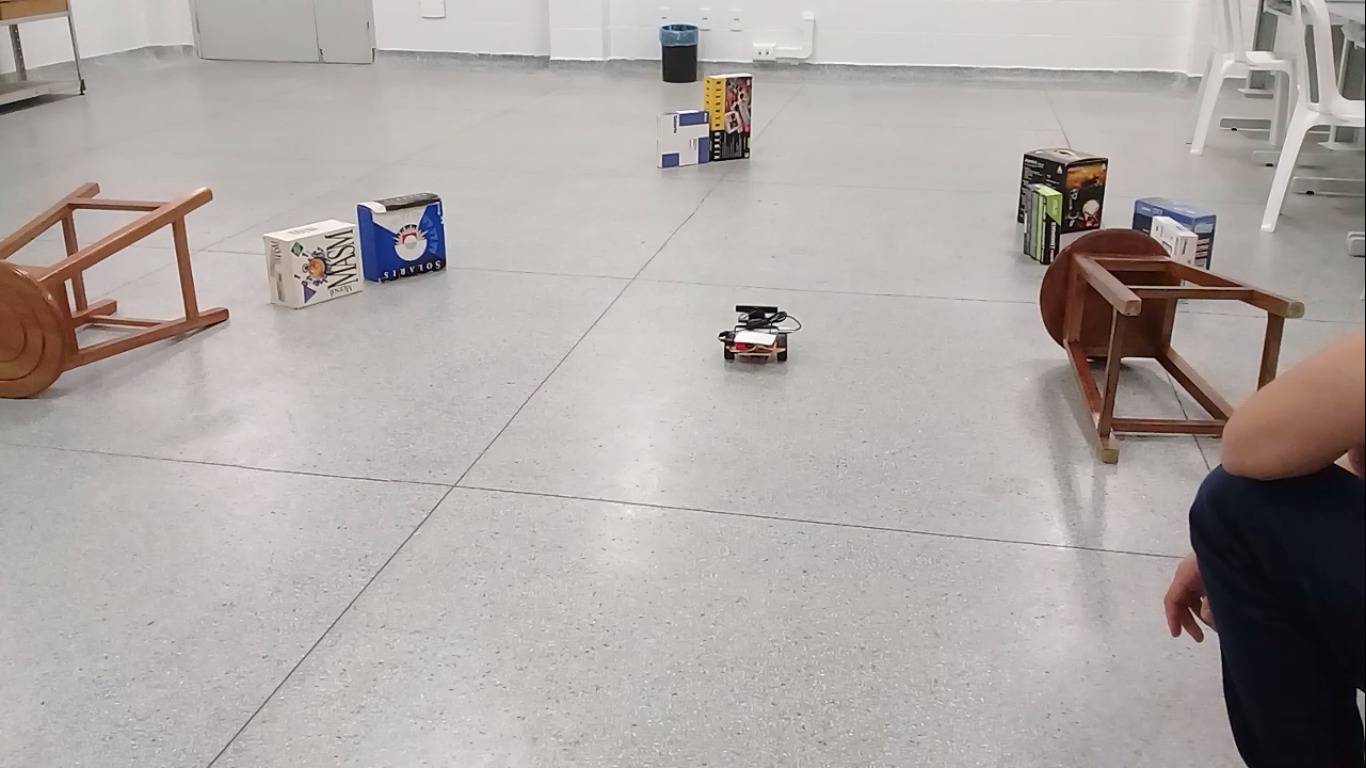}
		\caption{\label{fig:teste_online_cena_4}}
	\end{subfigure}
	\qquad
	\begin{subfigure}[b]{0.4\linewidth}
		\includegraphics[width=\linewidth]{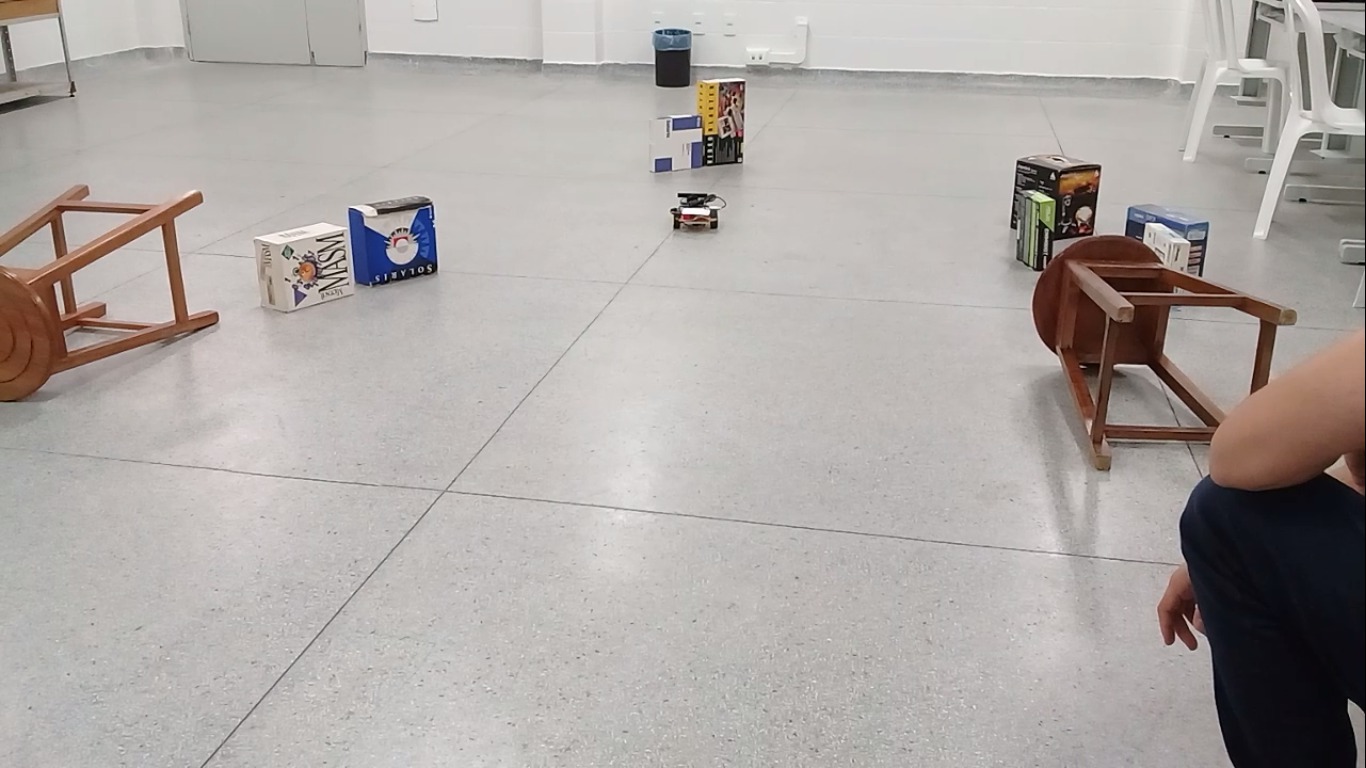}
		\caption{\label{fig:teste_online_cena_5}}
	\end{subfigure}
	\qquad
	\begin{subfigure}[b]{0.4\linewidth}
		\includegraphics[width=\linewidth]{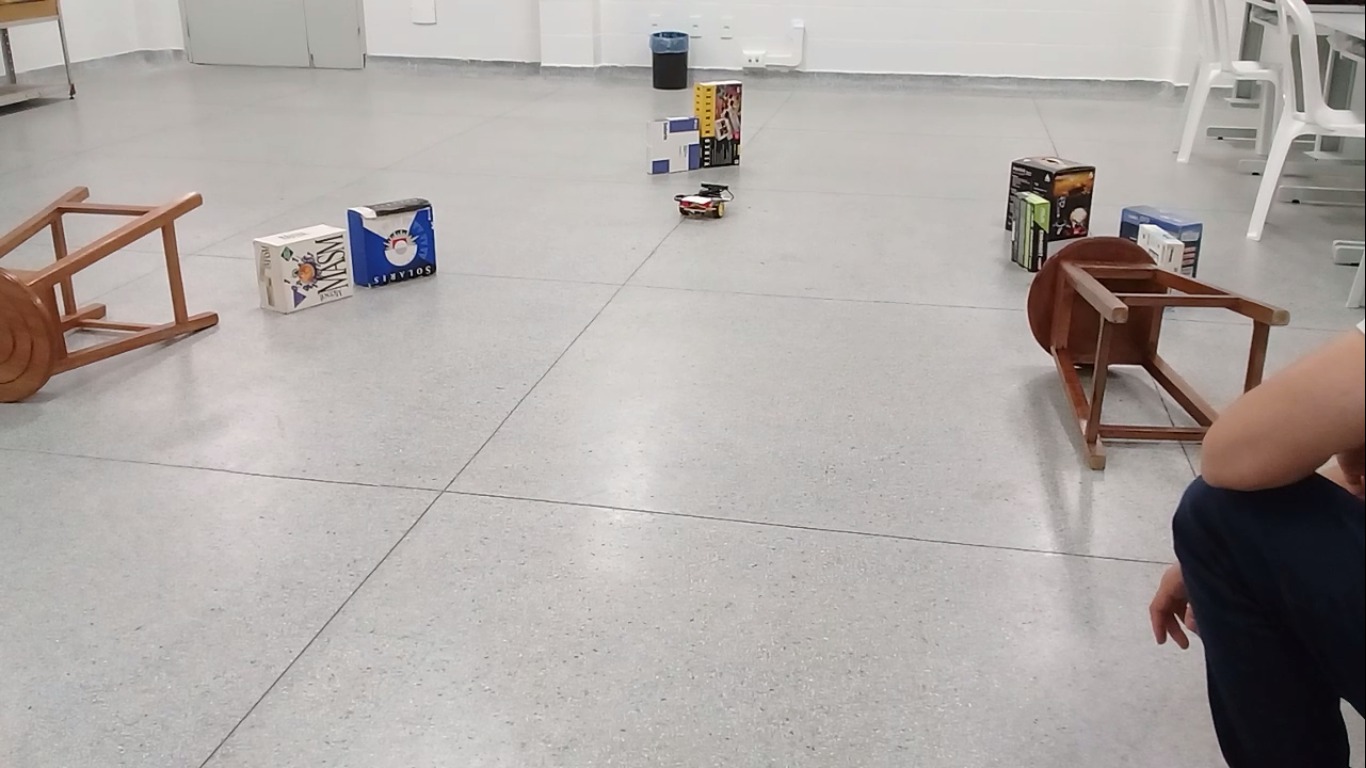}
		\caption{\label{fig:teste_online_cena_6}}
	\end{subfigure}
	\caption{Navegação autônoma do robô ao longo do circuito de teste. \ref{fig:teste_online_cena_1} e \ref{fig:teste_online_cena_2} ilustram a aproximação do robô ao primeiro obstáculo e o seu desvio, respectivamente. Já \ref{fig:teste_online_cena_3} e \ref{fig:teste_online_cena_4} exibem, respectivamente, a aproximação do robô ao segundo obstáculo e o seu desvio. Por fim, \ref{fig:teste_online_cena_5} e \ref{fig:teste_online_cena_6} apresentam a aproximação do robô ao terceiro obstáculo e o seu desvio, respectivamente. \label{fig:avaliacao_online}}
\end{figure}

O vídeo disponibilizado em \url{http://www.youtube.com/v/hzyKAGhQExg?rel=0} apresenta a navegação do robô ao longo do circuito de teste. Já a \autoref{fig:avaliacao_online} exibe algumas cenas desta gravação. Ao analisá-la, é possível notar que o robô desviou dos dois primeiros obstáculos antes de aproximar-se mais do que $70cm$. No entanto, os desvios realizados foram válidos, já que o robô encontrava-se em rota de colisão com os mesmos. Sendo assim, pode-se destacar que em nenhum momento o robô colidiu com os obstáculos contidos no circuito. Além disso, observa-se que todos os desvios foram realizados para a direção esperada. Finalmente, percebe-se que o controle dos atuadores permitiu a realização de curvas precisas, através das quais o robô pôde se posicionar na direção de todos os obstáculos subsequentes.

%% file: Conteudo/Conclusao.tex
\chapter{Conclusão}
\label{cap:conclusao}
Durante este trabalho foi possível investigar a utilização de técnicas de visão computacional para o desenvolvimento de um sistema robótico de baixo custo, capaz de navegar de maneira autônoma e baseado na plataforma Raspberry Pi. Mais especificamente, puderam-se revisar diferentes abordagens, propostas em trabalhos recentes, relacionadas à construção de sistemas autônomos cuja única forma de sensoriamento corresponde à visão monocular. A partir desta revisão, percebeu-se a maior adequação do fluxo óptico como característica primária a ser considerada por sistemas de navegação que devem apresentar baixo custo computacional. Sendo assim, explorou-se a definição de fluxo óptico e avaliou-se sua utilização para a tarefa de reconhecimento de obstáculos. Com base na investigação realizada, desenvolveu-se um robô autônomo capaz de identificar obstáculos a partir da classificação de fluxo óptico. O processo de desenvolvimento foi composto por duas etapas: a construção do \textit{hardware} e a implementação do \textit{software}. Durante a primeira etapa, realizou-se a seleção e a integração dos diferentes componentes de \textit{hardware} utilizados para a construção da plataforma física. Por sua vez, a segunda etapa consistiu em, inicialmente, configurar e a avaliar a placa Raspberry Pi. Em seguida, implementou-se o sistema de navegação cuja estimação de fluxo óptico baseia-se no algoritmo de Lucas-Kanade e cuja classificação é realizada através de uma SVM.

O sistema desenvolvido foi avaliado com base na sua performance \textit{offline} e \textit{online}. Para o primeiro tipo de avaliação, implementou-se uma estratégia automática de coleta de padrões de fluxo óptico já rotulados, com o propósito de construir-se uma base de dados contendo amostras de cenários reais. Em seguida, utilizou-se o método de validação cruzada \textit{k-fold} para treinar o classificador a partir de um determinado conjunto de elementos contidos na base e aplicá-lo sobre o grupo restante, de forma iterativa e exclusiva. Durante cada iteração, foram coletados o tempo exigido pelo sistema para executar cada etapa da classificação e a matriz de confusão do classificador. Ao final, os parâmetros médios obtidos foram utilizados para comparar a qualidade do sistema desenvolvido com a dos propostos pelos trabalhos relacionados. Ao fim da comparação, constatou-se que o sistema implementado apresentou acurácia superior e custo de aquisição inferior ao da maioria dos trabalhos citados. Além disso, sua frequência de processamento superou as apresentadas pelos sistemas comparados. Já durante a avaliação \textit{online}, o sistema desenvolvido foi capaz de identificar e de desviar corretamente de todos os obstáculos inseridos no ambiente percorrido, sendo que parte destes elementos não foram apresentados ao robô durante seu treinamento. Sendo assim, ao fim das avaliações concluiu-se que a plataforma robótica construída atingiu os objetivos definidos.

Para trabalhos futuros, pretende-se explorar a utilização de modelos de classificação probabilísticos, através dos quais seja possível desenvolver um sistema de navegação capaz de prever obstáculos com base em padrões de fluxo óptico estimados no passado \cite{Lipton2015}. Além disso, deseja-se experimentar novas estratégias para a extração de características do fluxo estimado sobre uma cena, de modo a reduzir-se a dimensão dos padrões apresentados ao classificador. Finalmente, pretende-se desenvolver um protocolo de teste que permita aferir a capacidade de generalização do sistema em contraste à apresentada por aqueles baseados em sensores de alcance.

%% file: Pos_Textual/Apendices.tex
\begin{apendicesenv}

% Imprime uma página indicando o início dos apêndices
\partapendices

% ----------------------------------------------------------
\chapter{Configuração da Raspberry Pi para o Desenvolvimento do Sistema de Navegação}
\label{ap:config_rasp}
% ----------------------------------------------------------

Inicialmente, realizou-se, num computador à parte, o \textit{download} da imagem do sistema operacional Raspbian \cite{Raspbian}, o qual é baseado no Debian. Apesar de haver suporte para vários outros sistemas operacionais (baseados ou não em Linux), optou-se pela utilização do Raspbian pelo fato deste ser o sistema oficial da Raspberry Pi. Em seguida, a imagem obtida foi gravada num cartão Micro SD de 32GB e 10MB/s (\autoref{fig:cartao_sd}) através do \textit{software} Win32 Disk Imager (\autoref{fig:win32}). Após a gravação, este cartão foi inserido na Raspberry Pi, a qual foi conectada a uma fonte de energia, a um \textit{mouse} e a um teclado com interfaces USB, a uma rede local por meio de conexão \textit{Ethernet} e a um monitor com interface HDMI. Ao ligar a Raspberry Pi, realizou-se o \textit{login} no sistema com as credenciais padrões fornecidas pelo fabricante e executou-se a ferramenta de configuração da plataforma. Através desta, permitiu-se que o sistema operacional ocupasse todo o espaço fornecido pelo cartão SD. Além disso, configurou-se o sistema para sempre iniciar a partir da linha de comando. Finalmente, realizou-se a atualização do seu \textit{firmware}.

\begin{figure}[htb]
	\centering
	\includegraphics[width=0.4\linewidth]{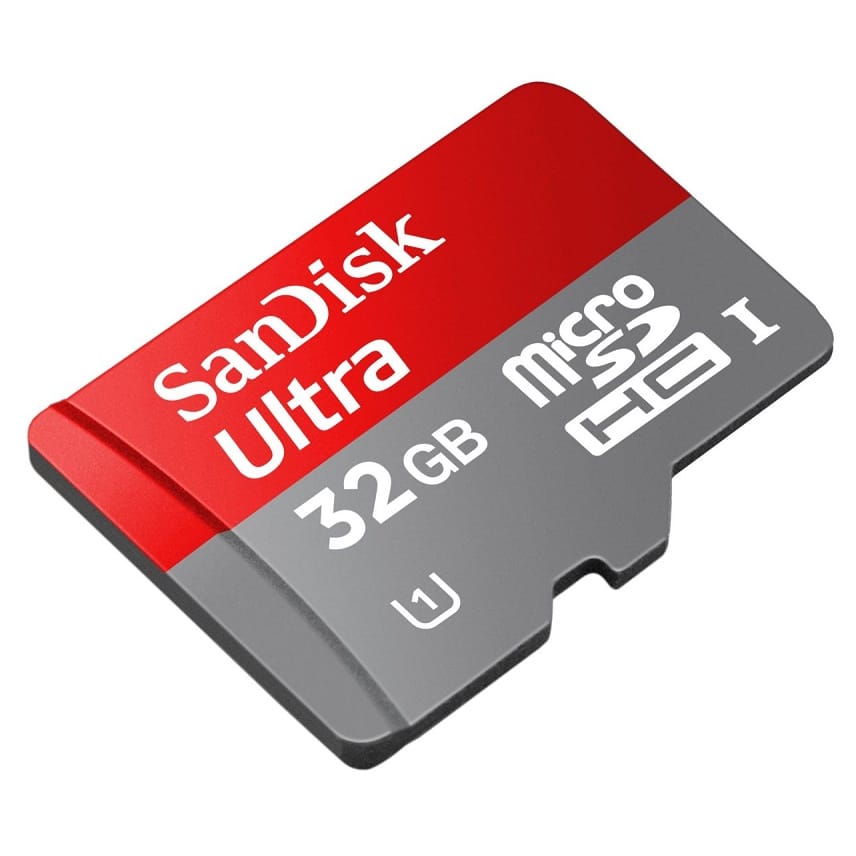}
	\caption{Cartão SD no qual foi gravada a imagem do Raspbian \cite{SanDisk2017}.\label{fig:cartao_sd}}
\end{figure}

\begin{figure}[htb]
	\centering
	\includegraphics[width=0.5\linewidth]{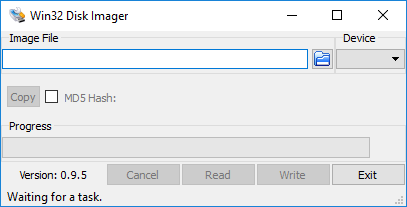}
	\caption{\textit{Software} Win32 Disk Imager utilizado para gravar a imagem do Raspbian no cartão SD \cite{Sourceforge2017}.\label{fig:win32}}
\end{figure}

Após configurar o sistema operacional da Raspberry Pi, foi preciso conectá-la a uma rede \textit{Wi-Fi}. Para isso, executou-se o assistente fornecido através da interface gráfica do sistema (\autoref{fig:raspbian}). Uma vez conectada, removeu-se a conexão \textit{Ethernet}. Em seguida, instalou-se o \textit{software} PuTTY (\autoref{fig:conexao_putty}) num computador à parte, por meio do qual seria possível realizar uma conexão SSH com o servidor embutido no Raspbian e, assim, controlar a Raspberry Pi remotamente. Uma vez estabelecida a conexão, puderam-se remover o \textit{mouse}, o teclado e o monitor até então conectados à Raspberry Pi. Por fim, também instalou-se, neste mesmo computador de controle, o \textit{software} FileZilla, através do qual seria possível transferir arquivos entre tal máquina e o Raspbian por meio do protocolo SFTP.

\begin{figure}[htb]
	\centering
	\includegraphics[width=0.7\linewidth]{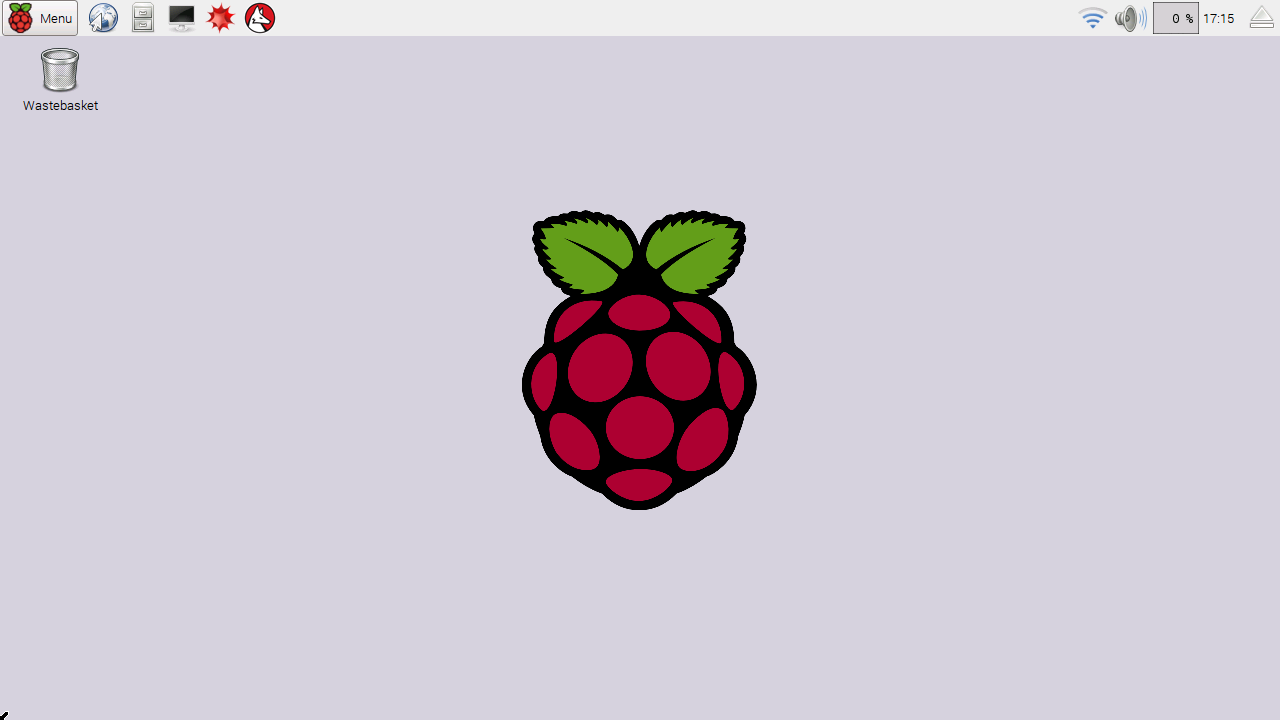}
	\caption{Interface gráfica do Raspbian. No canto superior direito é possível acessar o assistente para configuração de rede.\label{fig:raspbian}}
\end{figure}

\begin{figure}[htb]
	\centering
	\begin{subfigure}[b]{0.7\linewidth}
		\includegraphics[width=\linewidth]{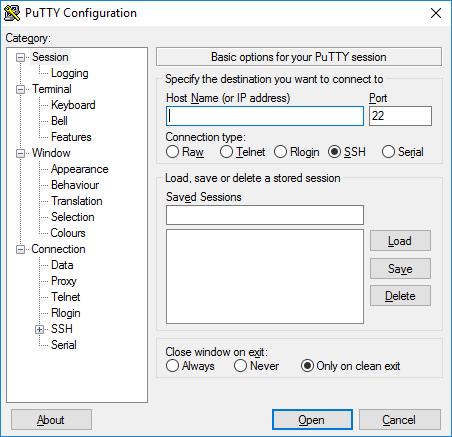}
		\caption{Tela principal do \textit{software} PuTTy.\label{fig:putty}}
	\end{subfigure}
	\\
	\begin{subfigure}[b]{0.7\linewidth}
		\includegraphics[width=\linewidth]{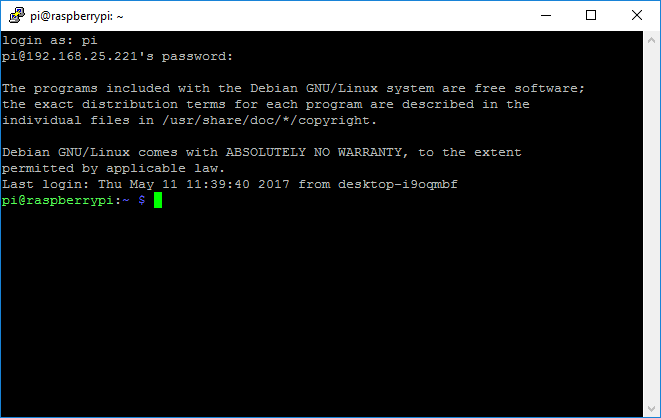}
		\caption{Sessão SSH entre o cliente PuTTy e o Raspbian.\label{fig:sessao_ssh}}
	\end{subfigure}
	\caption{Conexão SSH a partir do \textit{software} PuTTy.\label{fig:conexao_putty}}
\end{figure}

Uma vez realizada a conexão da Raspberry Pi com o computador de controle, foi preciso instalar as ferramentas utilizadas para o desenvolvimento do sistema de navegação a ser executado. A primeira delas foi o Python 2.7 \cite{Foundation2017}, a partir do qual seriam desenvolvidos todos os códigos do sistema. Esta escolha foi feita devido à simplicidade e, consequentemente, à praticidade oferecidas pela linguagem. Além disso, as funções fornecidas pelas bibliotecas a serem utilizadas já se encontrariam compiladas, de modo que os \textit{scripts} em Python apenas realizariam suas chamadas. Sendo assim, o desempenho do sistema não seria prejudicado. 

Com o suporte a Python definido, instalou-se a versão 3.2 da biblioteca multiplataforma de visão computacional e aprendizado de máquina OpenCV \cite{OpenCV3-2}. Esta biblioteca de código aberto contém mais de 2500 algoritmos otimizados, sendo utilizada mundialmente por diversas empresas, grupos de pesquisa e órgãos governamentais \cite{OpenCV}. Para instalá-la, primeiramente foram adicionados alguns pacotes responsáveis por permitir o carregamento de imagens e vídeos de diferentes formatos, tais como JPEG, PNG, AVI, dentre outros. Após, realizou-se o \textit{download} do código fonte da OpenCV. Antes de compilá-lo, foi preciso instalar a biblioteca de processamento numérico Numpy \cite{van_der_Walt_2011}. Finalmente, compilou-se a OpenCV. Dentre os parâmetros de compilação configurados, habilitou-se a utilização do recurso TBB \cite{Intel2017}. A partir do mesmo, permite-se que o código da OpenCV seja executado paralelamente sobre os 4 núcleos de processamento da Raspberry Pi.

Por fim, foi instalado na Raspberry Pi a biblioteca de aprendizado de máquina Scikit-Learn \cite{scikit-learn}, a qual disponibiliza ferramentas relevantes para o desenvolvimento do sistema, como a implementação da PCA \cite{hotelling1933analysis} e da SVM \cite{Chih-WeiHsuChih-ChungChang2008}.

% ----------------------------------------------------------
\chapter{Avaliação da Raspberry Pi 3 Model B}
\label{ap:avaliacao_rasp}
% ----------------------------------------------------------

Para avaliar o desempenho da Raspberry Pi 3 Model B e analisar sua viabilidade para este projeto, foram elaborados testes de desempenho baseados na frequência de captura de imagens através da câmera e na execução de algoritmos frequentemente utilizados em aplicações de visão computacional, os quais encontram-se implementados na OpenCV. Os testes realizados consistiram na medição do tempo levado pela Raspberry Pi para executar as seguintes operações:

\begin{itemize}
	\item Capturar imagem;
	\item Calcular DFT;
	\item Aplicar Filtro de Canny;
	\item Estimar fluxo óptico.
\end{itemize}

O primeiro teste consistiu na medição do tempo levado pela plataforma para capturar 120 imagens através da câmera utilizada neste trabalho. O objetivo foi estimar a taxa de captura média da Raspberry Pi. Este teste é relevante pois apesar da câmera em questão ser capaz de capturar até 30 quadros por segundo, a taxa final de imagens capturadas pela Raspberry Pi pode ser influenciada pela latência da conexão USB, pelo escalonamento de processos no sistema, dentre outros fatores. Já o segundo teste consistiu em medir-se o tempo do cálculo da DFT (\textit{Discrete Fourier Transform}) sobre um vídeo contendo 60 quadros. Este experimento foi realizado 50 vezes, sendo que ao final coletou-se o tempo médio levado pela plataforma. Por sua vez, o terceiro teste consistiu em avaliar a performance da Raspberry Pi ao pré-processar um determinado vídeo contendo 480 quadros e aplicar sobre este o Filtro de Canny, o qual corresponde a um filtro passa-alta muito utilizado para a detecção de bordas. Durante o teste, mediu-se o tempo levado para converter cada quadro em níveis de cinza, aplicar um filtro passa-baixa Gaussiano e em seguida o Filtro de Canny. Ao final, armazenou-se o tempo médio de processamento por quadro. Finalmente, o último experimento consistiu na medição do tempo de estimação do fluxo óptico sobre um determinado vídeo contendo 230 imagens. Para isso, utilizou-se o algoritmo de Lucas-Kanade com base numa distribuição de 100 pontos fixos. Ao final, coletou-se o tempo médio gasto por quadro.

Além de executá-los na Raspberry Pi, estes testes também foram realizados num \textit{desktop}, cujas principais especificações são: processador I7 4470K 3,5GHz, memória 16GB 800MHz, armazenamento 2TB 6GB/s e sistema operacional Windows 8.1. Após a execução de cada teste, foi medida a quantidade média de imagens processadas por segundo em cada plataforma, sendo que o tamanho das imagens consideradas correspondeu sempre a $640x480$ \textit{pixels}. Os resultados finais obtidos são apresentados na tabela a seguir:

\begin{table}[htb]
	\IBGEtab{%
		\caption{Resultados dos testes de desempenho\label{tab:resultados_desempenho}}
	}{%
		\begin{tabular}{ccc}
			\toprule
			\textbf{Experimentos}    & \textbf{Média de FPS no Raspberry Pi} & \textbf{Média de FPS no Desktop} \\
			\midrule
			Taxa de captura & 25,28					   & 22,78					  \\
			Cálculo da DFT  & 27,68                    & 451,81               \\
			Filtro de Canny & 29,81                    & 455,34               \\
			Fluxo óptico    & 36,69                    & 472,36               \\
			\bottomrule              
		\end{tabular}
	}{%
		%sem fonte
	}
\end{table}

Ao analisar os resultados apresentados na \autoref{tab:resultados_desempenho}, percebe-se inicialmente que a taxa de captura da Raspberry Pi ficou próxima à máxima permitida pela câmera. Nota-se também que a taxa obtida pelo \textit{desktop} foi inferior, fato este justificado pela possível incompatibilidade do sistema operacional com o formato de codificação utilizado pela câmera. Considerando-se os demais testes, nota-se que o desempenho da Raspberry Pi foi cerca de 15 vezes inferior ao do \textit{desktop}, uma vez que suas especificações de \textit{hardware} são significativamente inferiores. Além disso, é comum que em sistemas autônomos baseados em visão computacional sejam utilizadas câmeras de baixo a médio custo com frequência máxima de captura igual à 30 quadros por segundo. Como a frequência média de processamento da Raspberry Pi encontrou-se próxima deste valor, percebe-se a existência de uma margem razoável de tempo que pode ser preenchida com mais operações sem que haja perda significativa de performance. Sendo assim, com base nos resultados de desempenho obtidos ao longo dos testes, pôde-se concluir que a utilização da Raspberry Pi 3 Model B para o desenvolvimento do sistema proposto neste trabalho é viável.

% ----------------------------------------------------------
\chapter{Seleção dos Pontos Considerados para a Estimação do Fluxo Óptico}
\label{ap:selecao_pontos}

Antes de realizar-se a extração do fluxo óptico, primeiramente deve-se determinar quais pontos da imagem serão monitorados. Esta escolha é relevante pelo fato de que quanto maior a quantidade de pontos, maior será a complexidade computacional envolvida no cálculo do fluxo. Ao passo que quanto menor tal quantidade, menor a resolução do padrão de movimento visualizado. Além disso, a distribuição dos pontos escolhidos também é relevante para o problema de detecção de obstáculos, uma vez que as regiões da imagem com maior concentração de pontos tornam-se mais relevantes para a determinação do padrão de movimento. Ou seja, a partir de distribuições não uniformes de pontos monitorados é possível realçar a contribuição de determinadas regiões da imagem para a indicação da existência de obstáculo.

Para determinar a melhor distribuição de pontos de monitoramento visando o problema de detecção de obstáculos, foram realizados testes visuais com base na aplicação do algoritmo de Lucas-Kanade sobre determinados vídeos de navegação em cenários reais. O objetivo destes testes foi avaliar visualmente se o fluxo obtido evidenciava de forma satisfatória o movimento dos obstáculos contidos no ambiente. A \autoref{fluxo_optico_obtido} apresenta os resultados de alguns desses testes.

\begin{figure}[htb]
	\centering
	\begin{subfigure}[b]{0.4\linewidth}
		\includegraphics[width=\linewidth]{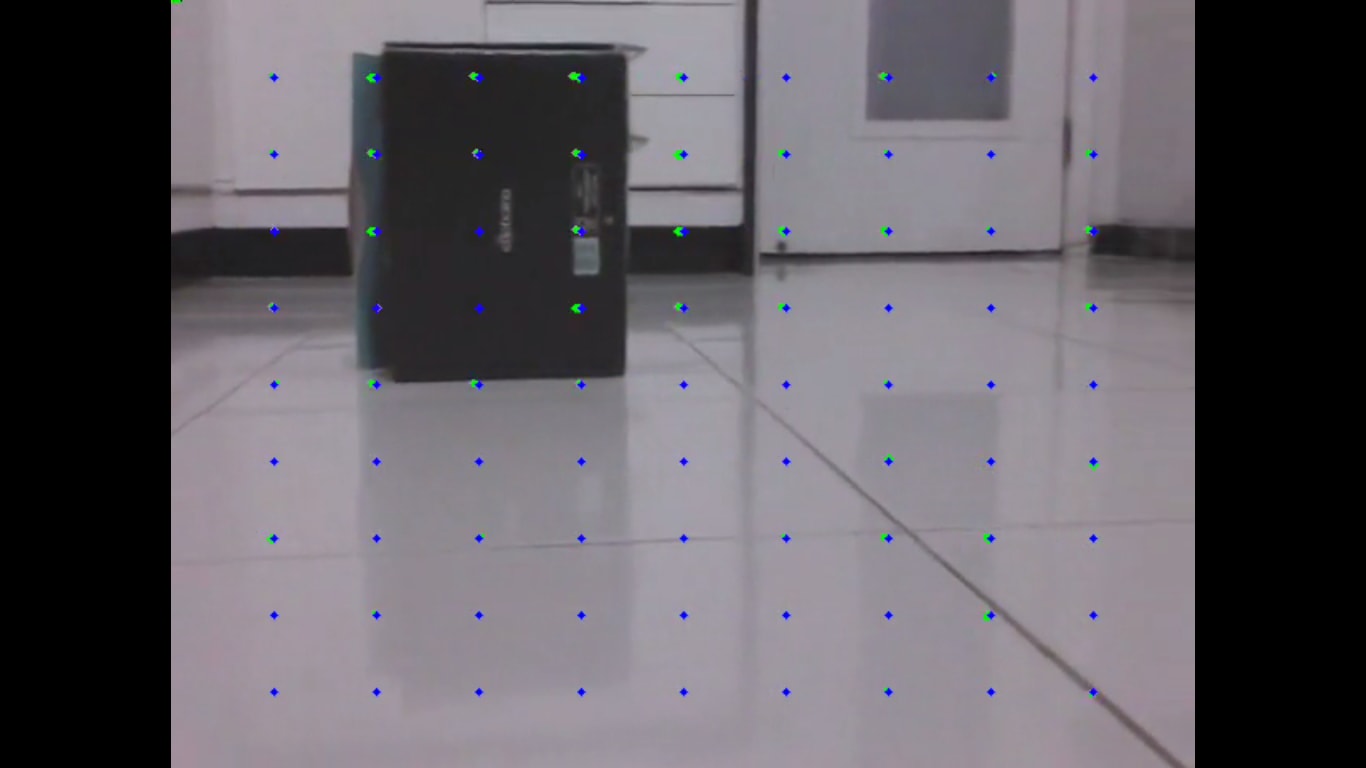}
		\caption{\label{fig:fluxo_pontos_fixos}}
	\end{subfigure}
	~
	\begin{subfigure}[b]{0.4\linewidth}
		\includegraphics[width=\linewidth]{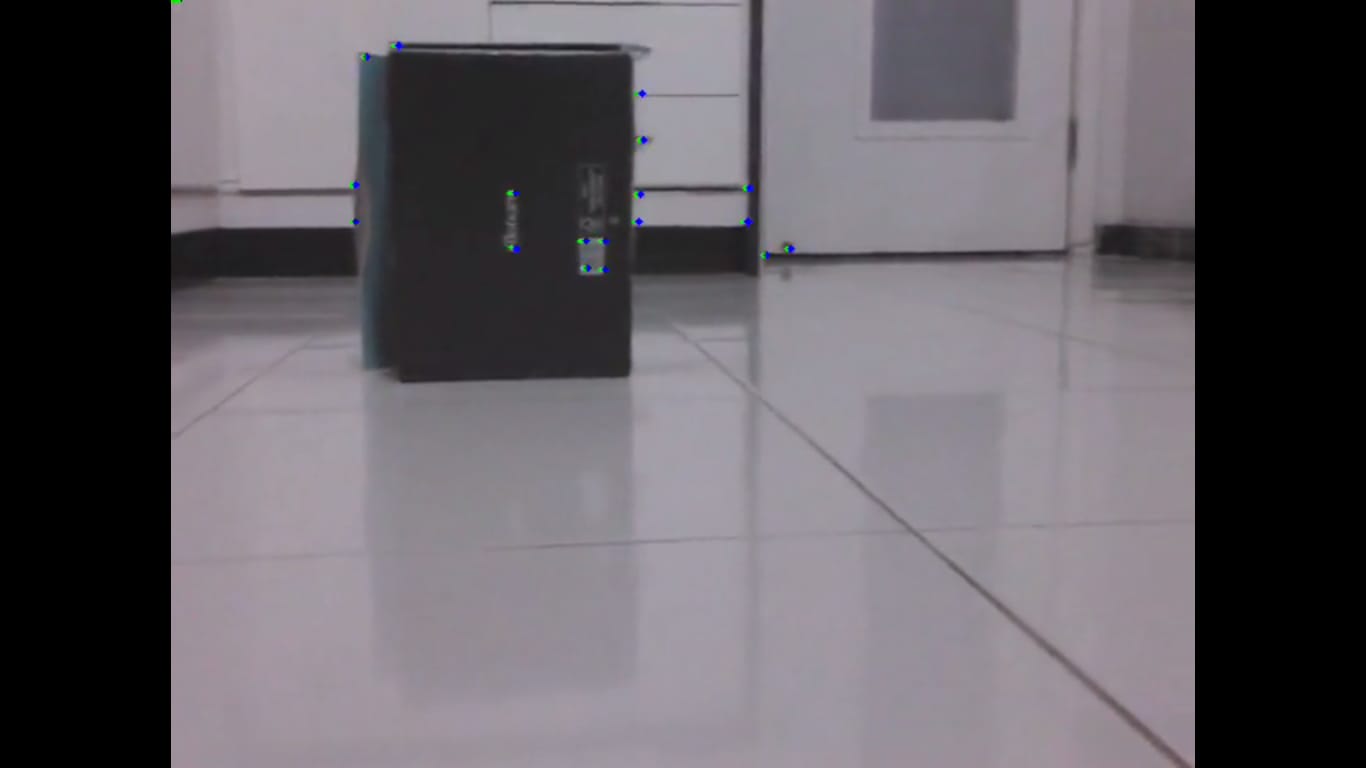}
		\caption{\label{fig:fluxo_melhores_pontos}}
	\end{subfigure}
	\qquad
	\begin{subfigure}[b]{0.4\linewidth}
		\includegraphics[width=\linewidth]{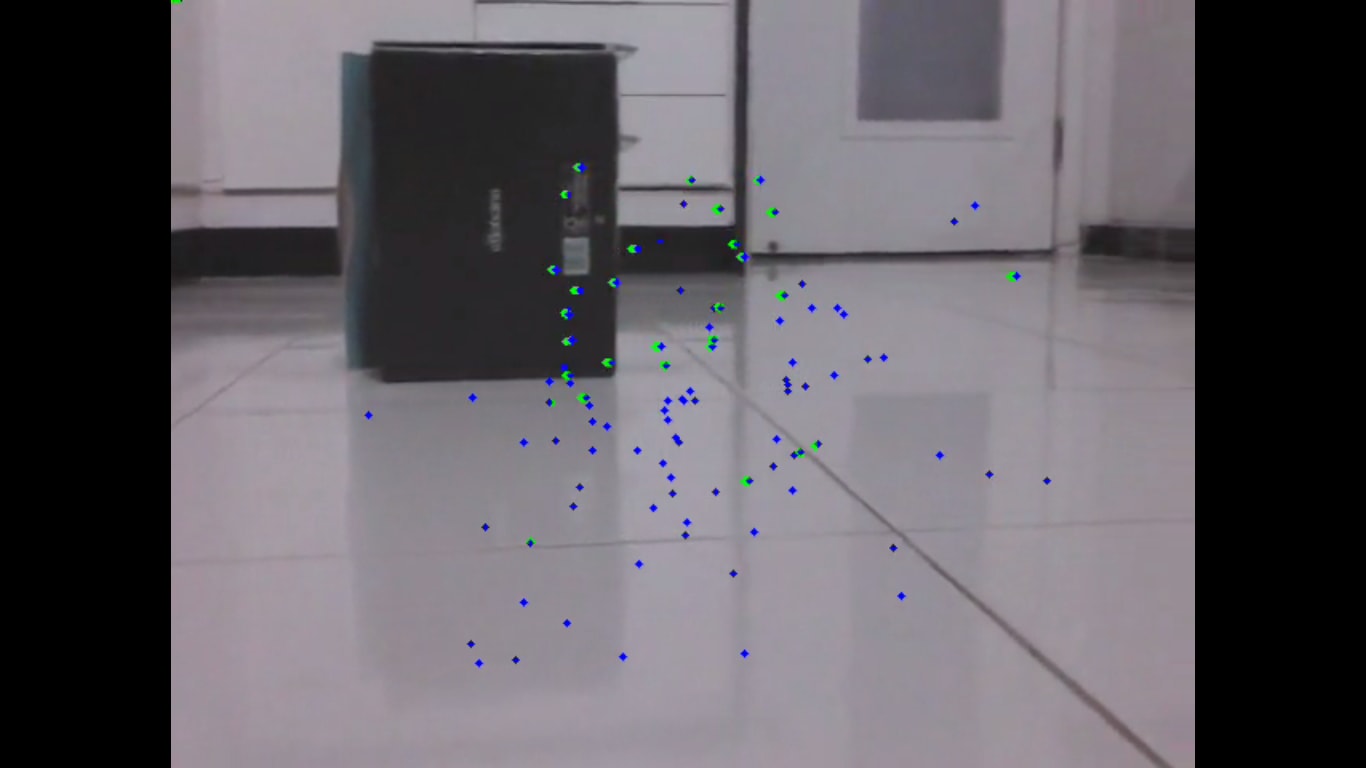}
		\caption{\label{fig:fluxo_pontos_gaussianos}}
	\end{subfigure}
	~
	\begin{subfigure}[b]{0.4\linewidth}
		\includegraphics[width=\linewidth]{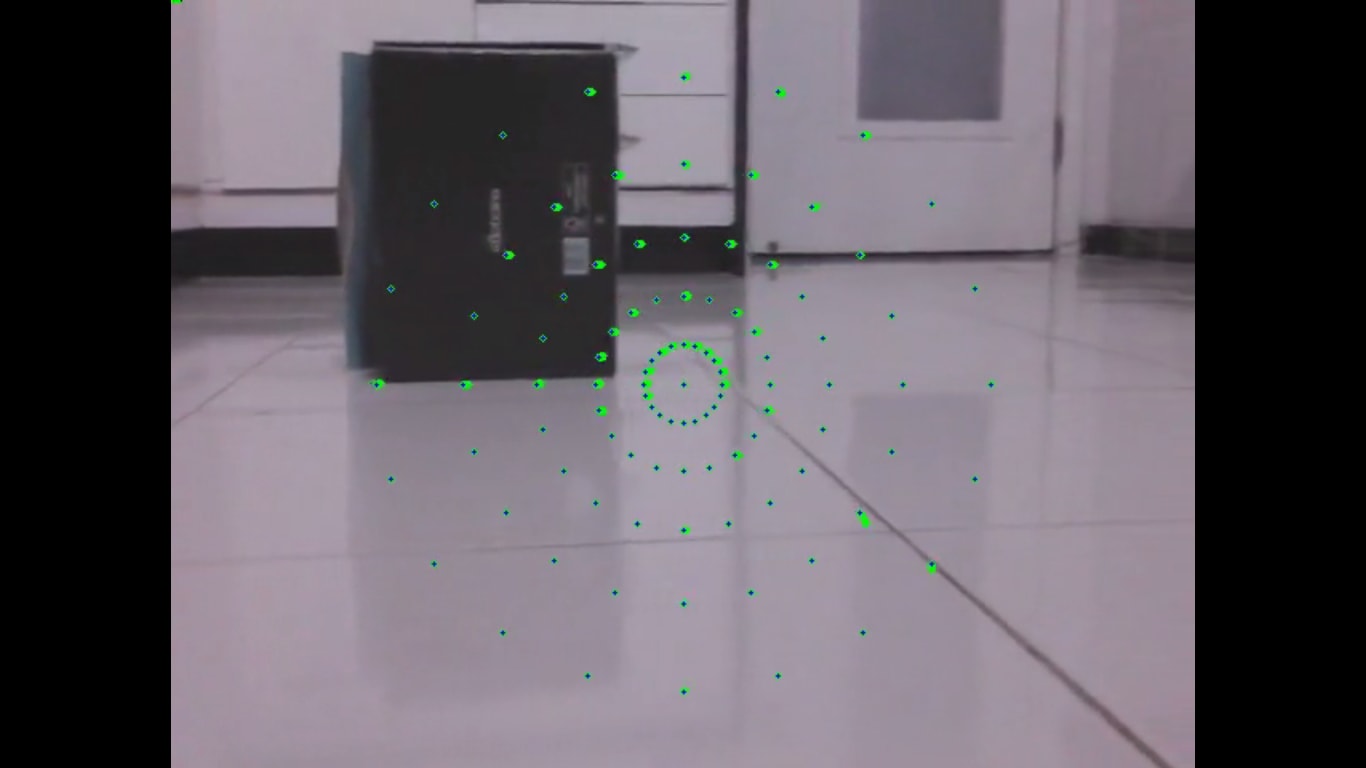}
		\caption{\label{fig:fluxo_pontos_circulares}}
	\end{subfigure}
	\caption{Diferentes aplicações do algoritmo de Lucas-Kanade. \ref{fig:fluxo_pontos_fixos} apresenta o fluxo sobre 100 pontos equidistantes,  \ref{fig:fluxo_melhores_pontos} apresenta o fluxo sobre os pontos com melhores características, \ref{fig:fluxo_pontos_gaussianos} apresenta o fluxo sobre pontos posicionados segundo uma distribuição gaussiana e \ref{fig:fluxo_pontos_circulares} apresenta o fluxo sobre pontos posicionados de forma circular.\label{fluxo_optico_obtido}}
\end{figure}

A \autoref{fig:fluxo_pontos_fixos} exibe o fluxo calculado sobre 100 pontos igualmente espaçados. Pelo fato de abranger todas as regiões da imagem, esta abordagem aparenta ser adequada para análises da movimentação geral do ambiente. Por exemplo, é possível estimar qual direção apresenta maior grau de movimentação média e, assim, direcionar o robô para a direção contrária. Porém, tal abordagem não aparenta ser viável para a identificação de obstáculos, já que a concentração de pontos de análise sobre obstáculos específicos tende a ser insuficiente para uma possível classificação. Já o fluxo exibido na \autoref{fig:fluxo_melhores_pontos} foi calculado apenas sobre pontos com características visuais distintas, como bordas e texturas realçadas. Tais pontos foram obtidos utilizando-se o algoritmo de Shi-Tomasi \cite{Shi1994}. Apesar da maior concentração dos pontos considerados, percebe-se que esta abordagem pode levar à análise de pontos que não sejam relevantes para a navegação do robô. É o caso de um corredor cujas paredes tenham texturas mais realçadas do que possíveis obstáculos que se encontrem logo à frente do caminho. 

Por sua vez, a \autoref{fig:fluxo_pontos_gaussianos} apresenta o fluxo calculado sobre 100 pontos posicionados de acordo com uma distribuição gaussiana cuja origem encontra-se no centro da imagem. Esta abordagem teve como objetivo gerar um foco sobre a região mais relevante para a detecção de obstáculos. Dessa forma, garante-se que tal região sempre será monitorada. Ao mesmo tempo, as demais regiões também são analisadas, porém com menor relevância. Essa abordagem aparenta ser mais adequada para este tipo de problema, uma vez que concentra em seu foco pontos suficientes para a classificação de obstáculos específicos. No entanto, a assimetria desta distribuição torna mais difícil a cobertura de obstáculos de maneira uniforme ao longo da navegação. Devido a isso, realizou-se o teste final com base na distribuição apresentada na \autoref{fig:fluxo_pontos_circulares}. Esta mantém a mesma propriedade da distribuição vista na \autoref{fig:fluxo_pontos_gaussianos}, uma vez que concentra a maior parte dos pontos no centro da imagem e reduz tal concentração de maneira semelhante a uma distribuição gaussiana. Além disso, esta distribuição é simétrica, de modo que as regiões com mesma distância do centro possuem a mesma probabilidade de cobrirem um obstáculo.

Sendo assim, com base nos testes realizados, escolheu-se pela distribuição circular vista na \autoref{fig:fluxo_pontos_circulares} para a extração do fluxo óptico neste trabalho.

% ----------------------------------------------------------
\chapter{\textit{Support Vector Machine} (SVM)}
\label{ap:svm}

Máquinas de Vetores de Suporte (SVMs, do Inglês \textit{Support Vector Machines}) correspondem a uma técnica de aprendizado de máquina baseada na maximização da margem que separa padrões. Para isso, esta técnica considera o mapeamento dos padrões para dimensões geralmente maiores que a dos vetores de características, de modo a encontrar-se o hiperplano que otimiza tal separação \cite{Theodoridis2003}. A formulação mais simples de uma SVM consiste na implementação linear com margem rígida. Para este caso, considere-se que o vetor de características de um determinado conjunto $X$ de padrões de treinamento seja dado por $x_i$, $i=1,2...N$. Tais padrões estão associados a apenas uma de duas possíveis classes, $w_1$ e $w_2$. Estas, por sua vez, são linearmente separáveis. Assim, o objetivo é encontrar um hiperplano
	
	\begin{equation}
		g(x) = w^Tx + w_0 = 0
	\end{equation} 

\noindent que classifique corretamente todos os vetores de treinamento. A \autoref{fig:svm_possiveis_fronteiras} ilustra dois possíveis hiperplanos capazes de separar os padrões de duas classes distintas. Como pode-se perceber, o hiperplano ilustrado pela reta contínua aparenta ser mais adequado para classificar novos padrões, uma vez que o mesmo provê maior espaço para que as amostras de cada classe sejam distribuídas sem que a regra de separação seja violada. Em outras palavras, o classificador formado por este hiperplano apresenta maior capacidade de generalização. Assim, entende-se que de modo geral o hiperplano a ser escolhido para a criação de classificadores deve ser capaz de maximizar a margem existente entre as duas classes separadas.

\begin{figure}[htb]
	\centering
	\includegraphics[width=0.7\linewidth]{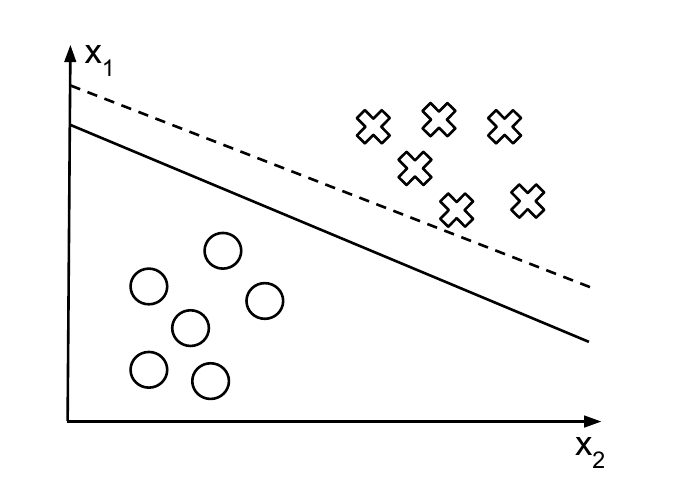}
	\caption{Exemplo de possíveis hiperplanos capazes de separar padrões de duas classes diferentes.\label{fig:svm_possiveis_fronteiras}}
\end{figure}

Para encontrar tal hiperplano, considera-se inicialmente que sua distância a um determinado ponto corresponde a

\begin{equation}
	z = \frac{|g(x)|}{||w||}
\end{equation}

Em seguida, assume-se que o valor de $g(x)$ sobre os pontos mais próximos de $w_1$ e $w_2$ (padrões preenchidos na \autoref{fig:svm_maximizacao_margens}) corresponda a $+1$ para $w_1$ e $-1$ para $w_2$. Esta consideração é equivalente às seguintes afirmações:

\begin{enumerate}
	\item A largura da margem desejada corresponde a $\frac{1}{||w||} + \frac{1}{||w||} = \frac{2}{||w||}$
	\item A restrição a ser atingida é igual a 
	
	\begin{equation}
		w^Tx + w_0 \geq 1 \textnormal{, } \forall x \in w_1
	\end{equation}
	
	\begin{equation}
	w^Tx + w_0 \leq -1 \textnormal{, } \forall x \in w_2
	\end{equation}
\end{enumerate}

\begin{figure}[htb]
	\centering
	\includegraphics[width=0.7\linewidth]{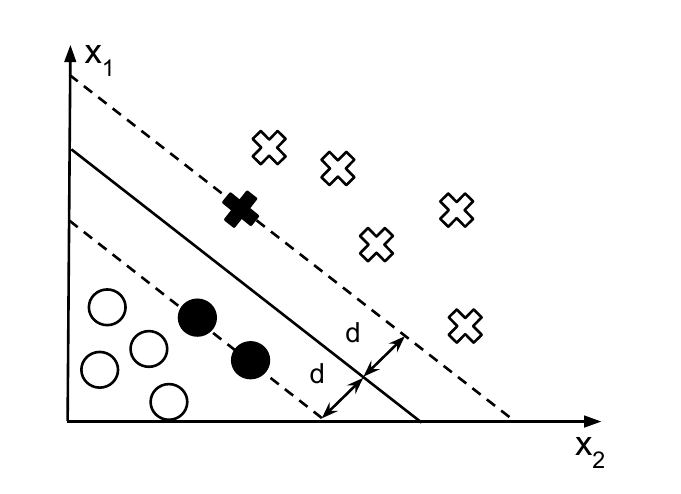}
	\caption{Exemplo de hiperplano (reta contínua) que maximiza a margem de separação entre padrões de duas classes distintas.\label{fig:svm_maximizacao_margens}}
\end{figure}

Finalmente, considerando que $y_i(x_i)$ corresponde ao valor da classe de $x_i$, tem-se que o problema de busca do hiperplano que maximiza as margens de separação é definido como:

\begin{equation} \label{eq:svm_minimizacao}
	\textnormal{Minimizar } J(w) \equiv \frac{1}{2}||w||^2
\end{equation}
\begin{equation} \label{eq:svm_restricoes}
	\textnormal{Sujeito a } y_i(w^Tx_i + w_0) \geq 1 \textnormal{, } i=1,2...N
\end{equation}

\noindent , sendo que as restrições da \autoref{eq:svm_restricoes} garantem que não haja dados de treinamento entre as margens de separação das classes. 

A \autoref{fig:svm_maximizacao_margens} ilustra o hiperplano ótimo (reta contínua) para um determinado problema de classificação. O classificador formado por este hiperplano é, portanto, chamado Máquina de Vetor de Suporte. Além disso, os vetores de suporte associados à máquina correspondem àqueles que atravessam os parões mais próximos à fronteira encontrada (os quais encontram-se preenchidos na \autoref{fig:svm_maximizacao_margens}).

Como dito anteriormente, a \autoref{eq:svm_minimizacao} e a \autoref{eq:svm_restricoes} definem a tarefa de busca de uma fronteira de separação linear. Para casos nos quais a mesma deve ser não-linear, consideram-se as seguintes equações \cite{Chih-WeiHsuChih-ChungChang2008}:

\begin{equation}
	\textnormal{Minimizar } J(w, w_0, \epsilon) \equiv \frac{1}{2}||w||^2 + C\sum_{i=1}^{N}{I(\epsilon_i)}
\end{equation}
\begin{equation}
	\textnormal{Sujeito a } y_i(w^T\phi(x_i) + w_0) \geq 1 - \epsilon_i \textnormal{, } i=1,2...N
\end{equation}
\begin{equation}
	\epsilon_i \geq 0 \textnormal{, } i=1,2...N
\end{equation}

\noindent , sendo $K(\phi(x_i), \phi(x_j)) = \phi(x_i)^T\phi(x_j)$ chamada função \textit{kernel}, $\epsilon_i$ conhecidas como \textit{variáveis de folga}, $\epsilon$ o  vetor de parâmetros $\epsilon_i$, $C$ o parâmetro de penalidade do termo de erro e

\begin{equation}
	I(\epsilon_i) =
	\left \{
	\begin{array}{cc}
	1, & \epsilon_i \geq 0  \\
	0, & \epsilon_i = 0 \\
	\end{array}
	\right.
\end{equation} 

\end{apendicesenv}

%% file: tcc.bbl
\providecommand{\abntreprintinfo}[1]{%
 \citeonline{#1}}
\setlength{\labelsep}{0pt}\begin{thebibliography}{}
\providecommand{\abntrefinfo}[3]{}
\providecommand{\abntbstabout}[1]{}
\abntbstabout{v-1.9.6 }

\end{thebibliography}


\providecommand{\abntreprintinfo}[1]{%
 \citeonline{#1}}
\setlength{\labelsep}{0pt}\begin{thebibliography}{}
\providecommand{\abntrefinfo}[3]{}
\providecommand{\abntbstabout}[1]{}
\abntbstabout{v-1.9.6 }

\bibitem[Angelici e Luiselli 2007]{Angelici2007}
\abntrefinfo{Angelici e Luiselli}{ANGELICI; LUISELLI}{2007}
{ANGELICI, F.~M.; LUISELLI, L. {Body Size and Altitude Partitioning of the
  Hares Lepus Europaeus and L. Corsicanus Living in Sympatry and Allopatry in
  Italy}.
\emph{Wildlife Biology}, v.~13, n. Toschi 1965, p. 251--257, 2007.
ISSN 0909-6396.}

\bibitem[Anton, Bivens e Davis 2014]{Anton2014}
\abntrefinfo{Anton, Bivens e Davis}{ANTON; BIVENS; DAVIS}{2014}
{ANTON, H.; BIVENS, I.; DAVIS, S. \emph{Cálculo - V2 (Portuguese Edition)}.
  Bookman, 2014.
ISBN 9788582602461. Dispon{\'\i}vel em:
  \url{https://www.amazon.com/C\%C3\%A1lculo-V2-Portuguese-Howard-Anton-ebook/dp/B015WUBGAU?SubscriptionId=0JYN1NVW651KCA56C102&tag=techkie-20&linkCode=xm2&camp=2025&creative=165953&creativeASIN=B015WUBGAU}.}

\bibitem[Barron, Fleet e Beauchemin 1994]{Barron1994}
\abntrefinfo{Barron, Fleet e Beauchemin}{BARRON; FLEET; BEAUCHEMIN}{1994}
{BARRON, J.~L.; FLEET, D.~J.; BEAUCHEMIN, S.~S. {Systems and Experiment
  Performance of optical flow techniques}.
\emph{International Journal of Computer Vision}, v.~12, n.~1, p. 43--77, 1994.
ISSN 0920-5691.
Dispon{\'\i}vel em: \url{http://link.springer.com/10.1007/BF01420984}.}

\bibitem[BeagleBoard.org 2016]{Beagle}
\abntrefinfo{BeagleBoard.org}{BEAGLEBOARD.ORG}{2016}
{BEAGLEBOARD.ORG. \emph{BeagleBoard.org - community supported open hardware
  computers for making}. 2016.
Dispon{\'\i}vel em: \url{https://beagleboard.org/}.}

\bibitem[Bekey 2005]{Bekey2005}
\abntrefinfo{Bekey}{BEKEY}{2005}
{BEKEY, G.~A. \emph{Autonomous Robots: From Biological Inspiration to
  Implementation and Control (Intelligent Robotics and Autonomous Agents
  series)}. A Bradford Book, 2005.
ISBN 9780262534185. Dispon{\'\i}vel em:
  \url{https://www.amazon.com/Autonomous-Robots-Inspiration-Implementation-Intelligent-ebook/dp/B006H2CD7I?SubscriptionId=0JYN1NVW651KCA56C102&tag=techkie-20&linkCode=xm2&camp=2025&creative=165953&creativeASIN=B006H2CD7I}.}

\bibitem[Boroujeni 2012]{Boroujeni2012}
\abntrefinfo{Boroujeni}{BOROUJENI}{2012}
{BOROUJENI, N.~S. {Fast obstacle detection using targeted optical flow}.
\emph{{\ldots} (ICIP), 2012 19th IEEE {\ldots}}, v.~2, n.~3, p. 65--68, 2012.}

\bibitem[Bouguet 2001]{Bouguet2001}
\abntrefinfo{Bouguet}{BOUGUET}{2001}
{BOUGUET, J.-Y. {Pyramidal implementation of the affine lucas kanade feature
  tracker—description of the algorithm}.
\emph{Pages.Slc.Edu}, v.~2, p.~3, 2001.
ISSN 10828907.}

\bibitem[Caldeira, Schneebeli e Sarcinelli-Filho 2007]{Caldeira2007}
\abntrefinfo{Caldeira, Schneebeli e Sarcinelli-Filho}{CALDEIRA; SCHNEEBELI;
  SARCINELLI-FILHO}{2007}
{CALDEIRA, E. M. D.~O.; SCHNEEBELI, H. J.~A.; SARCINELLI-FILHO, M. {An optical
  flow-based sensing system for reactive mobile robot navigation}.
\emph{Sba: Controle {\&} Automa{\c{c}}{\~{a}}o Sociedade Brasileira de
  Automatica}, v.~18, n.~3, p. 265--277, 2007.
ISSN 0103-1759.}

\bibitem[Canny 1986]{Canny_1986}
\abntrefinfo{Canny}{CANNY}{1986}
{CANNY, J. A computational approach to edge detection.
\emph{{IEEE} Transactions on Pattern Analysis and Machine Intelligence},
  Institute of Electrical and Electronics Engineers ({IEEE}), {PAMI}-8, n.~6,
  p. 679--698, nov 1986.}

\bibitem[Chaiyasoonthorn, Hongyim e Mitatha 2015]{Chaiyasoonthorn2015}
\abntrefinfo{Chaiyasoonthorn, Hongyim e Mitatha}{CHAIYASOONTHORN; HONGYIM;
  MITATHA}{2015}
{CHAIYASOONTHORN, S.; HONGYIM, N.; MITATHA, S. Building automatic packet report
  system to report position and radiation data for autonomous robot in the
  disaster area. In:  \emph{2015 15th International Conference on Control,
  Automation and Systems ({ICCAS})}. Institute of Electrical and Electronics
  Engineers ({IEEE}), 2015. Dispon{\'\i}vel em:
  \url{http://dx.doi.org/10.1109/ICCAS.2015.7364883}.}

\bibitem[{Chih-Wei Hsu, Chih-Chung Chang} e Lin
  2008]{Chih-WeiHsuChih-ChungChang2008}
\abntrefinfo{{Chih-Wei Hsu, Chih-Chung Chang} e Lin}{{Chih-Wei Hsu, Chih-Chung
  Chang}; LIN}{2008}
{{Chih-Wei Hsu, Chih-Chung Chang}; LIN, C.-J. {A Practical Guide to Support
  Vector Classification}.
\emph{BJU international}, v.~101, n.~1, p. 1396--400, 2008.
ISSN 1464-410X.
Dispon{\'\i}vel em:
  \url{http://www.csie.ntu.edu.tw/{~}cjlin/papers/guide/guide.p}.}

\bibitem[Cho et al. 2013]{Cho2013}
\abntrefinfo{Cho et al.}{CHO et al.}{2013}
{CHO, B.-S. et al. Positioning of a mobile robot based on odometry and a new
  ultrasonic lps.
\emph{International Journal of Control, Automation and Systems}, v.~11, n.~2,
  p. 333--345, 2013.
ISSN 2005-4092.
Dispon{\'\i}vel em: \url{http://dx.doi.org/10.1007/s12555-012-0045-x}.}

\bibitem[Conrad e DeSouza 2010]{Conrad2010}
\abntrefinfo{Conrad e DeSouza}{CONRAD; DESOUZA}{2010}
{CONRAD, D.; DESOUZA, G.~N. {Homography-based ground plane detection for mobile
  robot navigation using a Modified EM algorithm}.
\emph{2010 IEEE International Conference on Robotics and Automation}, p.
  910--915, 2010.
Dispon{\'\i}vel em:
  \url{http://ieeexplore.ieee.org/lpdocs/epic03/wrapper.htm?arnumber=5509457}.}

\bibitem[Corporation 2012]{Corporation2012}
\abntrefinfo{Corporation}{CORPORATION}{2012}
{CORPORATION, B. {BCM2835 ARM Peripherals}.
\emph{Science}, 2012.}

\bibitem[Croon 2015]{Croon2015}
\abntrefinfo{Croon}{CROON}{2015}
{CROON, G. C. H. E.~G. de. {Distance estimation with efference copies and
  optical flow maneuvers : a stability-based strategy .}
\emph{Bioinspiration {\&} Biomimetics}, IOP Publishing, v.~11, n.~1, p. 1--30,
  2015.
ISSN 1748-3190.
Dispon{\'\i}vel em: \url{http://dx.doi.org/10.1088/1748-3190/11/1/016004}.}

\bibitem[Croston 2017]{Croston2017}
\abntrefinfo{Croston}{CROSTON}{2017}
{CROSTON, B. \emph{RPi.GPIO 0.6.3}. 2017.
Dispon{\'\i}vel em: \url{https://pypi.python.org/pypi/RPi.GPIO}.}

\bibitem[Drucker et al. 1997]{Drucker1997}
\abntrefinfo{Drucker et al.}{DRUCKER et al.}{1997}
{DRUCKER, H. et al. {Support vector regression machines}.
\emph{Advances in Neural Information Processing Systems}, v.~1, p. 155--161,
  1997.
ISSN 10495258.
Dispon{\'\i}vel em:
  \url{http://papers.nips.cc/paper/1238-support-vector-regression-machines.pdf}.}

\bibitem[Duda, Hart e Stork 2001]{Duda2001}
\abntrefinfo{Duda, Hart e Stork}{DUDA; HART; STORK}{2001}
{DUDA, R.~O.; HART, P.~E.; STORK, D.~G. \emph{{Pattern Classification}}. 2001.
  680~p.}

\bibitem[Eletric 2016]{PowerAPC}
\abntrefinfo{Eletric}{ELETRIC}{2016}
{ELETRIC, S. \emph{Power Pack APC - Carregador Portátil - Power Bank -
  Schneider Eletric}. 2016.
Dispon{\'\i}vel em: \url{http://powerpackapc.com.br/}.}

\bibitem[Filho 2016]{Filho2016}
\abntrefinfo{Filho}{FILHO}{2016}
{FILHO, A. \emph{Fundamentos de Cálculo Numérico}. [S.l.]: Bookman, 2016.
ISBN 9788582603857.}

\bibitem[Fleet e Weiss 2005]{Fleet2005}
\abntrefinfo{Fleet e Weiss}{FLEET; WEISS}{2005}
{FLEET, D.; WEISS, Y. {Optical Flow Estimation}.
\emph{Mathematical models for Computer Vision: The Handbook}, p. 239--257,
  2005.
ISSN 1941-0042.
Dispon{\'\i}vel em: \url{http://eprints.pascal-network.org/archive/00001065/}.}

\bibitem[Foundation 2017]{Foundation2017}
\abntrefinfo{Foundation}{FOUNDATION}{2017}
{FOUNDATION, P.~S. \emph{Python 2.7.13 documentation}. 2017.
Dispon{\'\i}vel em: \url{https://docs.python.org/2/}.}

\bibitem[Foundation 2016]{RaspberryGPIO}
\abntrefinfo{Foundation}{FOUNDATION}{2016a}
{FOUNDATION, R.~P. \emph{GPIO - Raspberry Pi Documentation}. 2016.
Dispon{\'\i}vel em:
  \url{https://www.raspberrypi.org/documentation/usage/gpio-plus-and-raspi2/}.}

\bibitem[Foundation 2016]{RaspberryPower}
\abntrefinfo{Foundation}{FOUNDATION}{2016b}
{FOUNDATION, R.~P. \emph{Power Supply - Raspberry Pi Documentation}. 2016.
Dispon{\'\i}vel em:
  \url{https://www.raspberrypi.org/documentation/hardware/raspberrypi/power/README.md}.}

\bibitem[Foundation 2016]{Raspberry}
\abntrefinfo{Foundation}{FOUNDATION}{2016c}
{FOUNDATION, R.~P. \emph{Raspberry Pi - Teach, Learn, and Make with Raspberry
  Pi}. 2016.
Dispon{\'\i}vel em: \url{https://www.raspberrypi.org/}.}

\bibitem[Foundation 2016]{Raspberry3}
\abntrefinfo{Foundation}{FOUNDATION}{2016d}
{FOUNDATION, R.~P. \emph{Raspberry Pi 3 Model B - Raspberry Pi}. 2016.
Dispon{\'\i}vel em:
  \url{https://www.raspberrypi.org/products/raspberry-pi-3-model-b/}.}

\bibitem[Foundation 2016]{Raspbian}
\abntrefinfo{Foundation}{FOUNDATION}{2016e}
{FOUNDATION, R.~P. \emph{RaspbianAbout - Raspbian}. 2016.
Dispon{\'\i}vel em: \url{https://www.raspbian.org/RaspbianAbout}.}

\bibitem[Ganganath, Cheng e Tse 2015]{Ganganath2015}
\abntrefinfo{Ganganath, Cheng e Tse}{GANGANATH; CHENG; TSE}{2015}
{GANGANATH, N.; CHENG, C.-t.; TSE, C.~K. {A Constraint-Aware Heuristic Path
  Planner for Finding Energy-Efficient Paths on Uneven Terrains}.
\emph{IEEE Transactions on Industrial Informatics}, v.~11, n.~3, p. 601--611,
  2015.
ISSN 1551-3203.
Dispon{\'\i}vel em:
  \url{http://ieeexplore.ieee.org/lpdocs/epic03/wrapper.htm?arnumber=7061469}.}

\bibitem[Holdings 2016]{ARM}
\abntrefinfo{Holdings}{HOLDINGS}{2016}
{HOLDINGS, A. \emph{Home – ARM}. 2016.
Dispon{\'\i}vel em: \url{http://www.arm.com/}.}

\bibitem[Horn e Schunck 1981]{Horn1981}
\abntrefinfo{Horn e Schunck}{HORN; SCHUNCK}{1981}
{HORN, B. K.~P.; SCHUNCK, B.~G. {Determining optical flow}.
\emph{Artificial Intelligence}, v.~17, n.~1-3, p. 185--203, 1981.
ISSN 00043702.}

\bibitem[Hotelling 1933]{hotelling1933analysis}
\abntrefinfo{Hotelling}{HOTELLING}{1933}
{HOTELLING, H. Analysis of a complex of statistical variables into principal
  components.
\emph{Journal of educational psychology}, Warwick \& York, v.~24, n.~6, p.~417,
  1933.}

\bibitem[Incorporated 2002]{Incorporated2002}
\abntrefinfo{Incorporated}{INCORPORATED}{2002}
{INCORPORATED, T.~I. {LM317L 3-Terminal Adjustable Regulator LM317L 3-Terminal
  Adjustable Regulator}.
v.~5, n. March, p.~1--4, 2002.}

\bibitem[Incorporated 2016]{Incorporated2016}
\abntrefinfo{Incorporated}{INCORPORATED}{2016}
{INCORPORATED, T.~I. {LM2679 SIMPLE SWITCHER {\textregistered} 5-A Step-Down
  Voltage Regulator}.
2016.}

\bibitem[Indoware 2013]{Indoware2013}
\abntrefinfo{Indoware}{INDOWARE}{2013}
{INDOWARE. {Ultrasonic Ranging Module HC - SR04}.
\emph{Datasheet}, p.~1--4, 2013.
Dispon{\'\i}vel em: \url{http://www.micropik.com/PDF/HCSR04.pdf}.}

\bibitem[Intel 2017]{Intel2017}
\abntrefinfo{Intel}{INTEL}{2017}
{INTEL. \emph{Threading Building Blocks}. 2017.
Dispon{\'\i}vel em: \url{https://www.threadingbuildingblocks.org/}.}

\bibitem[Kadir et al. 2015]{Kadir2015}
\abntrefinfo{Kadir et al.}{KADIR et al.}{2015}
{KADIR, M.~A. et al. An autonomous industrial robot for loading and unloading
  goods. In:  \emph{2015 International Conference on Informatics, Electronics
  {\&} Vision ({ICIEV})}. Institute of Electrical and Electronics Engineers
  ({IEEE}), 2015. Dispon{\'\i}vel em:
  \url{http://dx.doi.org/10.1109/ICIEV.2015.7333984}.}

\bibitem[Kim et al. 2012]{Kim2012}
\abntrefinfo{Kim et al.}{KIM et al.}{2012}
{KIM, D. et al. Development of jellyfish removal robot system {JEROS}. In:
  \emph{2012 9th International Conference on Ubiquitous Robots and Ambient
  Intelligence ({URAI})}. Institute of Electrical and Electronics Engineers
  ({IEEE}), 2012. Dispon{\'\i}vel em:
  \url{http://dx.doi.org/10.1109/URAI.2012.6463092}.}

\bibitem[Kohavi et al. 1995]{kohavi1995study}
\abntrefinfo{Kohavi et al.}{KOHAVI et al.}{1995}
{KOHAVI, R. et al. A study of cross-validation and bootstrap for accuracy
  estimation and model selection. In:  STANFORD, CA. \emph{Ijcai}. [S.l.],
  1995. v.~14, n.~2, p. 1137--1145.}

\bibitem[Koo et al. 2013]{Koo_2013}
\abntrefinfo{Koo et al.}{KOO et al.}{2013}
{KOO, I.~M. et al. Development of a quadruped walking robot {AiDIN}-{III} using
  biologically inspired kinematic analysis.
\emph{International Journal of Control, Automation and Systems}, Springer
  Nature, v.~11, n.~6, p. 1276--1289, nov 2013.}

\bibitem[Kostavelis et al. 2016]{Kostavelis2016}
\abntrefinfo{Kostavelis et al.}{KOSTAVELIS et al.}{2016}
{KOSTAVELIS, I. et al. {Stereo-based Visual Odometry for Autonomous Robot
  Navigation}.
\emph{International Journal of Advanced Robotic Systems}, p.~1, 2016.
ISSN 1729-8806.
Dispon{\'\i}vel em:
  \url{http://www.intechopen.com/journals/international{\\_}journal{\\_}of{\\_}advanced{\\_}robotic{\\_}systems/stereo-based-visual-odometry-for-autonomous-r}.}

\bibitem[Kraus 2013]{Kraus2013}
\abntrefinfo{Kraus}{KRAUS}{2013}
{KRAUS, S. H. W. R. E. F. A.~L. \emph{The Biology of the Laboratory Rabbit}.
  Elsevier Science, 2013. Dispon{\'\i}vel em:
  \url{http://www.ebook.de/de/product/23226530/the\_biology\_of\_the\_laboratory\_rabbit.html}.}

\bibitem[Lee e Chung 2015]{Lee2015}
\abntrefinfo{Lee e Chung}{LEE; CHUNG}{2015}
{LEE, W.; CHUNG, W. Position estimation using multiple low-cost {GPS} receivers
  for outdoor mobile robots. In:  \emph{2015 12th International Conference on
  Ubiquitous Robots and Ambient Intelligence ({URAI})}. Institute of Electrical
  and Electronics Engineers ({IEEE}), 2015. Dispon{\'\i}vel em:
  \url{http://dx.doi.org/10.1109/URAI.2015.7358906}.}

\bibitem[LG 2016]{LG2016}
\abntrefinfo{LG}{LG}{2016}
{LG, G. \emph{LG Portugal :: Acessórios TV - Câmara Videoconferência
  AN-VC500}. 2016.
Dispon{\'\i}vel em:
  \url{http://www.lg.com/pt/acessorios-tv/lg-AN-VC500-camara-smart}.}

\bibitem[Li e Birchfield 2010]{Li2010}
\abntrefinfo{Li e Birchfield}{LI; BIRCHFIELD}{2010}
{LI, Y.; BIRCHFIELD, S.~T. {Image-based segmentation of indoor corridor floors
  for a mobile robot}.
\emph{IEEE/RSJ 2010 International Conference on Intelligent Robots and Systems,
  IROS 2010 - Conference Proceedings}, p. 837--843, 2010.
ISSN 2153-0858.}

\bibitem[Lins et al. 2016]{Lins2016}
\abntrefinfo{Lins et al.}{LINS et al.}{2016}
{LINS, R.~G. et al. {A Novel Machine Vision Approach Applied for Autonomous
  Robotics Navigation}.
\emph{Proceedings - 2015 IEEE International Conference on Systems, Man, and
  Cybernetics, SMC 2015}, p. 1912--1917, 2016.}

\bibitem[Lipton, Berkowitz e Elkan 2015]{Lipton2015}
\abntrefinfo{Lipton, Berkowitz e Elkan}{LIPTON; BERKOWITZ; ELKAN}{2015}
{LIPTON, Z.~C.; BERKOWITZ, J.; ELKAN, C. {A Critical Review of Recurrent Neural
  Networks for Sequence Learning}.
p. 1--38, 2015.
ISSN 9781450330633.
Dispon{\'\i}vel em: \url{http://arxiv.org/abs/1506.00019}.}

\bibitem[Lucas e Kanade 1981]{Lucas1981}
\abntrefinfo{Lucas e Kanade}{LUCAS; KANADE}{1981}
{LUCAS, B.~D.; KANADE, T. {An Iterative Image Registration Technique with an
  Application to Stereo Vision}.
\emph{Imaging}, v.~130, n.~x, p. 674--679, 1981.
ISSN 17486815.
Dispon{\'\i}vel em:
  \url{http://citeseerx.ist.psu.edu/viewdoc/download?doi=10.1.1.49.2019{\&}rep=rep1{\&}ty}.}

\bibitem[Pedregosa et al. 2011]{scikit-learn}
\abntrefinfo{Pedregosa et al.}{PEDREGOSA et al.}{2011}
{PEDREGOSA, F. et al. Scikit-learn: Machine learning in {P}ython.
\emph{Journal of Machine Learning Research}, v.~12, p. 2825--2830, 2011.}

\bibitem[Quesenberry e Carpenter 2011]{Quesenberry2011}
\abntrefinfo{Quesenberry e Carpenter}{QUESENBERRY; CARPENTER}{2011}
{QUESENBERRY, K.; CARPENTER, J.~W. \emph{Ferrets, Rabbits and Rodents -
  E-Book}. Elsevier Health Sciences, 2011. Dispon{\'\i}vel em:
  \url{http://www.ebook.de/de/product/23019441/katherine\_quesenberry\_james\_w\_carpenter\_ferrets\_rabbits\_and\_rodents\_e\_book.html}.}

\bibitem[{R Core Team} 2013]{RCT2013}
\abntrefinfo{{R Core Team}}{{R Core Team}}{2013}
{{R Core Team}. \emph{R: A Language and Environment for Statistical Computing}.
Vienna, Austria, 2013. Dispon{\'\i}vel em: \url{http://www.R-project.org/}.}

\bibitem[Rosenblatt 1958]{Rosenblatt1958}
\abntrefinfo{Rosenblatt}{ROSENBLATT}{1958}
{ROSENBLATT, F. The perceptron: A probabilistic model for information storage
  and organization in the brain.
\emph{Psychological Review}, American Psychological Association ({APA}), v.~65,
  n.~6, p. 386--408, 1958.}

\bibitem[Sanchez-Garcia et al. 2015]{Sanchez-Garcia2015}
\abntrefinfo{Sanchez-Garcia et al.}{SANCHEZ-GARCIA et al.}{2015}
{SANCHEZ-GARCIA, A.~J. et al. {Decision making for obstacle avoidance in
  autonomous mobile robots by time to contact and optical flow}.
\emph{25th International Conference on Electronics, Communications and
  Computers, CONIELECOMP 2015}, p. 130--134, 2015.}

\bibitem[SanDisk 2017]{SanDisk2017}
\abntrefinfo{SanDisk}{SANDISK}{2017}
{SANDISK. \emph{SanDisk | Líder Global em Soluções de Armazenamento de
  Memória Flash}. 2017.
Dispon{\'\i}vel em: \url{https://www.sandisk.com.br/}.}

\bibitem[Shankar, Vatsa e Sujit 2014]{Shankar2014}
\abntrefinfo{Shankar, Vatsa e Sujit}{SHANKAR; VATSA; SUJIT}{2014}
{SHANKAR, A.; VATSA, M.; SUJIT, P.~B. {Collision avoidance for a low-cost robot
  using SVM-based monocular vision}.
\emph{2014 IEEE International Conference on Robotics and Biomimetics, IEEE
  ROBIO 2014}, p. 277--282, 2014.}

\bibitem[Shi e Tomasi 1994]{Shi1994}
\abntrefinfo{Shi e Tomasi}{SHI; TOMASI}{1994}
{SHI, J.; TOMASI. Good features to track. In:  \emph{Proceedings of {IEEE}
  Conference on Computer Vision and Pattern Recognition {CVPR}-94}. Institute
  of Electrical and Electronics Engineers ({IEEE}), 1994. Dispon{\'\i}vel em:
  \url{http://dx.doi.org/10.1109/CVPR.1994.323794}.}

\bibitem[Siegwart e Nourbakhsh 2004]{Siegwart2004}
\abntrefinfo{Siegwart e Nourbakhsh}{SIEGWART; NOURBAKHSH}{2004}
{SIEGWART, R.; NOURBAKHSH, I.~R. \emph{Introduction to Autonomous Mobile
  Robots}. [S.l.]: Massachusetts Institute of Technology, 2004.}

\bibitem[Sourceforge 2017]{Sourceforge2017}
\abntrefinfo{Sourceforge}{SOURCEFORGE}{2017}
{SOURCEFORGE. \emph{Win32 Disk Imager / Wiki / Home}. 2017.
Dispon{\'\i}vel em:
  \url{https://sourceforge.net/p/win32diskimager/wiki/Home/}.}

\bibitem[Stringhini Ilana de Almeida~Souza 2011]{Stringhini2011}
\abntrefinfo{Stringhini Ilana de Almeida~Souza}{STRINGHINI ILANA DE
  ALMEIDA~SOUZA}{2011}
{STRINGHINI ILANA DE ALMEIDA~SOUZA, L. A. d. S. e. M.~M. D. \emph{Visão
  Computacional Usando OpenCV}. [S.l.]: Universidade Presbiteriana Mackenzie,
  2011.}

\bibitem[SUI et al. 2011]{JINXUE2011}
\abntrefinfo{SUI et al.}{SUI et al.}{2011}
{SUI, J. et al. {Laser Measurement Key Technologies and Application in Robot
  Autonomous Navigation}.
\emph{International Journal of Pattern Recognition and Artificial
  Intelligence}, v.~25, n.~07, p. 1127--1146, 2011.
ISSN 0218-0014.
Dispon{\'\i}vel em:
  \url{http://www.worldscientific.com/doi/abs/10.1142/S0218001411008920}.}

\bibitem[Szeliski 2010]{Szeliski2010}
\abntrefinfo{Szeliski}{SZELISKI}{2010}
{SZELISKI, R. \emph{Computer Vision}. Springer London, 2010. Dispon{\'\i}vel
  em:
  \url{http://www.ebook.de/de/product/19111262/richard\_szeliski\_computer\_vision.html}.}

\bibitem[Team 2016]{OpenCV}
\abntrefinfo{Team}{TEAM}{2016}
{TEAM, O.~D. \emph{ABOUT | OpenCV}. 2016.
Dispon{\'\i}vel em: \url{http://opencv.org/about.html}.}

\bibitem[Team 2017]{OpenCV3-2}
\abntrefinfo{Team}{TEAM}{2017}
{TEAM, O.~D. \emph{OpenCV 3.2 | OpenCV}. 2017.
Dispon{\'\i}vel em: \url{http://opencv.org/opencv-3-2.html}.}

\bibitem[{Texas Instruments Incorporated} 2016]{TII2016}
\abntrefinfo{{Texas Instruments Incorporated}}{{Texas Instruments
  Incorporated}}{2016}
{{Texas Instruments Incorporated}. {L293x Quadruple Half-H Drivers}.
\emph{Texas Instruments Incorporated}, p.~21, 2016.
Dispon{\'\i}vel em: \url{http://www.ti.com/lit/ds/symlink/l293.pdf}.}

\bibitem[Theodoridis 2003]{Theodoridis2003}
\abntrefinfo{Theodoridis}{THEODORIDIS}{2003}
{THEODORIDIS, K. S.~K. \emph{Pattern Recognition}. 4th. ed. [S.l.]: Elsevier
  Academic Press, 2003.}

\bibitem[Walt, Colbert e Varoquaux 2011]{van_der_Walt_2011}
\abntrefinfo{Walt, Colbert e Varoquaux}{WALT; COLBERT; VAROQUAUX}{2011}
{WALT, S. van~der; COLBERT, S.~C.; VAROQUAUX, G. The {NumPy} array: A structure
  for efficient numerical computation.
\emph{Computing in Science {\&} Engineering}, Institute of Electrical and
  Electronics Engineers ({IEEE}), v.~13, n.~2, p. 22--30, mar 2011.}

\bibitem[Wang, Chen e Yin 2015]{Wang2015}
\abntrefinfo{Wang, Chen e Yin}{WANG; CHEN; YIN}{2015}
{WANG, L.; CHEN, F.; YIN, H. {Detecting and tracking vehicles in traffic by
  unmanned aerial vehicles}.
\emph{Automation in Construction}, n.~May, 2015.
ISSN 09265805.}

\end{thebibliography}
